\documentclass[journal]{IEEEtran}
\usepackage{cite}
\usepackage[pdftex]{graphicx}
\usepackage{amsmath,bm,amssymb}
\usepackage{algorithm,algorithmic}
\usepackage{array}
\usepackage[colorlinks,linkcolor=blue,anchorcolor=blue, citecolor=blue]{hyperref}
\usepackage{fixltx2e}
\usepackage{stfloats}
\usepackage{url}
\usepackage{multirow}  
\usepackage{makecell}  
\usepackage{float}    
\usepackage{subfigure}

\begin{document}

	\title{Neural Gradient Regularizer}

	\author{Shuang~Xu,
		Yifan~Wang,
		Zixiang~Zhao,
		Jiangjun~Peng,
		Xiangyong~Cao,
		Deyu~Meng,
            Yulun~Zhang, 
		Radu~Timofte,
		Luc~Van~Gool 
  \thanks{Shuang Xu is with the School of Mathematics and Statistics, Northwestern Polytechnical University, Xi’an 710021, China, and also with the Research and Development Institute of Northwestern Polytechnical University in Shenzhen, Shenzhen 518063, China.}
		\thanks{Yifan Wang and Xiangyong Cao are with the School of Electronic and Information Engineering and the Key Laboratory for Intelligent Networks and Network Security, Ministry of Education, Xi’an Jiaotong University, Xi’an 710049, China.}
		\thanks{Zixiang~Zhao, Jiangjun Peng and Deyu Meng are with the School of Mathematics and Statistics, Xi’an Jiaotong University, Xi’an 710049, China.}
            \thanks{Yulun Zhang, Radu Timofte and Luc Van Gool are with the Computer Vision Lab, ETH Zurich, 8092 Z\"{u}rich, Switzerland.}
            

}


\maketitle

\begin{abstract}
	Owing to its significant success, the prior imposed on gradient maps has consistently been a subject of great interest in the field of image processing. Total variation (TV), one of the most representative regularizers, is known for its ability to capture the intrinsic sparsity prior underlying gradient maps. Nonetheless, TV and its variants often underestimate the gradient maps, leading to the weakening of edges and details whose gradients should not be zero in the original image (i.e., image structures is not describable by sparse priors of gradient maps). Recently, total deep variation (TDV) has been introduced, assuming the sparsity of feature maps, which provides a flexible regularization learned from large-scale datasets for a specific task. However, TDV requires to retrain the network with image/task variations, limiting its versatility. To alleviate this issue, in this paper, we propose a neural gradient regularizer (NGR) that expresses the gradient map as the output of a neural network. Unlike existing methods, NGR does not rely on any subjective sparsity or other prior assumptions on image gradient maps, thereby avoiding the underestimation of gradient maps. NGR is applicable to various image types and different image processing tasks, functioning in a zero-shot learning fashion, making it a versatile and plug-and-play regularizer. Extensive experimental results demonstrate the superior performance of NGR over state-of-the-art counterparts for a range of different tasks, further validating its effectiveness and versatility.
\end{abstract}

\begin{IEEEkeywords}
	Deep image prior, unsupervised deep learning, low-level vision, total variation
\end{IEEEkeywords}

\IEEEpeerreviewmaketitle

\section{Introduction}
\IEEEPARstart{A}{n} image restoration task generally aims at recovering a high-quality image $\bm{\mathcal{X}}$ from its corrupted observation $\bm{\mathcal{Y}}$, by solving the following optimization problem \cite{Microscopy_Images_Denoising_Review}:
\begin{equation}
	\min_{\bm{\mathcal{X}}\in\mathbb{R}^{H\times W\times C} } L(\bm{\mathcal{X}},\bm{\mathcal{Y}})+\lambda R(\bm{\mathcal{X}}),
\end{equation}
where $H$, $W$ and $C$ denote the height, width and the number of channels, respectively. $L(\bm{\mathcal{X}},\bm{\mathcal{Y}})$ denotes the data fidelity term measuring the difference between $\bm{\mathcal{X}}$ and $\bm{\mathcal{Y}}$, $ R(\bm{\mathcal{X}})$ denotes the regularizer term encoding the prior knowledge imposed on the recovered image, and $\lambda$ is a hyper-parameter. Over the past few decades, remarkable progress has been made in data prior modeling \cite{LRLRR,DCP,ROP}, with image gradient research being a focal point of discussion over the image restoration research field \cite{PanHS017,ZhengLKXGK13,Gradient_Profile_Prior,Learn_Gradient_Profile}.

\begin{figure*}
	\centering
	\includegraphics[width=.75\linewidth]{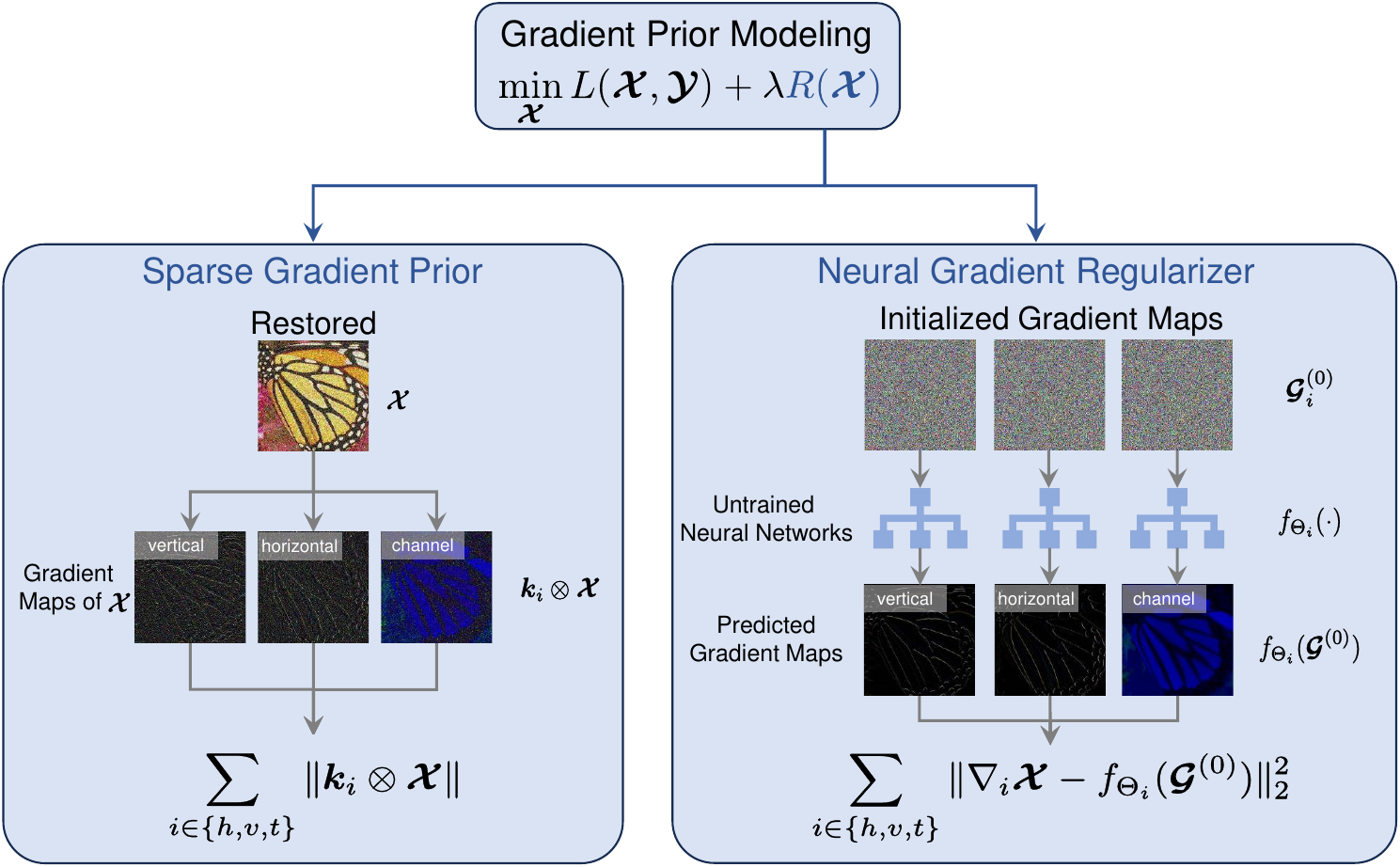}
	\caption{The mechanism of sparse gradient priors (left) that minimizes a certain norm of gradient/feature maps, and the proposed neural gradient regularizer (right) that encourages an untrained neural network to predict the gradient maps of a high-quality image. }
	\label{fig:toc}
\end{figure*}

Considering that adjacent pixel values vary smoothly, total variation (TV) \cite{total_variation_review} is devised to characterize this phenomenon. It is defined as ${\rm TV}(\bm{\mathcal{X}})=\|\nabla_{h}\bm{\mathcal{X}}\|_{1}+\|\nabla_{v}\bm{\mathcal{X}}\|_{1}$, where $\nabla_{h}$ and $\nabla_{v}$ denote the gradient operators along the horizontal and vertical axes, respectively. For brevity, TV can be compactly expressed as    
\begin{equation}  
	{\rm TV}(\bm{\mathcal{X}}) = \|\nabla \bm{\mathcal{X}}\|_{1},  
\end{equation}  
where $\nabla=(\nabla_{h},\nabla_{v})$. 
TV, with its widespread visual applications \cite{TV_deblurring,TV_inpainting,he2015total}, has been one of the most classic image gradient regularizers. Several noteworthy directions for enhancing TV are summarized as follows.

(i) Non-convex Total Variation: According to the definition of TV, it is the $\ell_1$-norm of gradients, implicitly assuming that gradients follow the Laplacian distribution. However, histograms of gradients in natural images reveal that this assumption is too rough to be always correct, and instead, the hyper-Laplacian distribution should be more properly considered \cite{NonConvexTV}. Therefore, the $\ell_p$-norm ($0<p<1$) of gradients is a more appropriate choice, leading to a non-convex TV formulation expressed as ${\rm TV}_{\ell_{p}}(\bm{\mathcal{X}}) = \|\nabla \bm{\mathcal{X}}\|_{p}^{p}$. Moreover, numerous studies have also validated the superior performance of other non-convex norm-based TV regularizers \cite{NonConvexTV2015,Nonconvex_Hybrid_Total_Variation}.

(ii) Directional Total Variation: Due to the inclusion of only horizontal and vertical neighbor information considered in conventional TV norms, edges in other directions are inevitably weakened, resulting in suboptimal and blurry image recovery in specific image processing tasks, such as rain streak removal \cite{LDCKTV_IEEESPL_2021} and seismic noise attenuation \cite{High_Order_Directional_Total_Variation}. To address this issue, DTV \cite{DTV_IEEESPL_2012} can be employed, defined as ${\rm DTV}(\bm{\mathcal{X}}) = \|\bm{k}_{\theta} \otimes \bm{\mathcal{X}}\|_{1}$, where $\bm{k}_{\theta}$ is a convolutional kernel modeling directional gradients and $\otimes$ denotes the convolution operator.

(iii) 3D Total Variation: When focusing on the smoothness along the channel axis in 3D images, an additional term can be incorporated. Specifically, ${{\rm TV}}_{\rm 3D}(\bm{\mathcal{X}})={{\rm TV}}(\bm{\mathcal{X}})+\|\nabla_{t}\bm{\mathcal{X}}\|_{1}$, where $\nabla_{t}$ denotes the gradient operator along the channel axis. For better characterizing sparsity of gradient maps along three axes, the enhanced 3D TV applies the sparsity measure to, instead of the gradient maps themselves, their subspace basis maps along all channels \cite{E3DTV}. More recently, the correlated TV (CTV) was proposed to better model channel smoothness by imposing a nuclear norm on spatial gradient maps, i.e., ${{\rm CTV}}(\bm{\mathcal{X}}) = \|\nabla \bm{\mathcal{X}}\|_{*}$ \cite{peng2022exact}. Tensor-CTV (t-CTV) extends this idea by using a tensor nuclear norm \cite{wang2023guaranteed}.

(iv) High-order Total Variation (HOTV): Traditional TV imposes norms on first-order gradients, leading to undesirable ``stair effects'' (i.e., resulting in piecewise constant images) \cite{TV_HOTV}. Second-order TV, defined as $\|\nabla^{2}\bm{\mathcal{X}}\|_{1}$, maintains the good properties of traditional TV near true edges and penalizes the formation of incorrect edges in areas that ought to maintain smoothness, potentially mitigating stair effects \cite{Second_order_TV_SIAMJSC_2000}. Furthermore, high-order TV preserves edges in the resultant deformation and is more robust to outlier noise \cite{Arbitrary_order_TV_PR_2023}.

(v) Total Generalized Variation (TGV): TGV, a state-of-the-art improvement over TV, also aims to mitigate stair effects by leveraging high-order gradients. Generally, a $k$-order TGV involves gradients of orders $i=1,2,\cdots,k$, and its kernel consists of polynomials with degrees less than $k$. To be more specific, 2-order TGV is defined as follows \cite{TGV}:
\begin{equation}\label{eq:TGV}
	{\rm TGV}^{2}_{\bm{\lambda}}(\bm{\mathcal{X}})=\min_{\bm{\mathcal{W}}} \lambda_{1}\|\nabla \bm{\mathcal{X}} - \bm{\mathcal{W}}\|_{1} + \lambda_{0}\|\mathcal{E}\bm{\mathcal{W}}\|_{1},
\end{equation}
where $\mathcal{E}=0.5(\nabla+\nabla^{\rm T})$ is the symmetrized gradient operator, and $\bm{\lambda}=(\lambda_0,\lambda_1)$ represents the hyper-parameters. It has been reported that 2-order TGV prefers piecewise linear reconstructions rather than piecewise constant ones, thereby preventing stair effects while still retains the edges that should be presented in the restored image \cite{TGV_piecewise_constant_func}.

(vi) Total Deep Variation (TDV): The aforementioned variants of TV can essentially be formulated as
\begin{equation}\label{eq:g_TV}
	g_{\rm TV}(\bm{\mathcal{X}})=f(\bm{k}\otimes\bm{\mathcal{X}}),
\end{equation}
where $f(\cdot)$ denotes a certain norm imposed on the transformed image $\bm{k}\otimes\bm{\mathcal{X}}$. However, both $f(\cdot)$ and $\bm{k}$ are fixed and required to be manually pre-specified. TDV makes Eq. (\ref{eq:g_TV}) learnable and is given by \cite{TDV,TDV2}
\begin{equation}\label{eq:TDV}
	{\rm TDV}(\bm{\mathcal{X}}) =\sum_{(i,j)}\bm{\mathcal{F}}_{ij}= \sum_{(i,j)} [\bm{w}\otimes {\rm CUNet}(\bm{k}\otimes \bm{\mathcal{X}})]_{ij}
\end{equation}
where $(i,j)$ indexes the pixel coordinates, $\bm{k}$ is a zero-mean convolutional kernel, ${\rm CUNet}$ represents the cascade of three U-shaped network, and $\bm{w}$ is a $1\times1$ convolutional kernel imposed on the output of CUNet to generate a one-channel feature map $\bm{\mathcal{F}}_{ij}$. Parameters $\bm{k}$, $\bm{w}$ and the weights in CUNet are learnable. From the definition of TDV, it is clear that TDV is more flexible than $g_{\rm TV}$, since CUNet could explore multi-scale prior and representative features.

A brief overview of the recent advancements in gradient prior modeling is provided above. Subsequently, we will delve into discussing their inherent limitations.

(i) The first four kinds of TV variants primarily aim to better characterize the specific sparsity prior of gradient maps, which is equivalent to local smoothness. TV and its variants measure the distance between gradient maps and zeros. While these methods yield commendable results for smooth regions when minimizing the TV regularizer, it is crucial to note that they also tend to weaken edges and details whose gradients should not be zeros. In essence, they consistently underestimate the intrinsic complex priors besides sparsity underlying gradient maps of general real-world images.

(ii) Regarding TDV, it is a black-box regularizer, the behavior of which is not easily comprehensible. The fundamental question of why minimizing the sum of $\bm{\mathcal{F}}_{ij}$ leads to successful image restoration remains unanswered. Consequently, it remains unclear as to what type of prior is being modeled by TDV.

(iii) Furthermore, the performance enhancement of TDV stems from training on a large-scale paired dataset for a specific task. Consequently, TDV could not be readily employed as an off-the-shelf regularizer in a plug-and-play manner, making it challenging to directly apply the learned TDV when the task or data changes. Even when they are unchanged, the generalization capability of TDV remains obscure. In both scenarios, retraining TDV becomes necessary. More critically, TDV cannot be trained when paired datasets are unavailable or difficult to collect, such as in the case of electron microscope images.

Based on these analyses, the ideal gradient regularizer should possess the following attributes: good performance, plug-and-play capabilities, and less requirement for training on large-scale paired datasets. In response to this burgeoning need, we present a neural gradient regularizer (NGR) in this paper. As depicted in Fig. \ref{fig:toc}, NGR seeks to recover the gradient map from a degraded image using a neural network, in a zero-shot learning fashion. This method does not rely on the assumption of sparse gradients, thereby distinguishing it from the existing techniques under Eq. (\ref{eq:g_TV}). In comparison to previous gradient regularizers, NGR exhibits an evident performance enhancement. We also apply NGR to multiple image processing tasks, and experimental results validate the widespread superiority of NGR over existing methods.

The remainder of this article is structured as follows: Section \ref{sec:Methods} presents the NGR. Section \ref{sec:experiments} reports the outcomes of numerical experiments. Finally, Section \ref{sec:Conclusion} summarizes the findings of this study.

\section{Methods}\label{sec:Methods}
\subsection{Neural gradient regularizer}
The success of TV can be attributed to the fact that most entries of the gradient maps approach zero. However, the downside of minimizing TV is that it underestimates the gradient of edges. Intuitively, incorporating more comprehensive information from gradient maps could potentially lead to better results. For instance, CTV and t-CTV, which simultaneously model local smoothness and low-rankness, aids in recovering finer details beyond what TV can achieve. Let us consider an ideal scenario where we have access to all the information of the ground-truth gradient maps, denoted as $\bm{\mathcal{G}}=(\bm{\mathcal{G}}_{h}, \bm{\mathcal{G}}_{v}, \bm{\mathcal{G}}_{t})$. It is reasonable to constrain the gradient map of the restored image to be equal to the ground-truth gradient map. Consequently, the model can be formulated as  
$$  
\min_{\bm{\mathcal{X}}} L(\bm{\mathcal{X}},\bm{\mathcal{Y}}), \quad \text{s.t. }  \nabla\bm{\mathcal{X}}=\bm{\mathcal{G}},  
$$  
where $\nabla=(\nabla_{h}, \nabla_{v}, \nabla_{t})$ represents the gradient operators along the three axes.

In practice, the ground-truth gradient map $\bm{\mathcal{G}}$ is often inaccessible, so we need to estimate it. Similar to the working mechanism of TDV, one plausible solution is to train a neural network that maps an observed corrupted image $\bm{\mathcal{Y}}$ to its clean gradient map $\bm{\mathcal{G}}$ for a specific task. However, it should be noted that this approach is effective only when the task and image type remain fixed. For instance, a network trained for the denoising task may not perform well for the inpainting task. Moreover, if the network is trained on RGB images, the network cannot be compatible with the test data consists of hyperspectral images (HSIs) or multispectral images (MSIs) with different channel numbers. In summary, this solution falls short of meeting the plug-and-play requirement.

To address this issue, we propose an estimation approach for gradient maps in the fashion of zero-shot learning. Partially inspired by deep image prior (DIP), the proposed neural gradient regularizer (NGR) encourages an untrained neural network  $f_{\Theta}(\cdot)=(f_{\Theta_{h}}(\cdot), f_{\Theta_{v}}(\cdot), f_{\Theta_{t}}(\cdot))$ to predict the gradient maps along three axes of a high-quality image. This can be expressed as 
\begin{equation}\label{eq:NGR_constraint}
	\min_{\bm{\mathcal{X}}, \Theta} L(\bm{\mathcal{X}},\bm{\mathcal{Y}}),  \quad {\rm s.t.} \nabla_{i}\bm{\mathcal{X}}=f_{\Theta_{i}}(\bm{\mathcal{G}}^{(0)}), i\in\{h,v,t\},
\end{equation}
where $\Theta=(\Theta_{h},\Theta_{v},\Theta_{t})$ denotes the collection of all learnable parameters in $f_{\Theta}(\cdot)$, $\bm{\mathcal{G}}^{(0)}$ is a randomly sampled variable, and $\nabla_{i}\bm{\mathcal{X}}$ represents the gradient of $\bm{\mathcal{X}}$ along its $i$th axis ($i\in\{h,v,t\}$). To solve this optimization problem, we rewrite the constraint equation as a term in the objective function:
\begin{equation}\label{eq:NGR}
	\min_{\bm{\mathcal{X}}, \Theta} L(\bm{\mathcal{X}},\bm{\mathcal{Y}}) + \sum_{i\in\{h,v,t\}} \frac{\lambda_{i}}{2} \| \nabla_{i}\bm{\mathcal{X}}-f_{\Theta_{i}}(\bm{\mathcal{G}}^{(0)})\|_{2}^2,
\end{equation}
where $\lambda_{i}$ is a hyper-parameter that controls the penalty strength for each axis ($i\in\{h,v,t\}$).

Eq. (\ref{eq:NGR}) does not impose any manually pre-defined assumptions on the image processing tasks or image types. The solution to Eq. (\ref{eq:NGR}) presented in the following section will illustrate that NGR is not reliant on large-scale dataset training. Consequently, NGR could be conveniently used as a flexible plug-and-play gradient modeling tool.

\textbf{Remark:} Readers familiar with DIP \cite{ulyanov2018deep,DIP_IJCV} might observe similarities between DIP and our proposed NGR, as both utilize untrained neural networks to embody prior knowledge and function in a zero-shot learning paradigm. However, several critical distinctions exist. Notably, DIP models the image prior rather than the gradient map prior. Moreover, the spectral bias theory \cite{spectral_bias_dip,SB_DIP} reveals that DIP learns low-frequency information more rapidly than high-frequency information, resulting in its relative inadequacy in capturing details and textures. Consequently, the incoherent mechanism of DIP hinders its capability to reconstruct details and textures. Conversely, considering that the observed image itself contains useful low-frequency structures, NGR is designed to better focus on recovering high-frequency information, thereby obviating the need to model low-frequency information, which could potentially facilitate the restoration quality of image details and textures. As will be demonstrated in the subsequent subsection, during training, NGR manipulates Eq. (\ref{eq:solve_x}) to merge the low-frequency information provided by the observed image and the high-frequency information recovered by the network. Most significantly, the results exhibited in Section \ref{sec:experiments} substantiate that NGR surpasses DIP by a gain of 3dB in terms of PSNR for the  inpainting task across several RGB and video datasets. This empirical evidence finely validates the superiority of NGR over DIP.

\subsection{Solution to NGR regularized image inpainting}
NGR is potentially applicable to a wide range of image processing tasks. As a case in point, we demonstrate how to address Eq. (\ref{eq:NGR}) in the context of image inpainting, which involves restoring missing entries in observed pixels, with their indices denoted as $\Omega$. Mathematically, image inpainting entails estimating the observed image, formulated as a tensor $\bm{\mathcal{X}}$, subject to the constraint $\mathcal{P}_{\Omega}(\bm{\mathcal{X}})=\mathcal{P}_{\Omega}(\bm{\mathcal{Y}})$, where $\mathcal{P}(\cdot)$ denotes the projection operator. In order to decouple $\mathcal{P}_{\Omega}(\cdot)$ and $\bm{\mathcal{X}}$, an auxiliary variable $\bm{\mathcal{K}}$ is introduced to compensate for unobserved entries, leading to the revised constraint $\mathcal{P}_{\Omega}(\bm{\mathcal{X}}+\bm{\mathcal{K}})=\mathcal{P}_{\Omega}(\bm{\mathcal{Y}})$, thereby facilitating the solution of Eq. (\ref{eq:NGR}). By combining NGR, an image inpainting problem can be addressed using the following optimization problem:  
\begin{equation}\label{eq:inpainting_NGR}  
	\left\lbrace  
	\begin{aligned}  
		\min_{\bm{\mathcal{X}}, \bm{\mathcal{K}}, \Theta} \quad & \delta_{\bm{\mathcal{K}}, \Omega} + \sum_{i\in\{h,v,t\}} \frac{\lambda_{i}}{2} \| \nabla_{i}\bm{\mathcal{X}}-f_{\Theta_{i}}(\bm{\mathcal{G}}^{(0)})\|_{2}^2, \\  
		{\rm s.t.} \quad  & \mathcal{P}_{\Omega}(\bm{\mathcal{X}}+\bm{\mathcal{K}})=\mathcal{P}_{\Omega}(\bm{\mathcal{Y}}),  
	\end{aligned}  
	\right.  
\end{equation}  
where $\delta_{\bm{\mathcal{K}}, \Omega}$ is an indicator function that constrains $\bm{\mathcal{K}}$ to be in the complement of $\Omega$, defined as:  
\begin{equation}  
	\delta_{\bm{\mathcal{K}}, \Omega}=  
	\begin{cases}  
		0, & \mathcal{P}_{\Omega}(\bm{\mathcal{K}})=0, \\  
		+\infty, & \text { otherwise }.  
	\end{cases}  
\end{equation}  
This constraint helps the elements of X exactly similar to those of Y in non-missing components, but flexibly valued in those missing parts. To solve this optimization problem, the Alternating Direction Method of Multipliers (ADMM) can be readily employed. The original problem is recast as a minimization of the augmented Lagrangian function:  
\begin{equation}\label{eq:Lagrangian func}  
	\begin{aligned}  
		\min_{\bm{\mathcal{X}}, \bm{\mathcal{K}}, \bm{\Lambda}, \Theta}  & \left\lbrace \delta_{\bm{\mathcal{K}}, \Omega} + \sum_{i\in\{h,v,t\}} \frac{\lambda_{i}}{2} \| \nabla_{i}\bm{\mathcal{X}}-f_{\Theta_{i}}(\bm{\mathcal{G}}^{(0)})\|_{2}^2 \right. \\  
		& \left. + \frac{\mu}{2} \|\mathcal{P}_{\Omega}(\bm{\mathcal{Y}})-\bm{\mathcal{X}}-\bm{\mathcal{K}}+\frac{\bm{\Lambda}}{\mu}\|_{2}^{2} \right\rbrace,  
	\end{aligned}  
\end{equation}  
where $\bm{\Lambda}$ is the Lagrangian multiplier and $\mu$ is a hyperparameter. The unknown variables are optimized iteratively.

(1) Updating $\Theta$: By fixing the variables other than $\Theta$, we derive the objective function for $\Theta$. The optimization problem can be formulated as follows:
\begin{equation}\label{eq:solve_theta}  
	\min_{\Theta} \sum_{i\in\{h,v,t\}} \frac{\lambda_{i}}{2} \|\nabla_{i}\bm{\mathcal{X}}-f_{\Theta_{i}}(\bm{\mathcal{G}}^{(0)})\|_{2}^{2}.  
\end{equation}
Given that $f_{\Theta_{i}}(\cdot)$ is a neural network composed of nonlinear operators, we can easily employ the Adam optimizer to solve this optimization problem.

(2) Updating $\bm{\mathcal{K}}$: The optimization of $\bm{\mathcal{K}}$ is straightforward, and we provide the expression as follows:
\begin{equation}\label{eq:solve_K}  
	\bm{\mathcal{K}} = \mathcal{P}_{\Omega}(\bm{\mathcal{Y}})-\bm{\mathcal{X}}+\frac{\Lambda}{\mu}, \quad \text{where} \quad \mathcal{P}_{\Omega}(\bm{\mathcal{K}})=0.  
\end{equation}

(3) Updating $\bm{\mathcal{X}}$: We update $\bm{\mathcal{X}}$ by solving the following optimization problem:
\begin{equation}
	\begin{aligned}
		\min_{\bm{\mathcal{X}}} \left\lbrace \sum_{i\in\{h,v,t\}} \frac{\lambda_{i}}{2} \| \nabla_{i}\bm{\mathcal{X}}-f_{\Theta_{i}}(\bm{\mathcal{G}}^{(0)})\|_{2}^2 + \right.\\
		\left.
		\frac{\mu}{2} \|\mathcal{P}_{\Omega}(\bm{\mathcal{Y}})-\bm{\mathcal{X}}-\bm{\mathcal{K}}+\frac{\bm{\Lambda}}{\mu}\|_{2}^{2} \right\rbrace,
	\end{aligned}
\end{equation}
Taking the derivative of this objective function with respect to $\bm{\mathcal{X}}$ and setting it to zero lead to the following equation:
\begin{equation}\label{eq:derivative_x}
	\sum_{i\in\{h,v,t\}} \lambda_{i}  \nabla_{i}^{\rm T}(\nabla_{i}\bm{\mathcal{X}}-f_{\Theta_{i}}(\bm{\mathcal{G}}^{(0)})) = \mu (\mathcal{P}_{\Omega}(\bm{\mathcal{Y}})-\bm{\mathcal{X}}-\bm{\mathcal{K}})+\bm{\Lambda}.
\end{equation}
After simple calculations, we obtain a linear system:
\begin{equation}\label{eq:linear_system}  
	\begin{aligned}
		&\left(\mu+\sum_{i\in\{h,v,t\}} \lambda_{i}  \nabla_{i}^{\rm T}\nabla_{i}\right)\bm{\mathcal{X}} = \\
		& \sum_{i\in\{h,v,t\}} \lambda_{i}\nabla_{i}^{\rm T}f_{\Theta_{i}}(\bm{\mathcal{G}}^{(0)}) + \mu (\mathcal{P}_{\Omega}(\bm{\mathcal{Y}})-\bm{\mathcal{K}})+\bm{\Lambda}.
	\end{aligned}
\end{equation}
Here, $\nabla_{i}^{\rm T}$ denotes the transposed operator of $\nabla_{i}$, and we denote the right-hand side of Eq. (\ref{eq:linear_system}) as $\bm{\mathcal{R}}$. The closed-form solution can be deduced using the following expression:
\begin{equation}\label{eq:solve_x}
	\bm{\mathcal{X}}=\mathcal{F}^{-1}\left(\frac{\mathcal{F}\left(\bm{\mathcal{R}}\right)}{\mu \mathbf{1}+\sum_{i\in\{h,v,t\}}\left|\mathcal{F}\left(\nabla_i\right)\right|^2}\right),
\end{equation}
In Eq. (\ref{eq:solve_x}), $|\cdot|^2$ represents the element-wise square operator, and $\mathcal{F}\left(\cdot\right)$ and $\mathcal{F}\left(\cdot\right)^{-1}$ denote the Fourier transform and its inverse, respectively.

(4) Update $\bm{\Lambda}$: According to general ADMM principles, the multipliers are updated by 
\begin{equation}\label{eq:solve_lambda}
	\bm{\Lambda} = \bm{\Lambda}+\mu(\mathcal{P}_{\Omega}(\bm{\mathcal{Y}})-\bm{\mathcal{X}}-\bm{\mathcal{K}}).
\end{equation}

Algorithm \ref{alg:ngr_inpainting} outlines the overall workflow. Taking the observation image $\bm{\mathcal{Y}}$, observation set $\Omega$, and hyper-parameters $\lambda_{i} (i=1,2,3)$ as input, the algorithm yields the restored image $\bm{\mathcal{X}}$. At step 1, $\bm{\mathcal{G}}^{(0)}$ is initialized by sampling from a uniform distribution. Notably, this algorithm dispenses with the need for training the network on large-scale datasets, thereby enabling NGR to operate in a zero-shot learning paradigm. We also apply NGR to image denoising problem, and please refer to supplementary document for details. 

\begin{algorithm}[H]
	\caption{NGR regularized image inpainting}
	\label{alg:ngr_inpainting}
	\begin{algorithmic}[1]
		\REQUIRE  
		$\bm{\mathcal{Y}}$, $\Omega$, $\lambda_{i}(i=1,2,3)$
		\ENSURE  
		$\bm{\mathcal{X}}$.
		\STATE Initialize $\bm{\mathcal{G}}^{(0)}$.
		\WHILE {not converged}
		\STATE Update $\Theta$ by applying Adam optimizer to Eq. (\ref{eq:solve_theta});
		\STATE Update $\bm{\mathcal{K}}$ by Eq. (\ref{eq:solve_K});
		\STATE Update $\bm{\mathcal{X}}$ by Eq. (\ref{eq:solve_x});
		\STATE Update $\bm{\Lambda}$ by Eq. (\ref{eq:solve_lambda}).
		\ENDWHILE
	\end{algorithmic}
\end{algorithm}

\begin{figure*}[t]
	\centering
	\subfigure[\centering\scriptsize Observation]    {\includegraphics[width=0.15\linewidth]{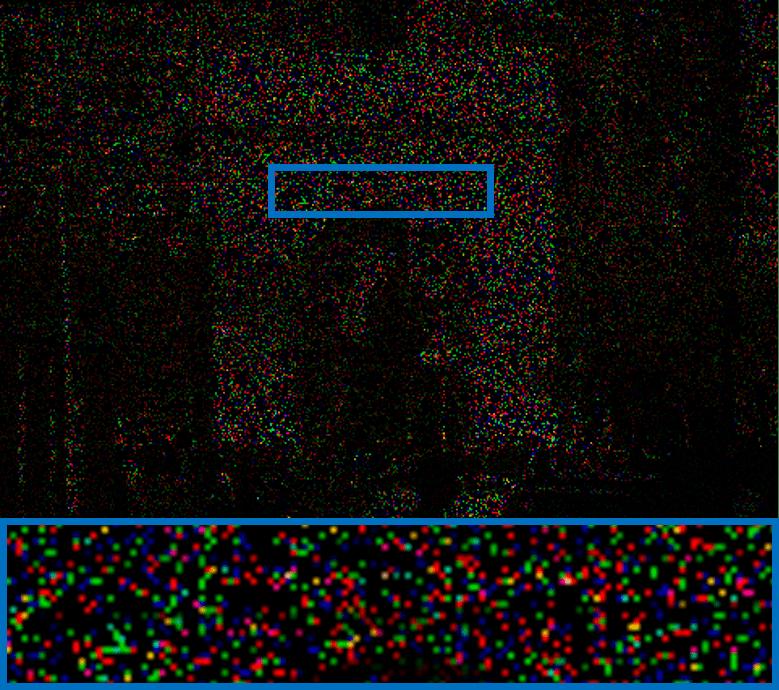} }
	\subfigure[\centering\scriptsize HaLRTC(17.40)]  {\includegraphics[width=0.15\linewidth]{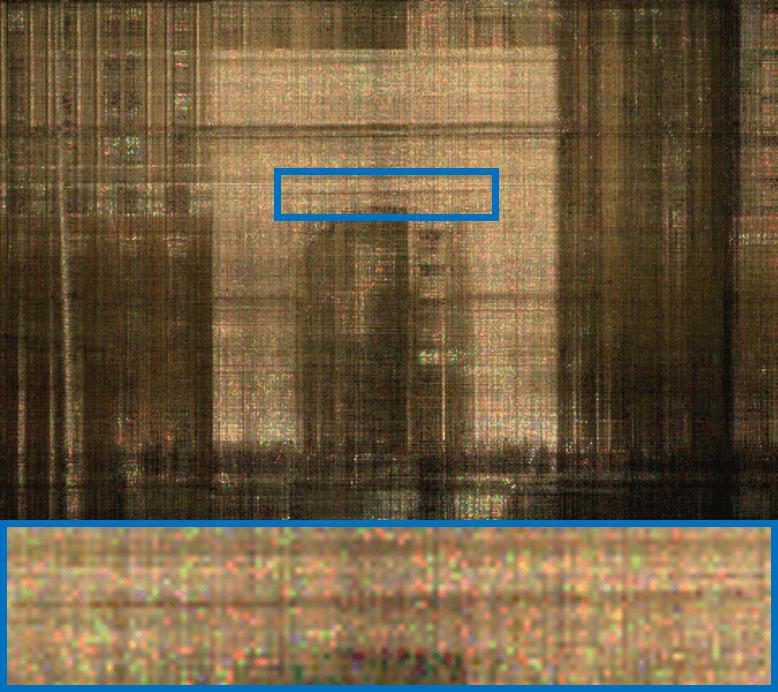} }
	\subfigure[\centering\scriptsize SPC-TV(22.04)]  {\includegraphics[width=0.15\linewidth]{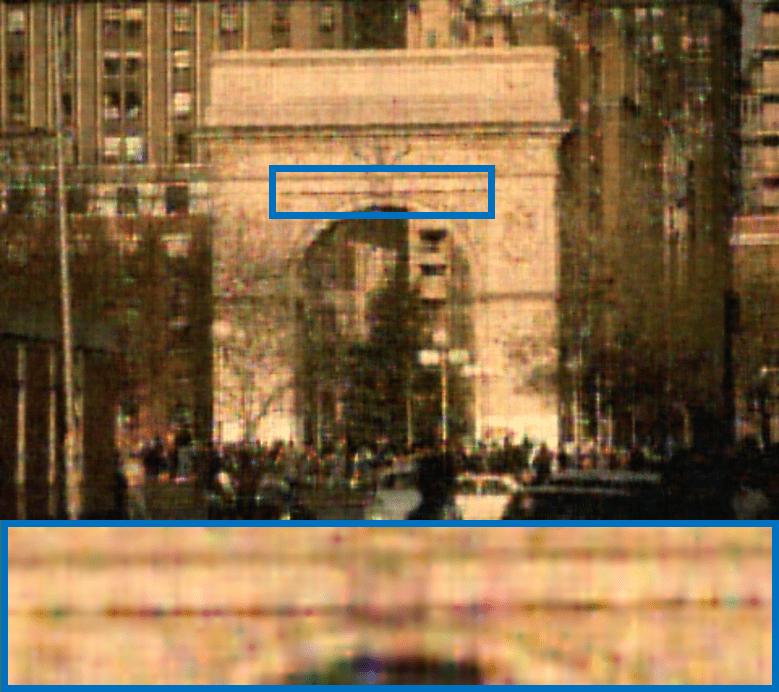} }
	\subfigure[\centering\scriptsize TNN-FFT(19.72)] {\includegraphics[width=0.15\linewidth]{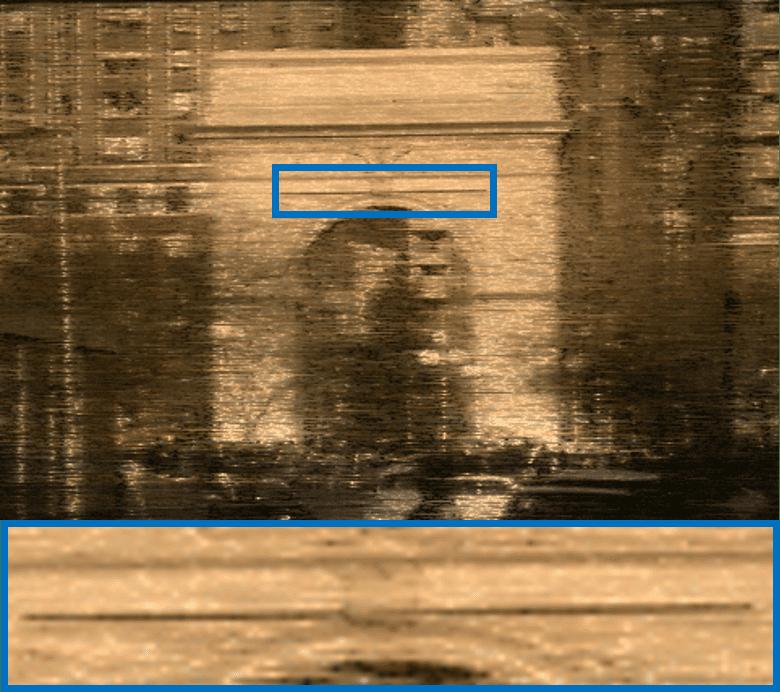} }
	\subfigure[\centering\scriptsize TNN-DCT(19.15)] {\includegraphics[width=0.15\linewidth]{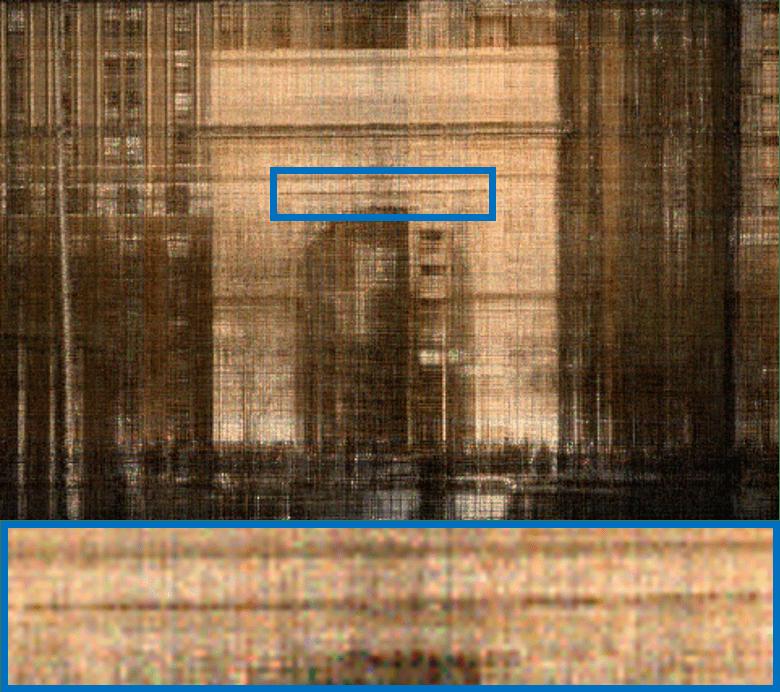} }
	\subfigure[\centering\scriptsize LRTC-TV(16.60)] {\includegraphics[width=0.15\linewidth]{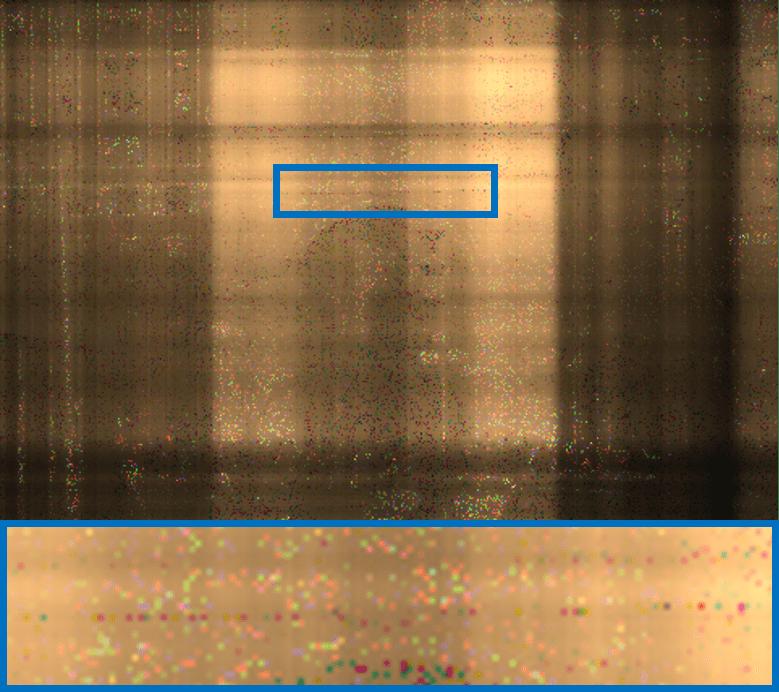} }
	\subfigure[\centering\scriptsize t-CTV(22.56)]   {\includegraphics[width=0.15\linewidth]{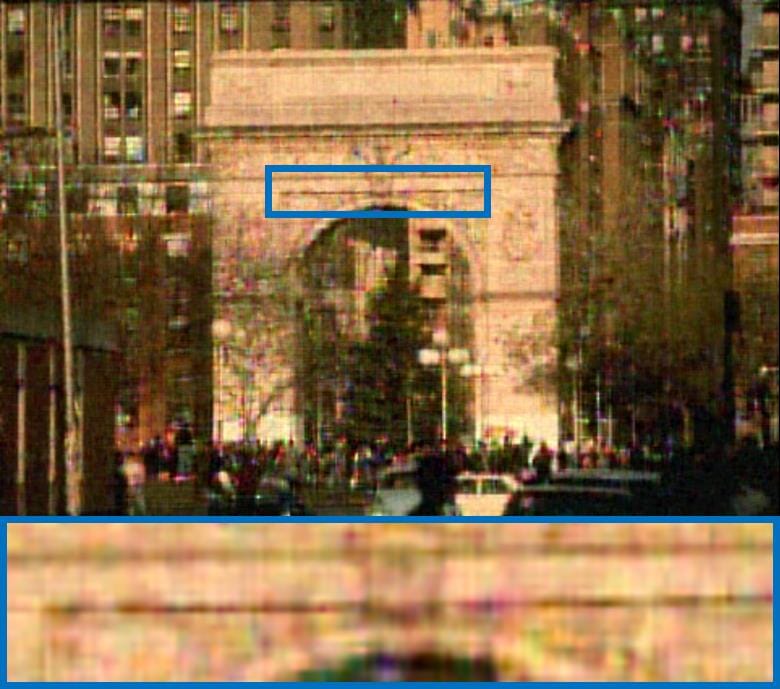} }
	\subfigure[\centering\scriptsize DIP(23.07)]     {\includegraphics[width=0.15\linewidth]{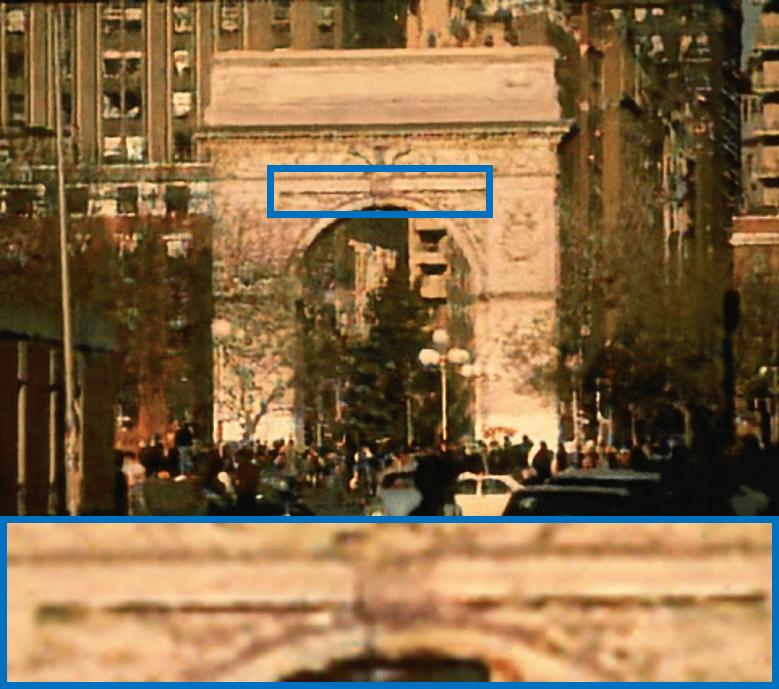} }
	\subfigure[\centering\scriptsize S2DIP(23.29)]   {\includegraphics[width=0.15\linewidth]{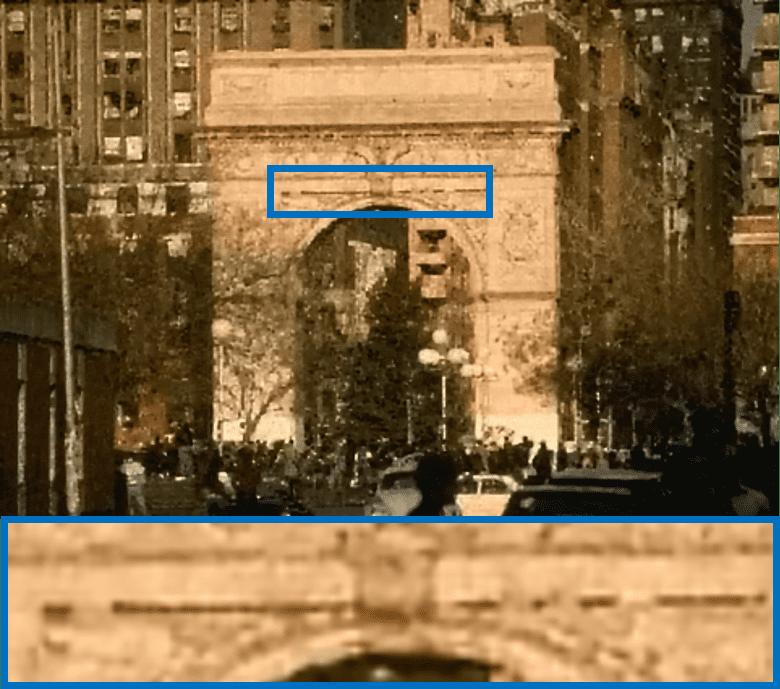} }
	\subfigure[\centering\scriptsize NGR(23.90)]     {\includegraphics[width=0.15\linewidth]{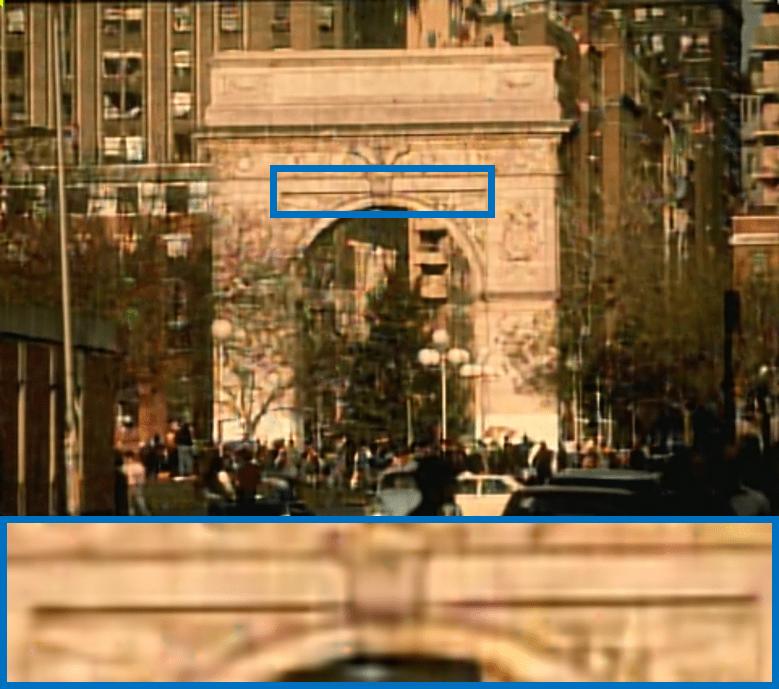} }
	\subfigure[\centering\scriptsize GT]             {\includegraphics[width=0.15\linewidth]{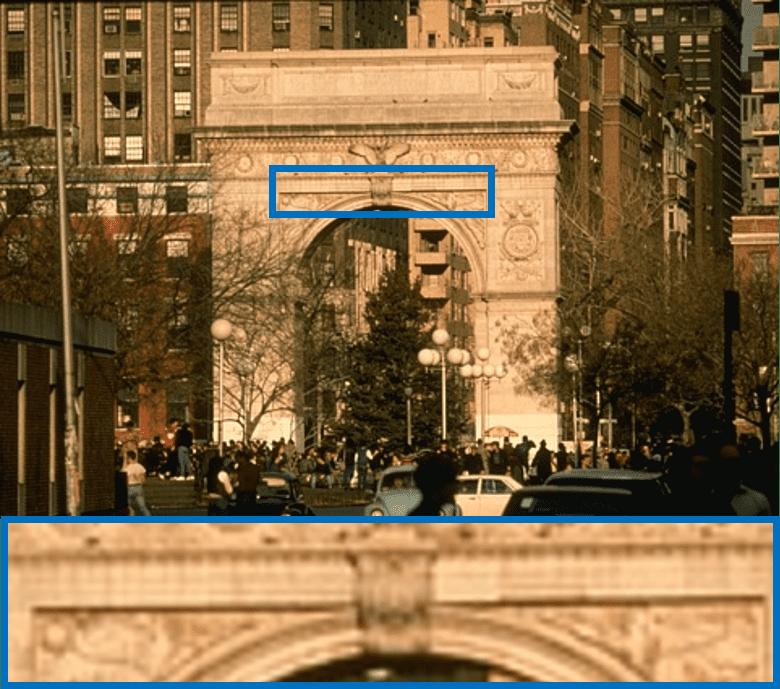} }
	\caption{GT, observed rgb image, inpainting visual results and corresponding PSNR values by methods for comparison on "148089" (SR = 10\%) from BSDS100 dataset. \label{RGB_inpainting fig}}
\end{figure*} 
\begin{figure*}[t]
	\centering
	\subfigure[\centering\scriptsize Observation]    {\includegraphics[width=0.15\linewidth]{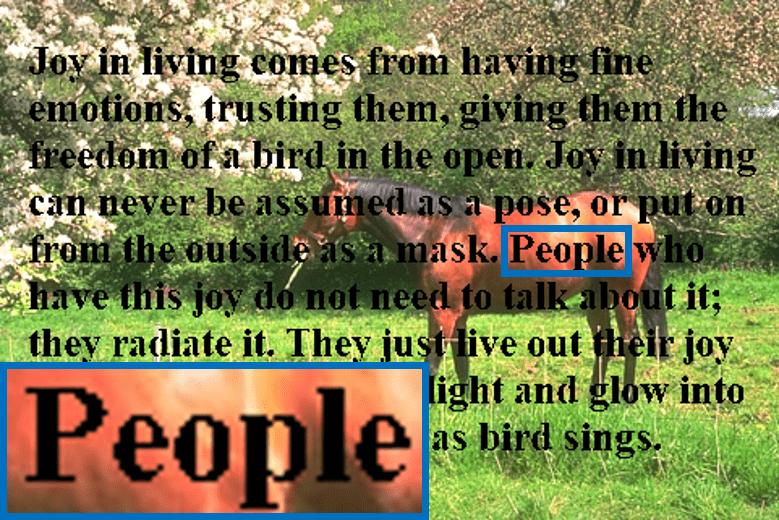} }
	\subfigure[\centering\scriptsize HaLRTC(25.90)]  {\includegraphics[width=0.15\linewidth]{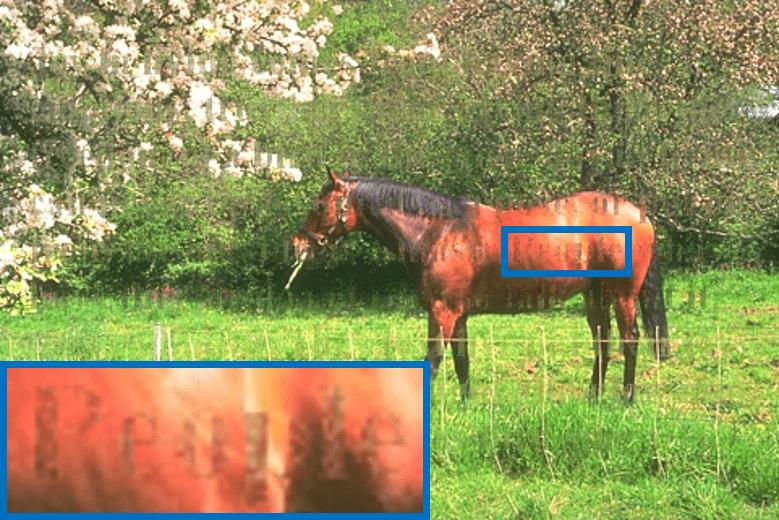} }
	\subfigure[\centering\scriptsize SPC-TV(25.98)]  {\includegraphics[width=0.15\linewidth]{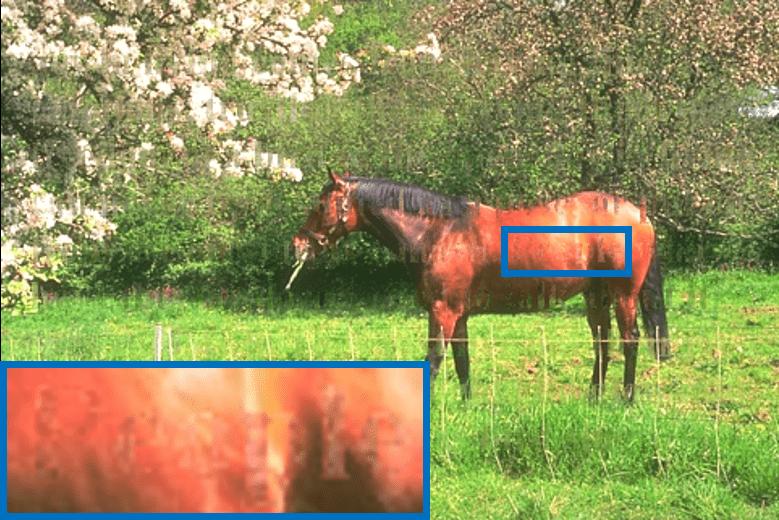} }
	\subfigure[\centering\scriptsize TNN-FFT(25.63)] {\includegraphics[width=0.15\linewidth]{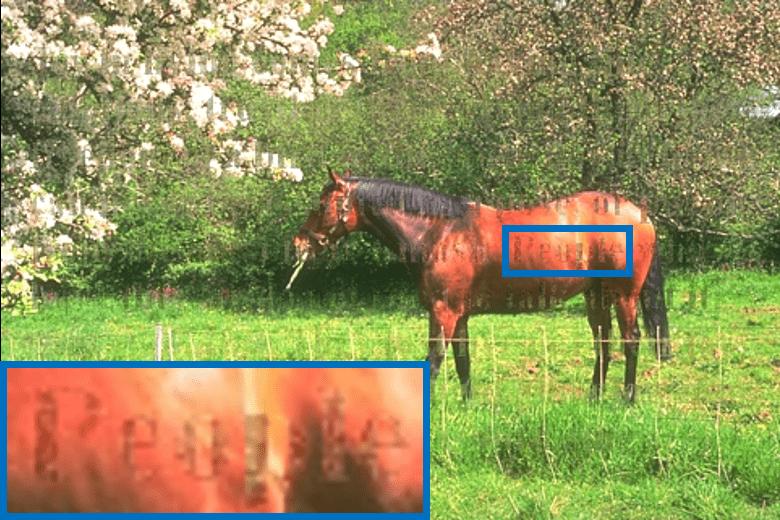} }
	\subfigure[\centering\scriptsize TNN-DCT(25.63)] {\includegraphics[width=0.15\linewidth]{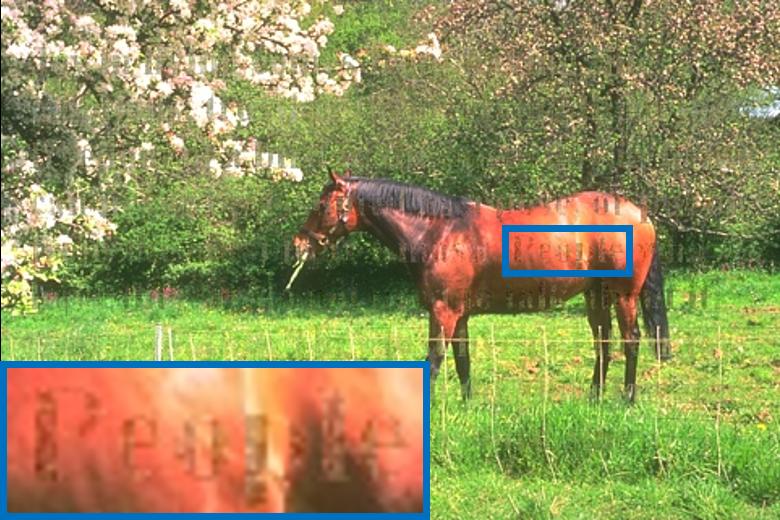} }
	\subfigure[\centering\scriptsize LRTC-TV(24.64)] {\includegraphics[width=0.15\linewidth]{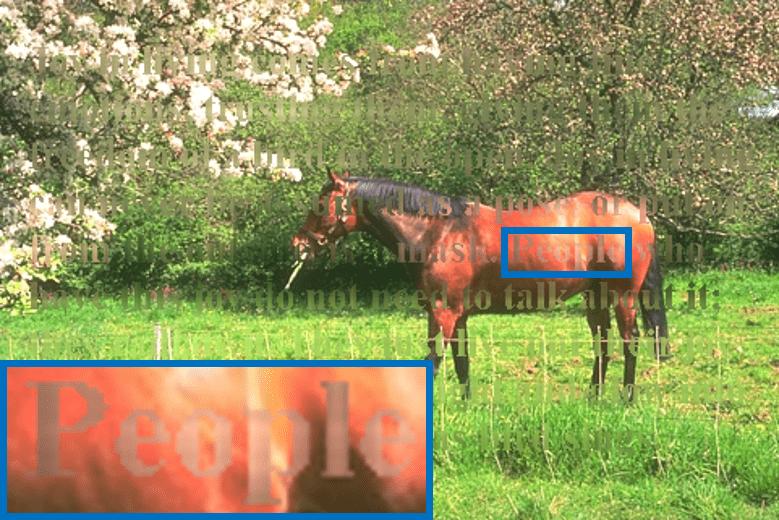} }
	\subfigure[\centering\scriptsize t-CTV(27.16)]   {\includegraphics[width=0.15\linewidth]{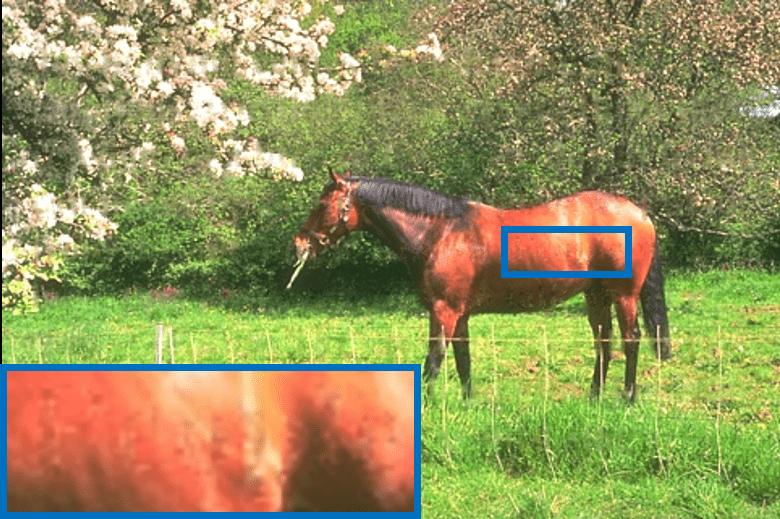} }
	\subfigure[\centering\scriptsize DIP(26.95)]     {\includegraphics[width=0.15\linewidth]{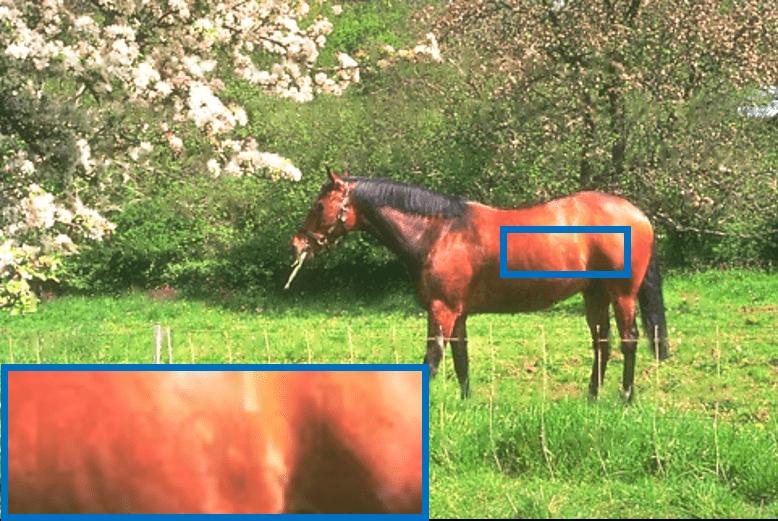} }
	\subfigure[\centering\scriptsize S2DIP(27.52)]   {\includegraphics[width=0.15\linewidth]{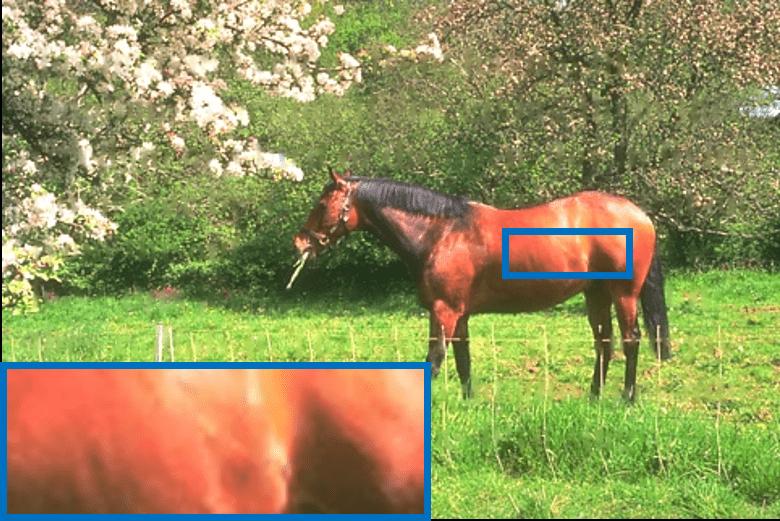} }
	\subfigure[\centering\scriptsize NGR(27.61)]     {\includegraphics[width=0.15\linewidth]{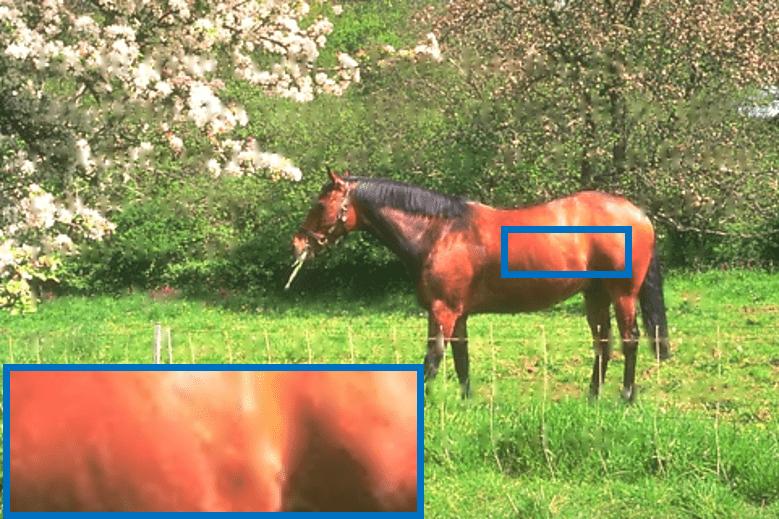} }
	\subfigure[\centering\scriptsize GT]             {\includegraphics[width=0.15\linewidth]{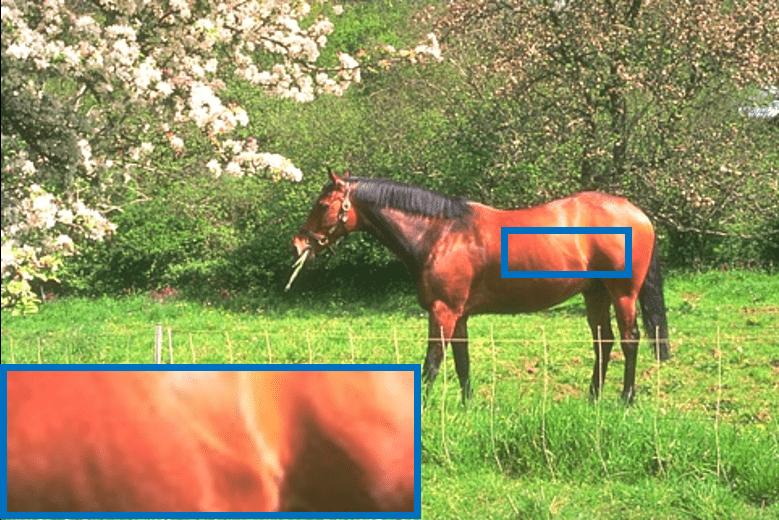} }
	\caption{GT, observed rgb image, text inpainting visual results and corresponding PSNR values by methods for comparison on "291000" from BSDS100 dataset. \label{text_inpainting fig}}
\end{figure*}

\subsection{Connection to previous works}
Connection to TGV: By introducing an auxiliary variable $\bm{\mathcal{W}}$, Eq. (\ref{eq:NGR}) can be equivalently reformulated as  
\begin{equation}  
	\begin{aligned}  
		\min_{\bm{\mathcal{X}}, \Theta} L(\bm{\mathcal{X}},\bm{\mathcal{Y}}) + \sum_{i\in\{h,v,t\}} \frac{\lambda_{i}}{2} \|\nabla_{i}\bm{\mathcal{X}}-\bm{\mathcal{W}}_i\|_{2}^2, \\  
		{\rm s.t. } \bm{\mathcal{W}}_i = f_{\Theta_{i}}(\bm{\mathcal{Z}}).  
	\end{aligned}  
\end{equation}  
Conversely, the TGV regularized problem can be written as  
\begin{equation}  
	\min_{\bm{\mathcal{W}}} L(\bm{\mathcal{X}},\bm{\mathcal{Y}}) + \lambda_{1}\|\nabla \bm{\mathcal{X}} - \bm{\mathcal{W}}\|_{1} + \lambda_{0}\|\mathcal{E}\bm{\mathcal{W}}\|_{1}.  
\end{equation}  
It is observed that there exists a strong connection between TGV and NGR, as both minimize the distance between the gradient map and the auxiliary variable $\bm{\mathcal{W}}$. However, TGV and NGR impose distinct constraints on $\bm{\mathcal{W}}$. While TGV promotes the a manually specified sparsity prior of $\mathcal{E}\bm{\mathcal{W}}$, NGR restricts $\bm{\mathcal{W}}$ to be the output of a neural network that can automatically extract intrinsic prior structures underlying the gradient map, facilitating a possibly more flexible and complex representation capability on expressing such information.

Connection to TDV: Both TDV and NGR share the common goal of characterizing gradient priors using neural networks, namely, they are data-driven regularizers. Nonetheless, they exhibit several distinctions. Firstly, Kobler et al. \cite{TDV} employed a gradient flow to minimize the energy functional regularized by TDV, where the training process was described as a mean-field optimal control problem, necessitating the supervised learning of the network on large-scale datasets for a fixed and specific task. Consequently, TDV can be interpreted as a discriminative prior. However, it requires TDV to be retrained if the task/data varies. In contrast, NGR operates within a zero-shot learning paradigm, thus potentially serving as a versatile and plug-and-play regularizer. Secondly, from the definition displayed in Eq. (\ref{eq:TDV}), the TDV approach primarily focuses on minimizing the sum of a feature map obtained by a neural network, making it difficult to comprehend which prior is encoded in the regularizer and why minimizing this regularizer leads to good performance. In contrast, as validated by section \ref{sec:gradient_estimation_process} and Fig. \ref{analysis}, NGR gradually generates a refined gradient map of a high-quality image, thereby facilitating an understanding of the behavior of NGR.

\begin{table}[t]
	\caption{The averaged quantitive results of RGB image inpainting on BSDS100, Set5 and USC-SIPI. The best results are marked in bold.\label{RGB_inpainting table}}
	\centering
	\resizebox{\linewidth}{!}{
		\begin{tabular}{c|c|cc|cc|cc}
			\Xhline{1.3pt}
			~ & SR & \multicolumn{2}{c|}{50\%} & \multicolumn{2}{c|}{30\%} &  \multicolumn{2}{c}{10\%} \\ \hline
			~ & Metrics & PSNR & SSIM & PSNR & SSIM & PSNR & SSIM \\ \Xhline{0.8pt}
			\multirow{8}[0]{*}{\rotatebox{90}{BSDS100}} & HaLRTC & 29.44  & 0.929  & 25.15  & 0.834  & 19.21  & 0.592  \\ 
			~ & SPC-TV & 30.95  & 0.951  & 28.00  & 0.906  & 24.28  & 0.793  \\ 
			~ & TNN-FFT & 32.96  & 0.971  & 27.76  & 0.910  & 22.27  & 0.730  \\ 
			~ & TNN-DCT & 32.13  & 0.919  & 26.49  & 0.758  & 20.85  & 0.436  \\
			~ & LRTC-TV & 29.69  & 0.918  & 24.46  & 0.782  & 17.23  & 0.493  \\ 
			~ & t-CTV & 35.84  & \textbf{0.981}  & 30.08  & 0.934  & 24.84  & \textbf{0.814}  \\ 
			~ & DIP & 35.30  & 0.962   & 30.81  & 0.905 & 25.88  & 0.75  \\ 
			~ & S2DIP & 36.55 & 0.973 & \textbf{32.08} & \textbf{0.936} & 26.28 & 0.798 \\ 
			~ & \textbf{NGR} & \textbf{37.37}  & 0.977 & 31.87  & 0.929 & \textbf{26.39}  & 0.785  \\ \hline
			\multirow{8}[0]{*}{\rotatebox{90}{Set5\quad\ }} & HaLRTC & 30.99  & 0.884  & 26.14  & 0.734  & 19.68  & 0.412  \\ 
			~ & SPC-TV & 29.54  & 0.836  & 25.87  & 0.701  & 19.96  & 0.417  \\ 
			~ & TNN-FFT & 31.90  & 0.879  & 26.23  & 0.701  & 19.50  & 0.357  \\
			~ & TNN-DCT & 32.11  & 0.883  & 26.44  & 0.710  & 19.63  & 0.366  \\ 
			~ & LRTC-TV & 31.85  & 0.875  & 23.72  & 0.631  & 18.58  & 0.405  \\
			~ & t-CTV & 37.70  & 0.955  & 32.28  & 0.894  & 25.65  & 0.703  \\
			~ & DIP & 37.88  & 0.964  & 34.50  & 0.936  & 29.18  & 0.866  \\ 
			~ & S2DIP & 38.11  & 0.969  & 33.93  & 0.939  & 27.77  & 0.844  \\ 
			~ & \textbf{NGR} & \textbf{40.33}  & \textbf{0.974}  & \textbf{35.84}  & \textbf{0.951}  & \textbf{29.70}  & \textbf{0.881}  \\\hline 
			\multirow{8}[0]{*}{\rotatebox{90}{USC-SIPI\quad\ }} & HaLRTC & 29.27  & 0.872  & 25.83  & 0.742  & 20.71  & 0.477  \\ 
			~ & SPC-TV & 27.91  & 0.834  & 25.39  & 0.721  & 20.72  & 0.475  \\ 
			~ & TNN-FFT & 29.12  & 0.848  & 25.62  & 0.703  & 20.24  & 0.408  \\
			~ & TNN-DCT & 29.29  & 0.852  & 25.78  & 0.709  & 20.39  & 0.416  \\ 
			~ & LRTC-TV & 28.68  & 0.850  & 22.48  & 0.629  & 18.45  & 0.451  \\ 
			~ & t-CTV & 31.30  & 0.911  & 28.79  & 0.843  & 24.69  & 0.694  \\ 
			~ & DIP & 32.27  & 0.934  & 30.20  & 0.884  & 26.70  & 0.764  \\ 
			~ & S2DIP & 31.33  & 0.930  & 28.94  & 0.880  & 25.37  & 0.766  \\
			~ & \textbf{NGR} & \textbf{32.92}  & \textbf{0.940} & \textbf{30.70}  & \textbf{0.898}  & \textbf{27.18}  & \textbf{0.788} \\    \Xhline{1.3pt}
		\end{tabular}
	}
\end{table}

\begin{table}[t]
	\caption{The averaged quantitive results of videos inpainting on eight videos. The best results are marked in bold.\label{video_inpainting table}}
	\centering
	\begin{tabular}{c|cc|cc|cc}
		\Xhline{1.3pt}
		SR & \multicolumn{2}{c|}{20\%} & \multicolumn{2}{c|}{15\%} &  \multicolumn{2}{c}{10\%} \\ \hline
		Metrics & PSNR & SSIM & PSNR & SSIM & PSNR & SSIM \\ \Xhline{0.8pt}
		HaLRTC & 25.56  & 0.790  & 24.24  & 0.742  & 22.54  & 0.676  \\
		SPC-TV & 30.74  & 0.895  & 30.01  & 0.882  & 28.86  & 0.859  \\ 
		TNN-FFT & 35.51  & 0.953  & 34.15  & 0.940  & 32.50  & 0.921  \\ 
		TNN-DCT & 35.67  & 0.954  & 34.28  & 0.941  & 32.58  & 0.923  \\ 
		LRTC-TV & 31.91  & 0.909  & 26.22  & 0.780  & 23.84  & 0.692  \\ 
		t-CTV & 38.35  & 0.975  & 37.06  & 0.968  & 35.39  & 0.958  \\ 
		DIP & 35.75  & 0.962  & 34.26  & 0.952  & 32.30  & 0.934  \\ 
		S2DIP & 35.96  & 0.967  & 35.45  & 0.965  & 34.68  & 0.960  \\ 
		\textbf{NGR} & \textbf{38.75}  & \textbf{0.979}  & \textbf{37.31}  & \textbf{0.973}  & \textbf{35.63}  & \textbf{0.964} \\\Xhline{1.3pt}
	\end{tabular}
\end{table}

\begin{figure*}[t]
	\centering
	\subfigure[\centering\scriptsize Observation]    {\includegraphics[width=0.15\linewidth]{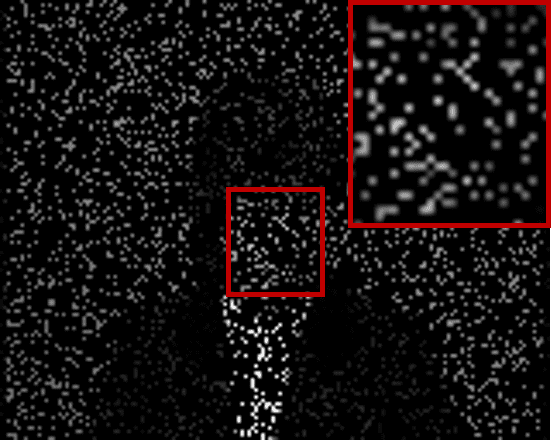} }
	\subfigure[\centering\scriptsize HaLRTC(26.53)]  {\includegraphics[width=0.15\linewidth]{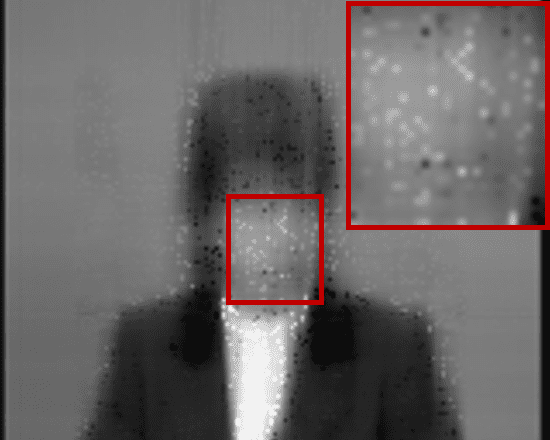} }
	\subfigure[\centering\scriptsize SPC-TV(25.65)]  {\includegraphics[width=0.15\linewidth]{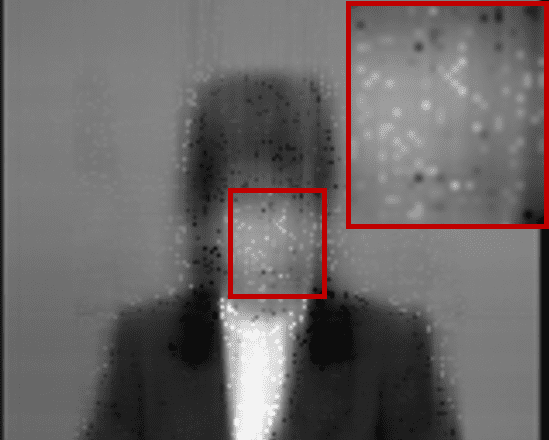} }
	\subfigure[\centering\scriptsize TNN-FFT(36.25)] {\includegraphics[width=0.15\linewidth]{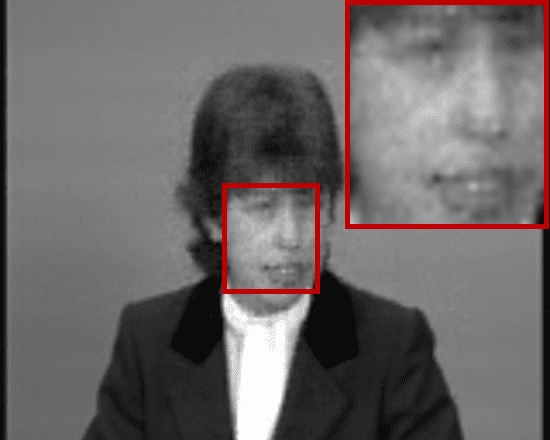} }
	\subfigure[\centering\scriptsize TNN-DCT(36.70)] {\includegraphics[width=0.15\linewidth]{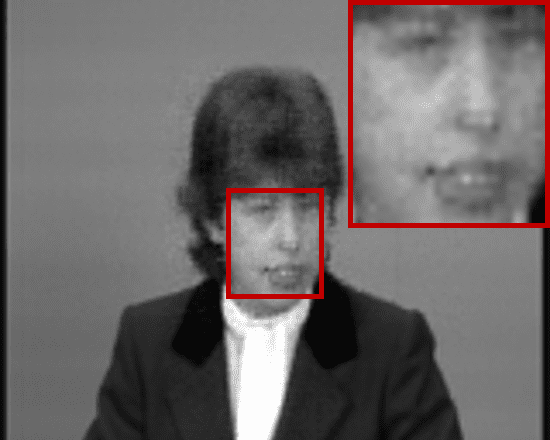} }
	\subfigure[\centering\scriptsize LRTC-TV(26.31)] {\includegraphics[width=0.15\linewidth]{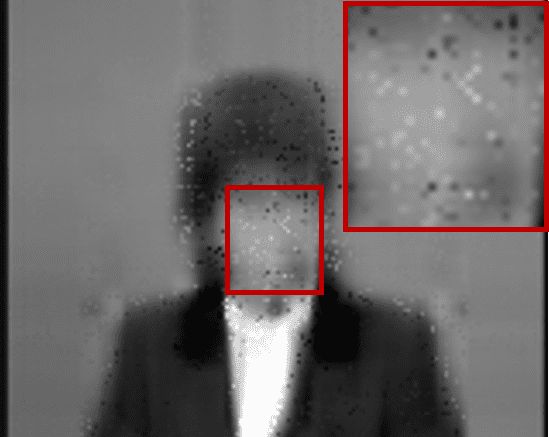} }
	\subfigure[\centering\scriptsize t-CTV(39.32)]   {\includegraphics[width=0.15\linewidth]{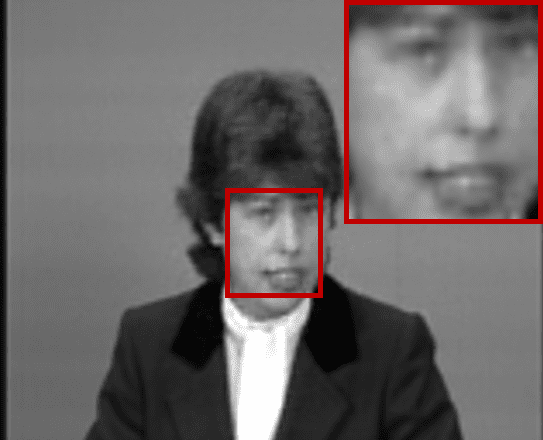} }
	\subfigure[\centering\scriptsize DIP(36.67)]     {\includegraphics[width=0.15\linewidth]{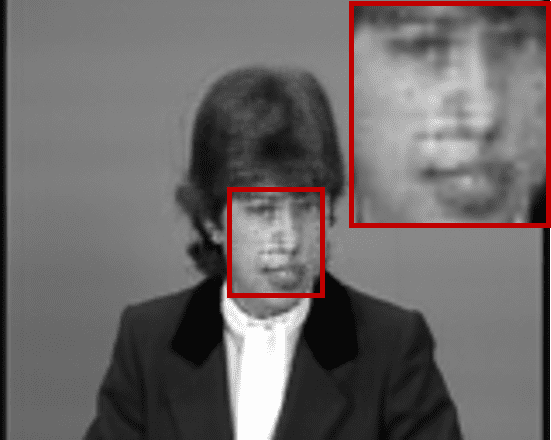} }
	\subfigure[\centering\scriptsize S2DIP(38.72)]   {\includegraphics[width=0.15\linewidth]{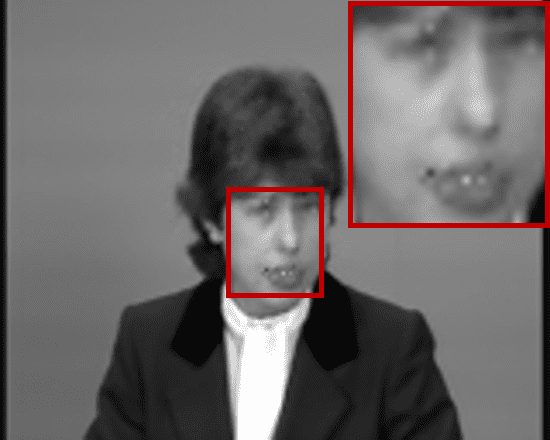} }
	\subfigure[\centering\scriptsize NGR(40.01)]     {\includegraphics[width=0.15\linewidth]{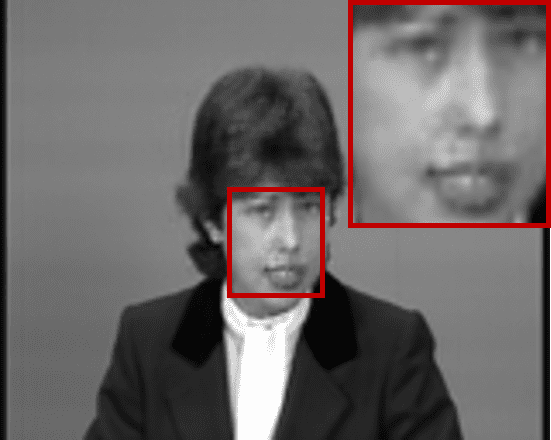} }
	\subfigure[\centering\scriptsize GT]             {\includegraphics[width=0.15\linewidth]{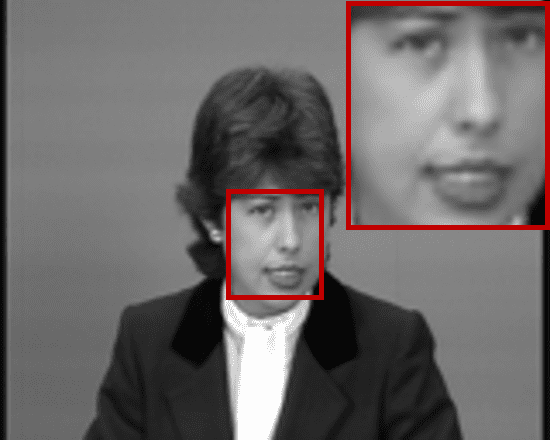} }
	\caption{GT, observed video and inpainting visual results, corresponding PSNR values by methods for comparison on claire dataset (SR = 15\%). The 178th frame is selected to display.\label{video_inpainting fig}}
\end{figure*} 

\section{Experiments}\label{sec:experiments}
This study mainly focuses on visual data inpainting and denoising tasks to testify the performance of NGR as well as other related SOTA methods. Specifically, we adopt the same network architecture as proposed by Ulyanov et al. in \cite{ulyanov2018deep}, which utilizes a U-Net with skip connections. It is important to note that, in contrast to other network architectures (discussed in Section \ref{backbone}), the NGR exhibits better robustness to the choice of backbone mainly attributed to its gradient prediction principle.

We employ peak signal-to-noise ratio (PSNR) and structural similarity (SSIM) as the evaluation metrics for all experiments. For HSI data, we additionally incorporate spectral angle mapper (SAM) and erreur relative global adimensionnelle de synthèse (ERGAS) metrics to more accurately assess the quality of channel-wise restoration. All experiments are conducted on a server equipped with Python 3.9.0, PyTorch 2.0.0, and Nvidia GeForce RTX 2080Ti GPUs. Our source code is publicly available at \url{https://github.com/yyfz/Neural-Gradient-Regularizer}, facilitating reproducibility and further development of the proposed method.

\subsection{Visual data inpainting}
We first demonstrate the applicability of the proposed NGR to inpainting tasks involving diverse visual data types, encompassing RGB images, videos, and HSIs. Furthermore, we verify the effectiveness of the NGR in addressing a specific image inpainting task, namely multi-temporal MSI decloud.

The comparison methods include: high accuracy low-rank tensor completion (HaLRTC) \footnote[1]{https://www.cs.rochester.edu/u/jliu/publications.html}\cite{liu2012tensor}, smooth PARAFAC tensor completion with TV (SPC-TV) \footnote[2]{https://ieeexplore.ieee.org/document/7502115/media\#media} \cite{yokota2016smooth},  tensor nuclear norm minimization using fast fourier transform (TNN-FFT) \footnote[3]{https://github.com/canyilu/tensor-completion-tensor-recovery}\cite{lu2018exact}, tensor nuclear norm minimization using discrete cosine transform (TNN-DCT) \footnote[4]{https://github.com/canyilu/Tensor-robust-PCA-and-tensorcompletion-under-linear-transform} \cite{lu2019low}, low-rank tensor completion with TV (LRTC-TV) \footnote[5]{https://github.com/zhaoxile/Tensor-completion-using-total-variation-and-low-rank-matrix-factorization}\cite{ji2016tensor}, tensor correlated TV (t-CTV) \footnote[6]{https://github.com/wanghailin97/Guaranteed-Tensor-Recovery-Fused-Low-rankness-and-Smoothness} \cite{wang2023guaranteed}, deep image prior (DIP) \footnote[7]{https://github.com/DmitryUlyanov/deep-image-prior, 
	
	https://github.com/acecreamu/deep-hs-prior} \cite{ulyanov2018deep, sidorov2019deep} and TV-SSTV constrained deep image prior (S2DIP) \footnote[8]{https://github.com/YisiLuo/S2DIP}\cite{luo2021hyperspectral}. All compared methods are carried out by the official implementation with recommended hyperparameters.

\subsubsection{RGB images inpainting}
Three color image datasets are applied to verify the performance of NGR. They are BSDS100 \cite{martin2001database}, set5 \cite{bevilacqua2012low} and five color images from USC-SIPI\footnote[9]{https://sipi.usc.edu/database/}. The evaluation encompasses three cases with sampling rate (SR) 50\%, 30\% and 10\%, respectively. 

It can be observed that NGR achieves superior performance in RGB image inpainting, as evidenced by the metrics listed in Table \ref{RGB_inpainting table}. When compared to SPC-TV and LRTC-TV, NGR exhibits superiority, even though they incorporate TV regularizer with low-rank prior. As a state-of-the-art TV variant, t-CTV can achieve competitive performance with NGR on the BSDS100 dataset, but with a significantly lower PSNR value and a higher SSIM value. However, on other datasets, t-CTV underperforms NGR in terms of both PSNR and SSIM, indicating the robust and consistent performance of NGR across different datasets.

\begin{figure*}[t]
	\centering
	\subfigure[\centering\scriptsize Observation]    {\includegraphics[width=0.15\linewidth]{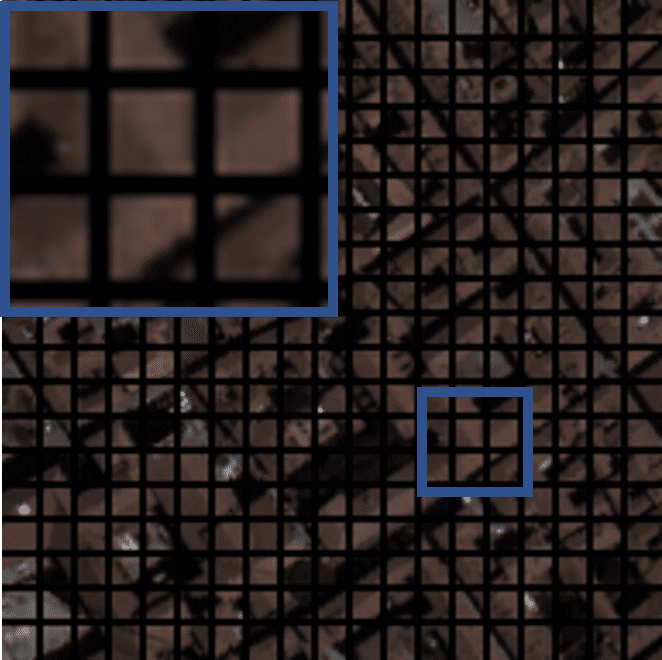} }
	\subfigure[\centering\scriptsize HaLRTC(18.62)]  {\includegraphics[width=0.15\linewidth]{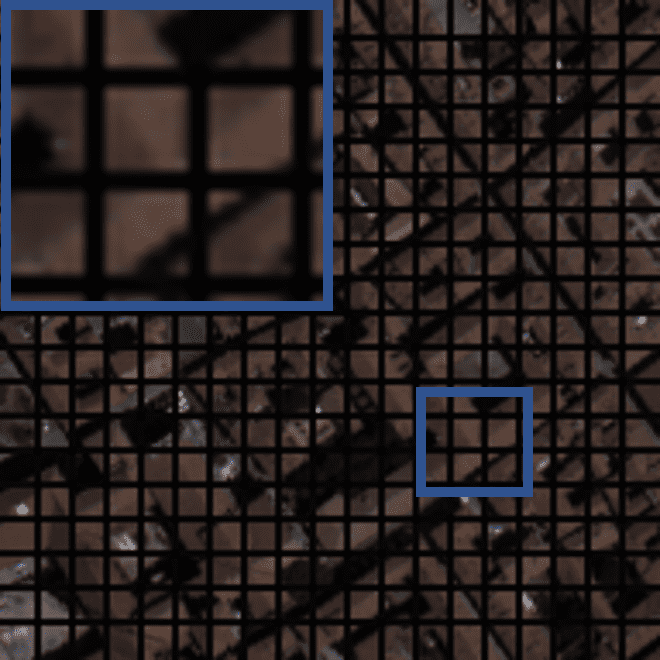} }
	\subfigure[\centering\scriptsize SPC-TV(21.26)]  {\includegraphics[width=0.15\linewidth]{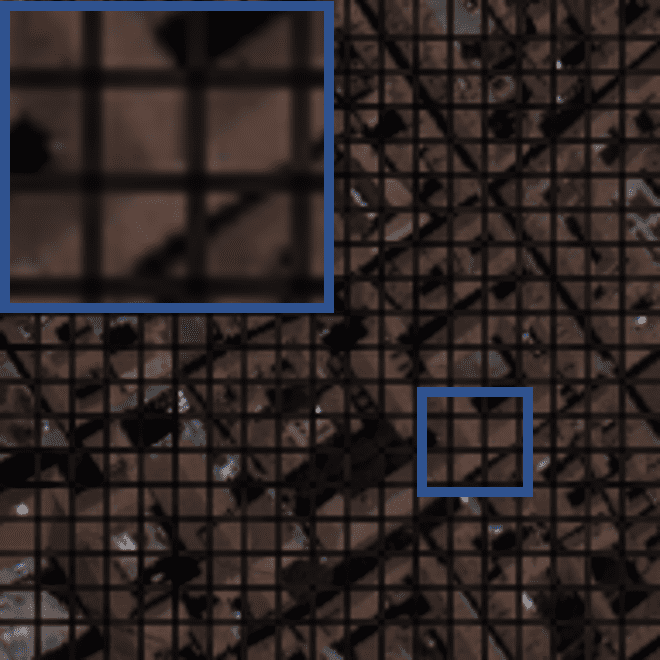} }
	\subfigure[\centering\scriptsize TNN-FFT(17.59)] {\includegraphics[width=0.15\linewidth]{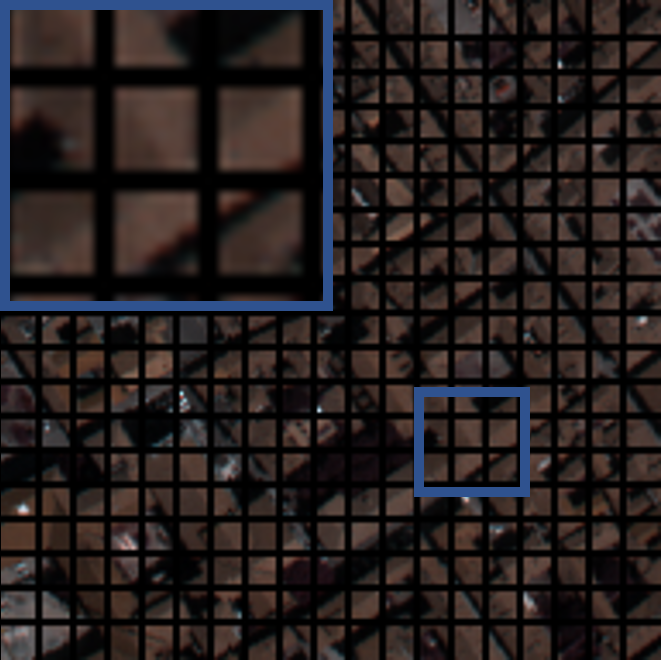} }
	\subfigure[\centering\scriptsize TNN-DCT(17.59)] {\includegraphics[width=0.15\linewidth]{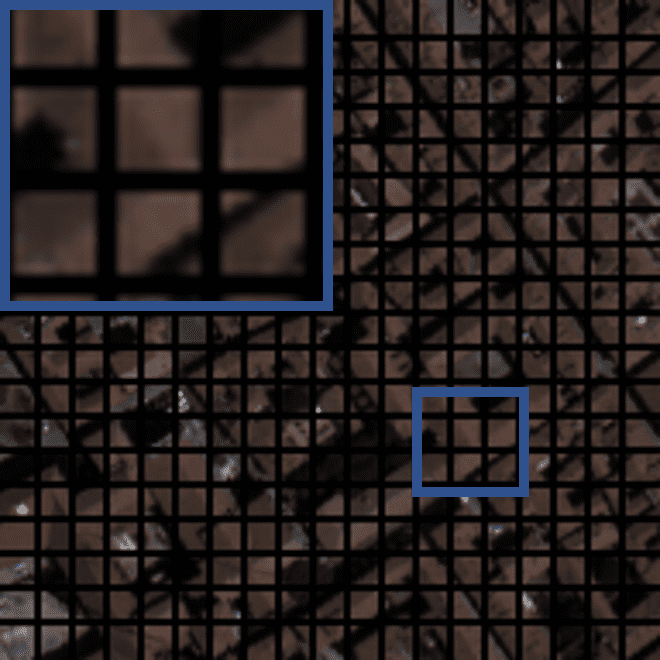} }
	\subfigure[\centering\scriptsize LRTC-TV(24.54)] {\includegraphics[width=0.15\linewidth]{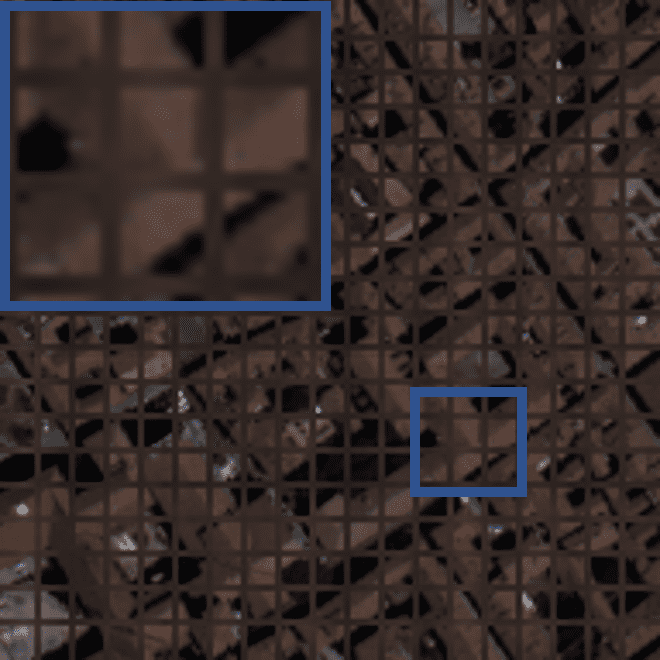} }
	\subfigure[\centering\scriptsize t-CTV(30.72)]   {\includegraphics[width=0.15\linewidth]{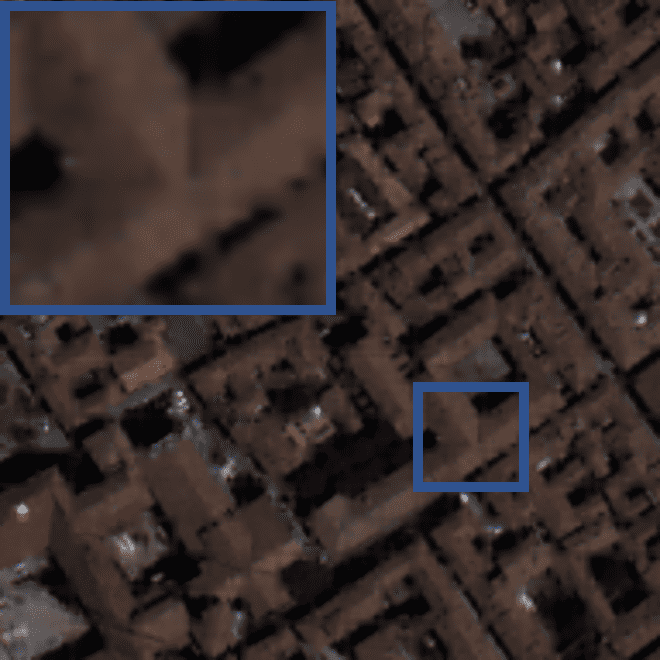} }
	\subfigure[\centering\scriptsize DIP(32.22)]     {\includegraphics[width=0.15\linewidth]{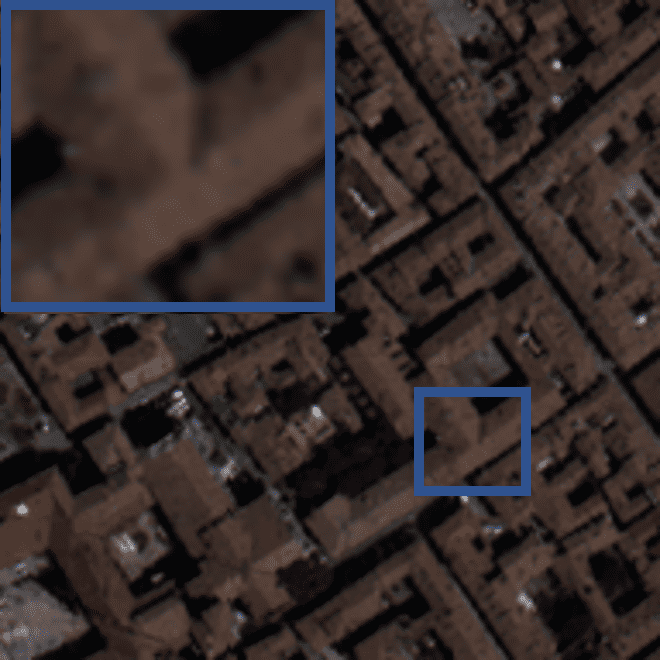} }
	\subfigure[\centering\scriptsize S2DIP(33.27)]   {\includegraphics[width=0.15\linewidth]{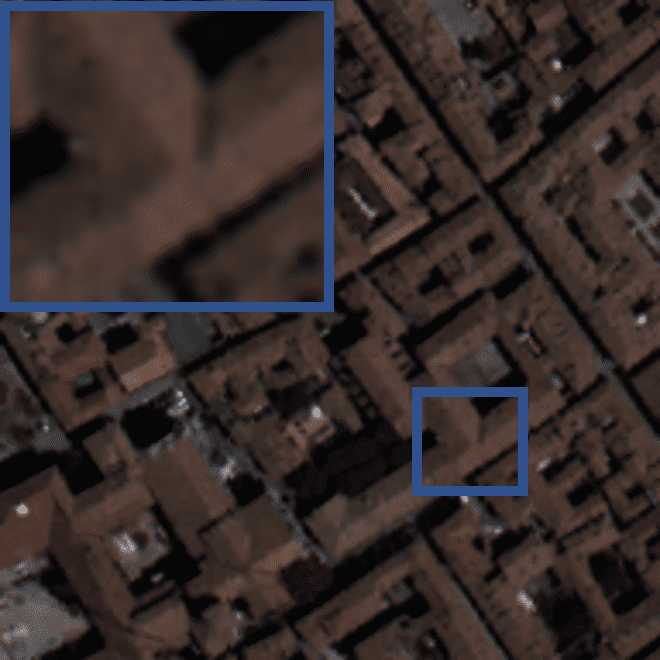} }
	\subfigure[\centering\scriptsize NGR(33.78)]     {\includegraphics[width=0.15\linewidth]{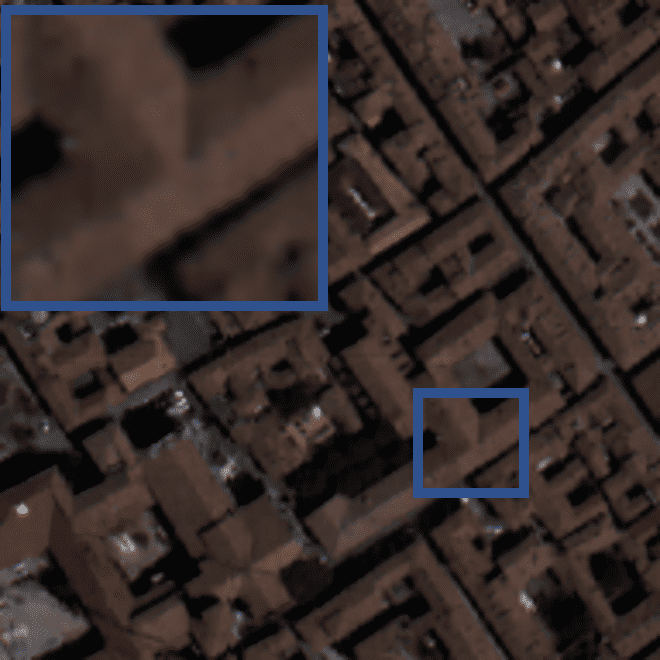} }
	\subfigure[\centering\scriptsize GT]             {\includegraphics[width=0.15\linewidth]{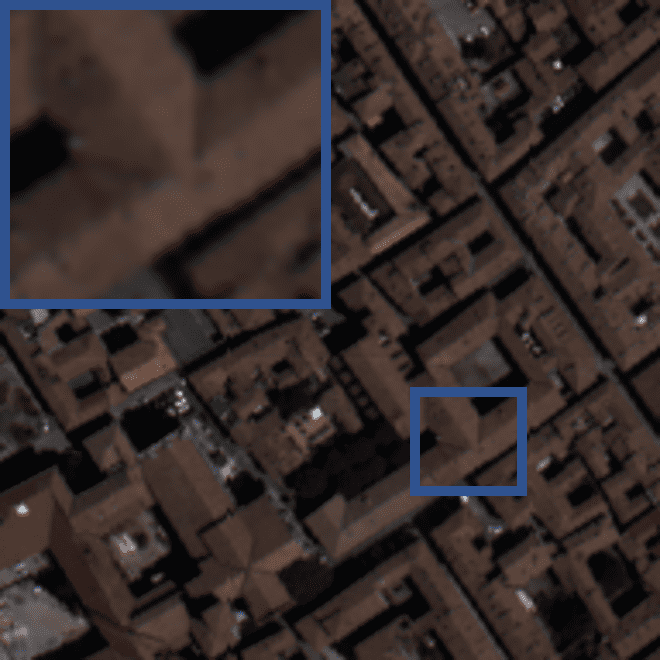} }
	\caption{GT, observed HSI, inpainting visual results and corresponding PSNR values by methods for comparison on PA. The pseudo images consisted of the 47-th, 20-th and 14-th bands are selected to display.\label{HSI_deadline fig}}
\end{figure*}

\begin{table*}[!t]
	\caption{The quantitive results of HSI inpainting on PA and WDC. The best results are marked in bold.\label{HSI inpainting table}}
	\centering
	\resizebox{\linewidth}{!}{
		\begin{tabular}{c|c|cccc|cccc|cccc|cccc}
			\Xhline{1.3pt}
			~ & SR & \multicolumn{4}{c|}{10\%} & \multicolumn{4}{c|}{7.5\%} & \multicolumn{4}{c|}{5\%} & \multicolumn{4}{c}{Deadlines} \\ \cline{1-18}
			~ & Metrics & PSNR & SSIM & SAM & ERGAS & PSNR & SSIM & SAM & ERGAS & PSNR & SSIM & SAM & ERGAS & PSNR & SSIM & SAM & ERGAS \\  \Xhline{0.8pt}
			\multirow{9}[0]{*}{PA} & HaLRTC & 21.02  & 0.404  & 11.202  & 44.04  & 20.36  & 0.345  & 11.172  & 47.50  & 19.77  & 0.296  & 10.971  & 50.87  & 18.62  & 0.455  & 3.166  & 57.92  \\
			~ & SPC-TV & 32.96  & 0.941  & 6.006  & 12.03  & 30.46  & 0.903  & 6.995  & 15.73  & 26.87  & 0.801  & 8.417  & 23.08  & 21.26  & 0.578  & 2.893  & 42.78  \\
			~ & TNN-FFT & 32.68  & 0.926  & 8.206  & 14.71  & 30.99  & 0.902  & 9.005  & 16.97  & 28.90  & 0.858  & 9.973  & 20.52  & 17.59  & 0.394  & 30.529  & 65.24  \\ 
			~ & TNN-DCT & 38.10  & 0.980  & 3.970  & 6.71  & 35.37  & 0.964  & 4.974  & 9.05  & 31.94  & 0.928  & 6.526  & 13.17  & 17.59  & 0.394  & 30.527  & 65.24  \\
			~ & LRTC-TV & 36.46  & 0.968  & 4.835  & 7.98  & 35.44  & 0.961  & 5.434  & 8.88  & 33.75  & 0.943  & 6.708  & 10.94  & 24.54  & 0.707  & 2.818  & 29.35  \\
			~ & t-CTV & 37.02  & 0.968  & 5.675  & 9.25  & 35.04  & 0.956  & 6.522  & 10.94  & 32.66  & 0.933  & 7.689  & 13.64  & 30.72  & 0.911  & 2.022  & 14.44  \\
			~ & DIP & 45.51  & 0.996  & 1.819  & 2.85  & 43.16  & 0.994  & 2.134  & 3.78  & 40.54  & 0.990  & 2.488  & 5.10  & 32.22  & 0.930  & 1.914  & 12.25  \\
			~ & S2DIP & 42.89  & 0.994  & 2.147  & 3.76  & 40.86  & 0.991  & 2.412  & 4.78  & 38.64  & 0.986  & 2.824  & 6.13  & 33.27  & 0.951  & 1.724  & 10.82  \\ 
			~ & \textbf{NGR} & \textbf{46.39}  & \textbf{0.997}  & \textbf{1.676}  & \textbf{2.65}  & \textbf{44.87}  & \textbf{0.996}  & \textbf{1.805}  & \textbf{3.10}  & \textbf{41.74}  & \textbf{0.992}  & \textbf{2.365}  & \textbf{4.66}  & \textbf{33.78}  & \textbf{0.962}  & \textbf{1.540}  & \textbf{10.21}\\ \hline
			\multirow{9}[0]{*}{WDC} & HaLRTC & 22.49  & 0.524  & 13.947  & 49.02  & 21.69  & 0.462  & 14.654  & 53.75  & 20.84  & 0.398  & 15.457  & 59.25  & 20.32  & 0.572  & 5.409  & 61.43  \\
			~ & SPC-TV & 32.77  & 0.934  & 6.773  & 15.54  & 31.03  & 0.906  & 7.707  & 18.81  & 28.36  & 0.840  & 9.451  & 25.36  & 23.40  & 0.705  & 4.656  & 43.18  \\ 
			~ & TNN-FFT & 33.54  & 0.939  & 7.993  & 14.82  & 31.95  & 0.917  & 9.071  & 17.44  & 30.03  & 0.880  & 10.490  & 21.35  & 18.70  & 0.454  & 31.720  & 73.91  \\ 
			~ & TNN-DCT & 34.30  & 0.948  & 7.363  & 13.03  & 32.61  & 0.927  & 8.450  & 15.67  & 30.57  & 0.890  & 9.869  & 19.62  & 18.70  & 0.454  & 31.719  & 73.91  \\ 
			~ & LRTC-TV & 34.99  & 0.943  & 10.886  & 13.18  & 34.32  & 0.934  & 12.673  & 14.55  & 31.24  & 0.865  & 19.267  & 22.33  & 20.37  & 0.562  & 5.961  & 61.15  \\ 
			~ & t-CTV & 37.56  & 0.973  & 5.425  & 9.57  & 35.71  & 0.962  & 6.327  & 11.45  & 33.30  & 0.939  & 7.704  & 14.72  & 29.92  & 0.897  & 2.600  & 20.68  \\ 
			~ & DIP & 44.01  & 0.993  & 2.609  & 4.48  & 43.00  & 0.992  & 2.801  & 4.97  & 41.37  & 0.989  & 3.069  & 5.87  & 28.83  & 0.838  & 3.421  & 23.46  \\
			~ & S2DIP & 39.84  & 0.982  & 3.553  & 6.80  & 38.95  & 0.979  & 3.764  & 7.47  & 37.00  & 0.968  & 4.353  & 9.20  & 29.82  & 0.901  & 2.476  & 20.82  \\
			~ & \textbf{NGR} & \textbf{45.92}  & \textbf{0.995}  & \textbf{2.136}  & \textbf{3.64}  & \textbf{43.87}  & \textbf{0.993}  & \textbf{2.447}  & \textbf{4.46}  & \textbf{41.60}  & \textbf{0.990}  & \textbf{2.935}  & \textbf{5.66} & \textbf{30.64}  & \textbf{0.917}  & \textbf{2.291}  & \textbf{19.02} \\ \Xhline{1.3pt}
		\end{tabular}
	}
\end{table*}

The superiority of the proposed method is exemplified by its better performance in edge preservation, as displayed in Fig. \ref{RGB_inpainting fig}. t-CTV and S2DIP generate evident artifacts, while NGR faithfully retains the edge information. Additionally, these methods are applied to a more difficult task, text inpainting. NGR achieves the highest PSNR among all competing methods, as shown in Fig. \ref{text_inpainting fig}. Most methods result in faint text imprints around the horse, but only S2DIP and NGR provide fine restoration without text imprints. Notably, S2DIP utilizes a combination of three regularizers (i.e. DIP, TV and SSTV) to achieve this, while NGR employs only one automatically learned regularizer.

\subsubsection{Videos inpainting}
Eight widely used videos\footnote[10]{http://trace.eas.asu.edu/yuv/index.html} are selected, and SR is chosen as 20\%, 15\% and 10\%. As shown in Table \ref{video_inpainting table}, NGR achieves the highest PSNR and SSIM among all compared methods, with a consistent performance gain of 0.2dB-0.4dB over t-CTV in terms of PSNR. The visual results presented in Fig. \ref{video_inpainting fig} demonstrate that NGR performs best in restoring facial details and structures, while other methods either fail to achieve complete restoration or result in local blurriness.

\subsubsection{HSIs inpainting}
Next, the experiment is conducted on two typical HSI datasets, Pavia Centre (PA) \footnote[11]{https://www.ehu.eus/ccwintco/index.php/Hyperspectral\_Remote
	
	\_Sensing\_Scenes\#Pavia\_Centre\_scene} and Washington D.C. (WDC) \footnote[12]{https://engineering.purdue.edu/${\sim}$biehl/MultiSpec/hyperspectral.html}, with SR configured as 10\%, 7.5\% and 5\%, respectively. Additionally, remote sensing HSI data is often corrupted by deadlines, where entire rows/columns of pixels are missing. Deadline removal is a more challenging task since there is no additional information available for the missing regions, which are contiguous blocks absent in all channels. Therefore, particular attention requires to be specifically paid to deadline removal.

The metrics reported in Table \ref{HSI inpainting table} indicate that NGR outperforms all other competing methods. Only NGR achieves a high-precision restoration, with PSNR exceeding 41dB and SSIM surpassing 0.99 for cases with randomly missing pixels. As for deadline removal, the recovered images displayed in Fig \ref{HSI_deadline fig} demonstrate that conventional methods such as TV regularization, low-rank regularization, or their combination are insufficient to recover from such severely corrupted HSI data. On the other hand, approaches based on correlated TV and untrained neural networks manage to achieve satisfactory image restoration, but NGR attains the best results through precise predictions of gradient maps.

\begin{figure*}[t]
	\centering
	\subfigure[\centering\scriptsize Observation]    {\includegraphics[width=0.15\linewidth]{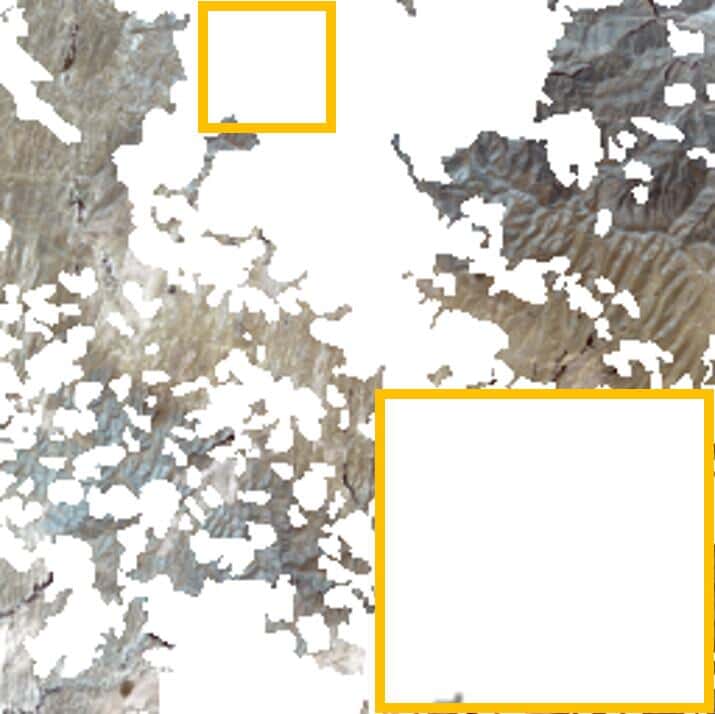} }
	\subfigure[\centering\scriptsize HaLRTC(23.89)]  {\includegraphics[width=0.15\linewidth]{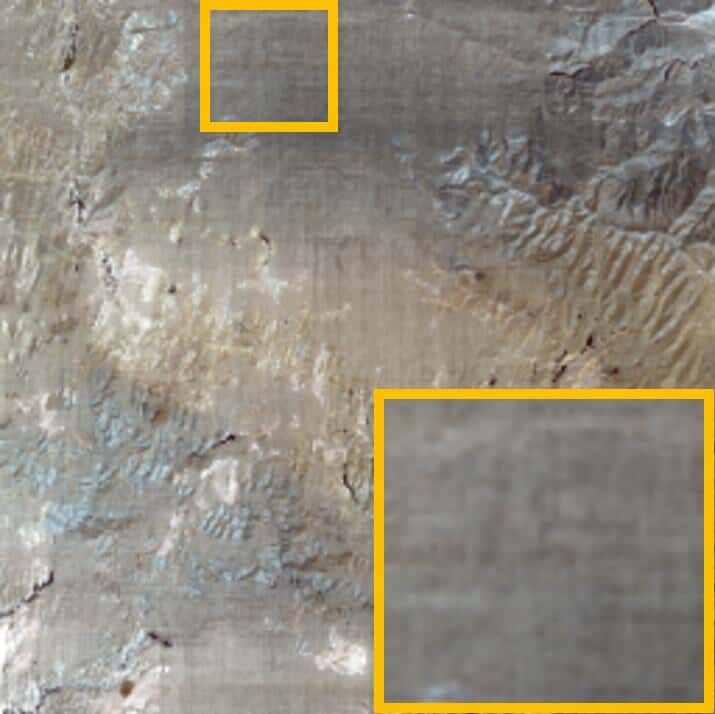} }
	\subfigure[\centering\scriptsize SPC-TV(27.59)]  {\includegraphics[width=0.15\linewidth]{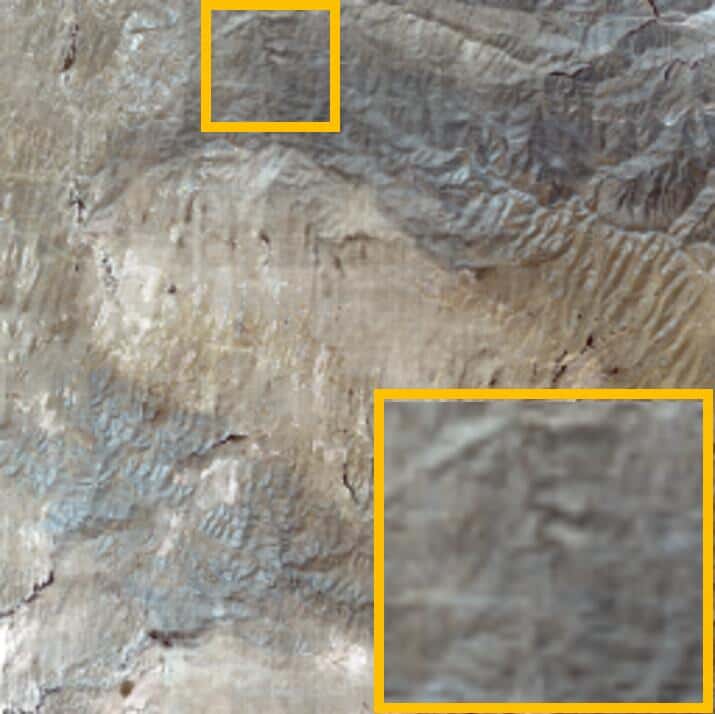} }
	\subfigure[\centering\scriptsize TNN-FFT(27.39)] {\includegraphics[width=0.15\linewidth]{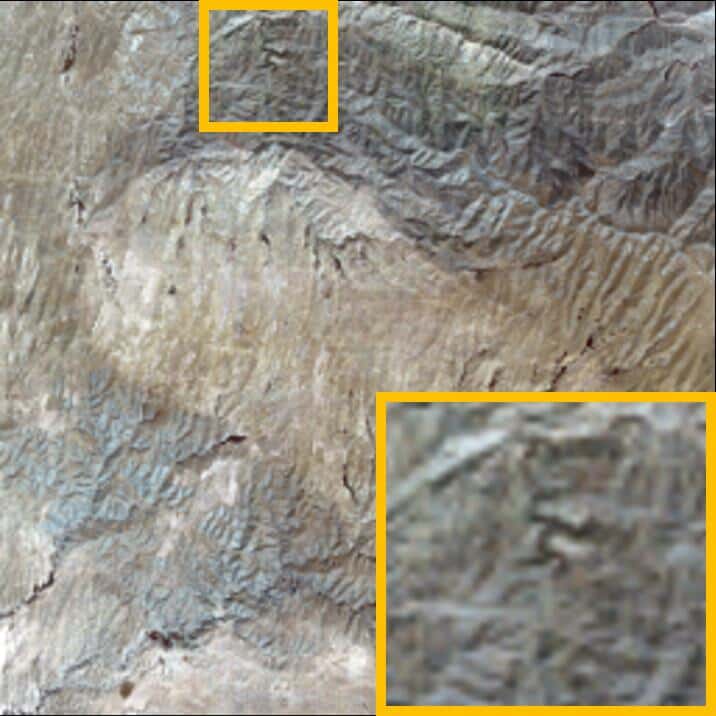} }
	\subfigure[\centering\scriptsize TNN-DCT(28.32)] {\includegraphics[width=0.15\linewidth]{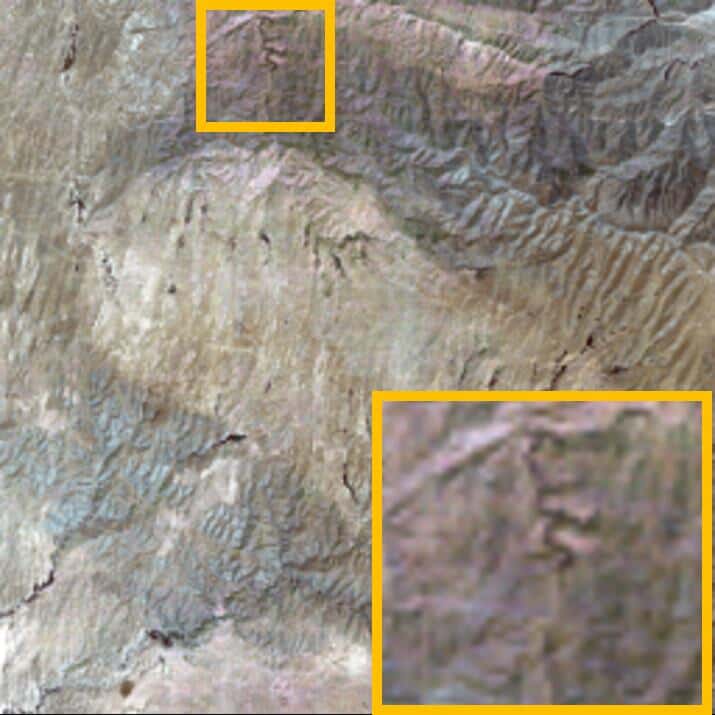} }
	\subfigure[\centering\scriptsize LRTC-TV(23.01)] {\includegraphics[width=0.15\linewidth]{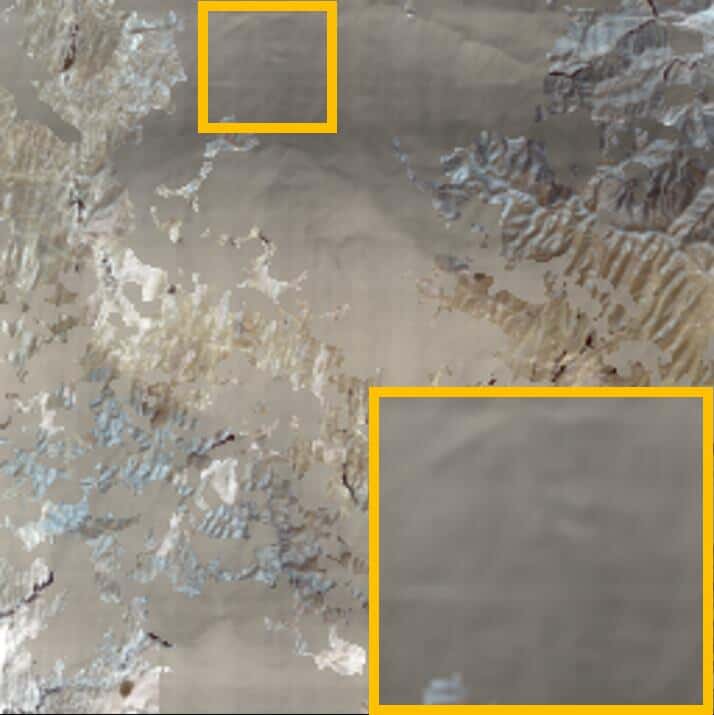} }
	\subfigure[\centering\scriptsize t-CTV(28.13)]   {\includegraphics[width=0.15\linewidth]{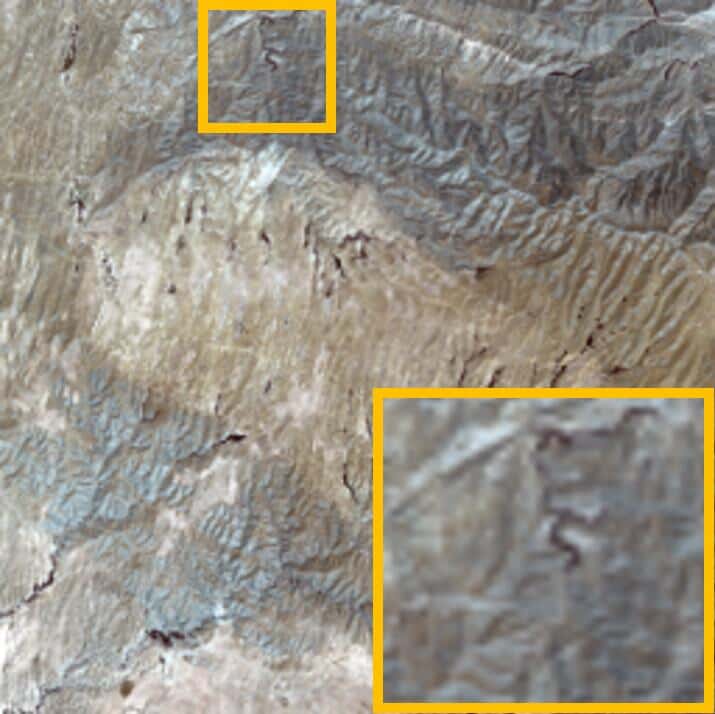} }
	\subfigure[\centering\scriptsize DIP(35.45)]     {\includegraphics[width=0.15\linewidth]{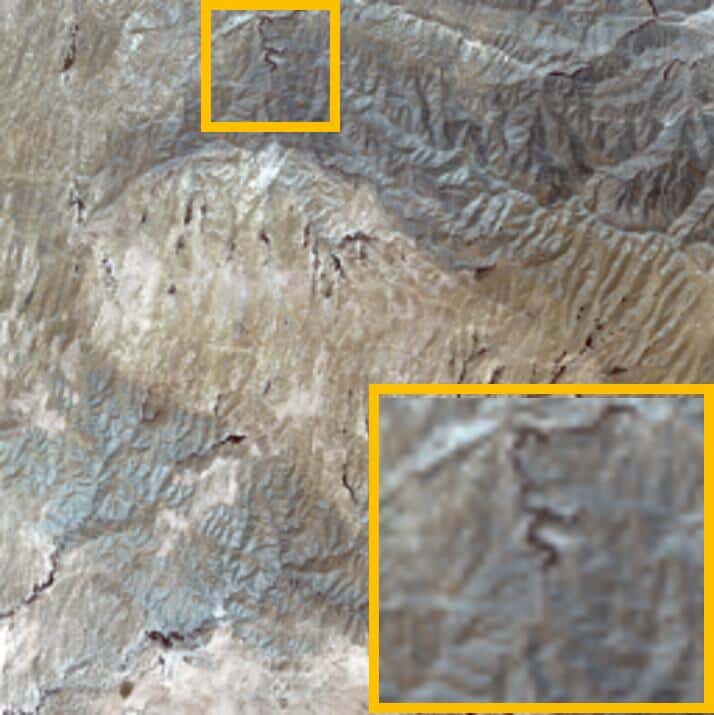} }
	\subfigure[\centering\scriptsize S2DIP(29.46)]   {\includegraphics[width=0.15\linewidth]{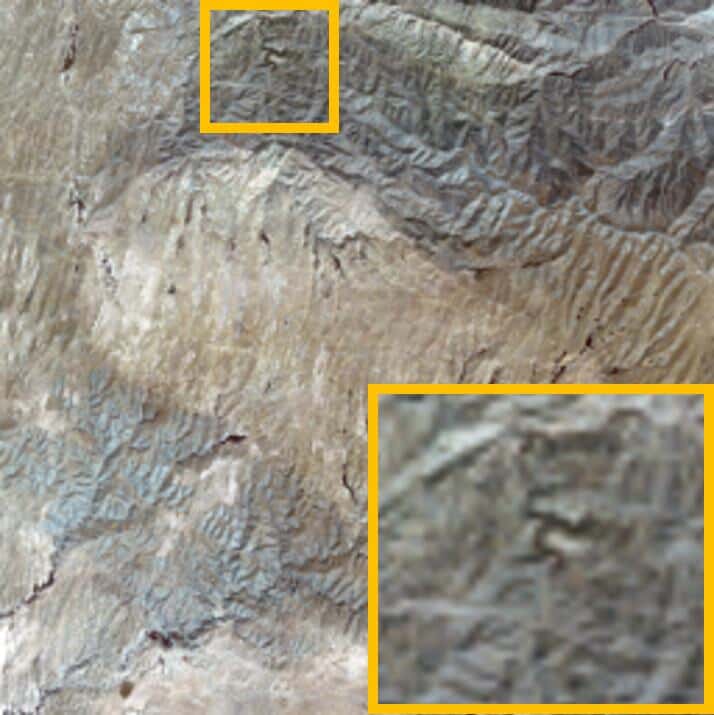} }
	\subfigure[\centering\scriptsize NGR(36.19)]     {\includegraphics[width=0.15\linewidth]{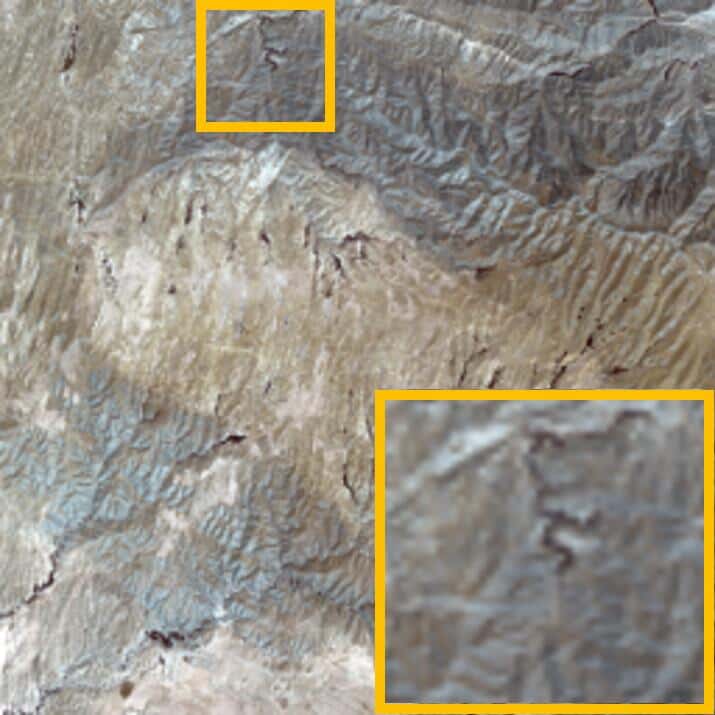} }
	\subfigure[\centering\scriptsize GT]             {\includegraphics[width=0.15\linewidth]{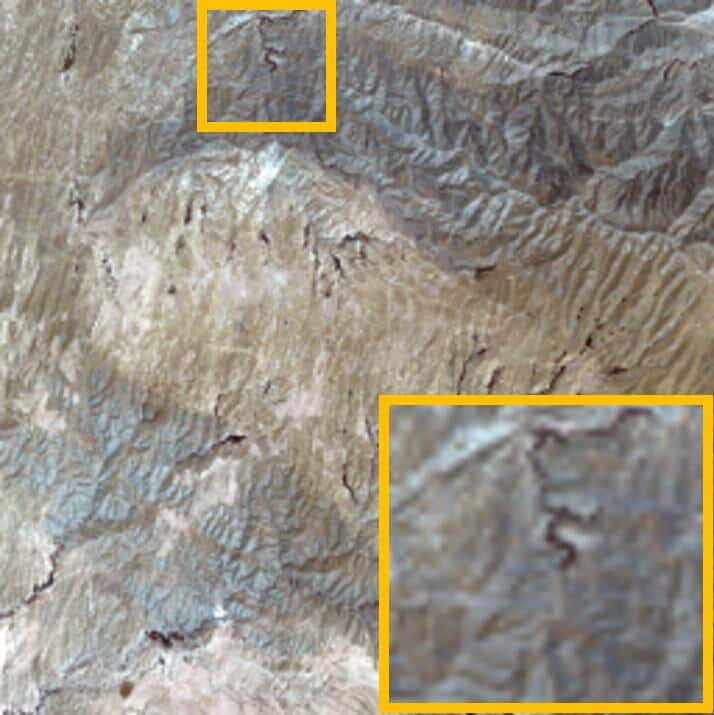} }
	\caption{GT, observed HSI, inpainting visual results and corresponding PSNR values by methods for comparison on Forish Mountain with large cloud mask. The pseudo images consisted of the 3-rd, 2-nd and 1-st bands are selected to display.\label{decloud fig}}
\end{figure*}

\begin{figure*}[h]
	\centering
	\subfigure[\centering\scriptsize Observation] {\includegraphics[width=0.15\linewidth]{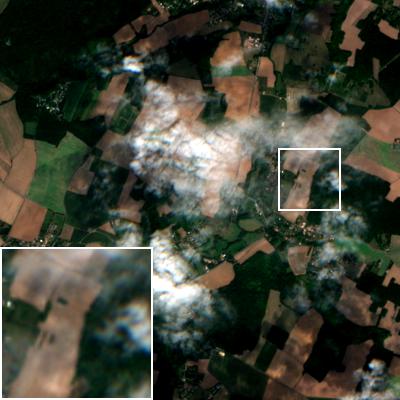} }
	\subfigure[\centering\scriptsize t-CTV]       {\includegraphics[width=0.15\linewidth]{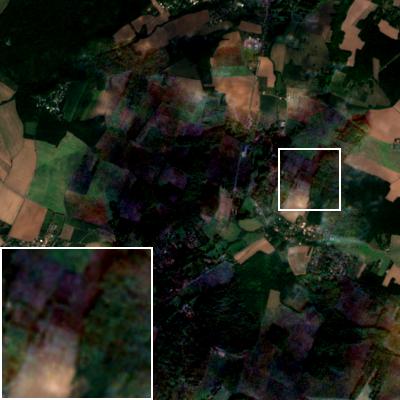} }
	\subfigure[\centering\scriptsize TNN-DCT]     {\includegraphics[width=0.15\linewidth]{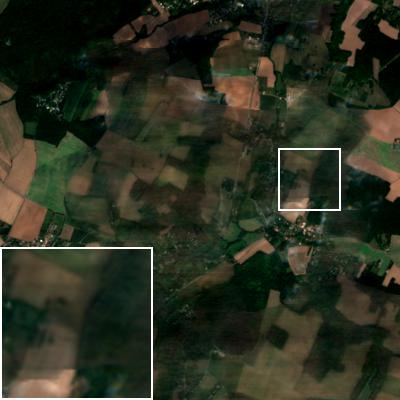} }
	\subfigure[\centering\scriptsize DIP]         {\includegraphics[width=0.15\linewidth]{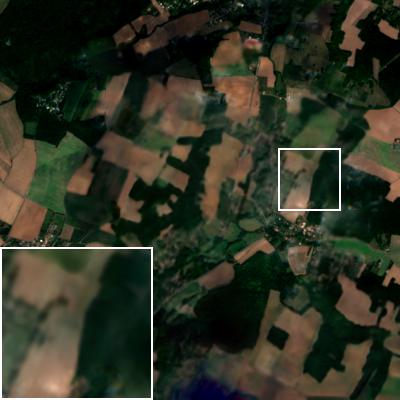} }
	\subfigure[\centering\scriptsize S2DIP]       {\includegraphics[width=0.15\linewidth]{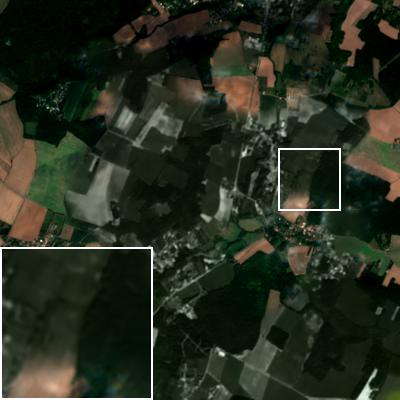} }
	\subfigure[\centering\scriptsize NGR]         {\includegraphics[width=0.15\linewidth]{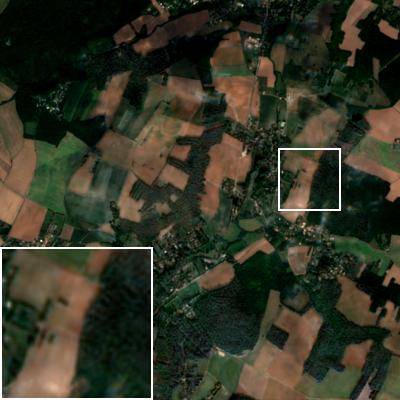} }
	\subfigure[\centering\scriptsize Observation] {\includegraphics[width=0.15\linewidth]{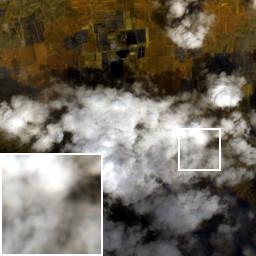} }
	\subfigure[\centering\scriptsize t-CTV]       {\includegraphics[width=0.15\linewidth]{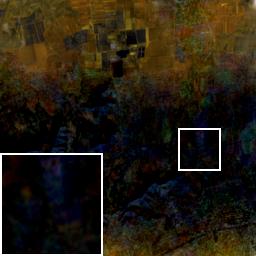} }
	\subfigure[\centering\scriptsize TNN-DCT]     {\includegraphics[width=0.15\linewidth]{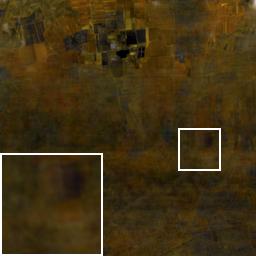} }
	\subfigure[\centering\scriptsize DIP]         {\includegraphics[width=0.15\linewidth]{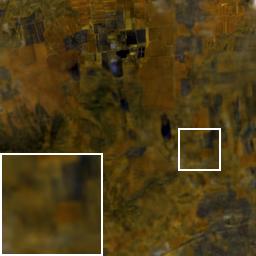} }
	\subfigure[\centering\scriptsize S2DIP]       {\includegraphics[width=0.15\linewidth]{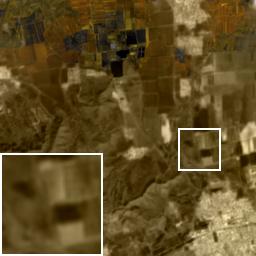} }
	\subfigure[\centering\scriptsize NGR]         {\includegraphics[width=0.15\linewidth]{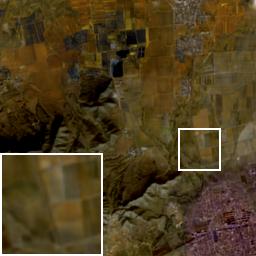} }
	\caption{The visual results on real-world data, France (the first row) and China (the second row), obtained by several representative methods.  \label{real_decloud fig}}
\end{figure*}
\subsubsection{Multi-temporal MSIs decloud}
Multi-temporal MSIs decloud is a special inpainting task for remote sensing, wherein cloud-free images of different timestamps are utilized to facilitate the restoration of cloud-affected images. Suppose there are MSIs captured at $T$ timestamps denoted as $\bm{\mathcal M}_t \in \mathbb R^{B\times H\times W} (t =1,2,\cdots ,T)$, with the assumption that $\bm{\mathcal{M}}_{1}$ is a cloudy MSI, while the others are cloud-free MSIs. These MSIs are concatenated along the channel dimension to construct a multi-temporal MSI, denoted as $\bm{\mathcal{Y}}=\left[\bm{\mathcal{M}}_{1},\bm{\mathcal{M}}_{2},\cdots,\bm{\mathcal{M}}_{T}\right]\in\mathbb{R}^{BT\times H\times W}$. Generally speaking, it aims to recover the missing values in cloudy regions in $\bm{\mathcal{M}}_{1}$.

The experimental setup remains consistent with the aforementioned configurations. The experimental data includes Forish Mountain with 8 bands and 4 timestamps, Forish Farmland with 8 bands and 4 timestamps, and Beijing with 6 bands and 4 timestamps. Three real cloud masks of varying sizes (small, medium and large) are selected from the WHU cloud dataset \cite{ji2020simultaneous}. Table \ref{decloud table} unequivocally demonstrates the superior performance of NGR over other methods across all evaluation metrics. Additionally, the visual results depicted in Fig. \ref{decloud fig} illustrate the remarkable capabilities of NGR in reconstruction of fine details from multi-temporal images.

Aside from synthetic data, Fig. \ref{real_decloud fig} presents the results obtained on two real-world datasets, namely, France and China. The findings indicate that t-CTV performs poorly on real-world datasets, largely due to its inability to effectively recover missing pixels in cloudy regions. Both TNN-DCT and S2DIP experience evident color inconsistency issues, while DIP tends to lose informative details. Conversely, NGR exhibits relatively more satisfactory texture reconstruction performance.

\begin{table}[!t]
	\caption{The quantitive results of multi-temporal MSIs decloud on Forish Farmland, Forish Mountain and Beijing. The best results are marked in bold.\label{decloud table}}
	\resizebox{\linewidth}{!}{
		\centering
		\begin{tabular}{c|c|cc|cc|cc}
			\Xhline{1.3pt}
			~ & Size & \multicolumn{2}{c|}{S} & \multicolumn{2}{c|}{M} & \multicolumn{2}{c}{L} \\ \hline
			~ & Metrics & PSNR & SSIM & PSNR & SSIM & PSNR & SSIM \\ \Xhline{0.8pt}
			\multirow{9}[0]{*}{\rotatebox{90}{Beijing}} & HaLRTC & 44.79  & 0.987  & 40.78  & 0.968  & 34.84  & 0.917  \\ 
			~ & SPC-TV & 47.11  & 0.990  & 43.72  & 0.976  & 38.38  & 0.943  \\ 
			~ & TNN-FFT & 45.80  & 0.989  & 41.41  & 0.971  & 35.78  & 0.920  \\ 
			~ & TNN-DCT & 46.27  & 0.989  & 41.65  & 0.970  & 35.84  & 0.919  \\ 
			~ & LRTC-TV & 41.68  & 0.977  & 39.77  & 0.960  & 34.91  & 0.902  \\ 
			~ & t-CTV & 47.36  & 0.992  & 42.93  & 0.977  & 36.80  & 0.935  \\ 
			~ & DIP & 51.70  & 0.996  & 47.94  & 0.990  & 39.74  & 0.962  \\ 
			~ & S2DIP & 48.74  & 0.993  & 44.99  & 0.981  & 38.00  & 0.931  \\ 
			~ & \textbf{NGR} & \textbf{52.12}  & \textbf{0.997}  & \textbf{48.91}  & \textbf{0.993}  & \textbf{41.03}  & \textbf{0.971}  \\ \hline
			\multirow{9}[0]{*}{\rotatebox{90}{Forish Mountain}} & HaLRTC & 34.09  & 0.961  & 28.60  & 0.872  & 23.89  & 0.715  \\ 
			~ & SPC-TV & 36.12  & 0.970  & 31.39  & 0.913  & 27.59  & 0.812  \\ 
			~ & TNN-FFT & 35.92  & 0.974  & 31.79  & 0.934  & 27.39  & 0.849  \\ 
			~ & TNN-DCT & 37.76  & 0.979  & 32.46  & 0.937  & 28.32  & 0.868  \\ 
			~ & LRTC-TV & 30.23  & 0.933  & 25.96  & 0.839  & 23.01  & 0.669  \\ 
			~ & t-CTV & 36.67  & 0.977  & 32.46  & 0.941  & 28.13  & 0.862  \\ 
			~ & DIP & 42.88  & 0.988  & 38.45  & 0.966  & 35.45  & 0.927  \\ 
			~ & S2DIP & 39.47  & 0.978  & 33.00  & 0.931  & 29.46  & 0.851  \\ 
			~ & \textbf{NGR} & \textbf{44.07}  & \textbf{0.986}  & \textbf{39.62}  & \textbf{0.966}  & \textbf{36.19}  & \textbf{0.934}  \\ \hline
			\multirow{9}[0]{*}{\rotatebox{90}{Forish Farmland}} & HaLRTC & 29.42  & 0.948  & 24.44  & 0.838  & 20.56  & 0.635  \\ 
			~ & SPC-TV & 29.91  & 0.952  & 25.80  & 0.865  & 22.19  & 0.707  \\ 
			~ & TNN-FFT & 30.03  & 0.956  & 25.77  & 0.878  & 21.79  & 0.729  \\ 
			~ & TNN-DCT & 29.76  & 0.954  & 25.34  & 0.869  & 21.39  & 0.707  \\ 
			~ & LRTC-TV & 27.65  & 0.927  & 23.44  & 0.808  & 20.35  & 0.592  \\ 
			~ & t-CTV & 31.01  & 0.963  & 26.58  & 0.893  & 22.63  & 0.759  \\ 
			~ & DIP & 31.79  & 0.970  & 28.00  & 0.916  & 23.78  & 0.807  \\ 
			~ & S2DIP & 30.47  & 0.956  & 26.35  & 0.882  & 22.24  & 0.720  \\ 
			~ & \textbf{NGR} & \textbf{32.21}  & \textbf{0.972}  & \textbf{28.54}  & \textbf{0.923}  & \textbf{24.44}  & \textbf{0.822}\\ \Xhline{1.3pt}
		\end{tabular}
	}
\end{table}

\subsubsection{Brief summary} 
In this section, we have conducted a comprehensive analysis to compare the performance of different competing methods on the inpainting problem with various types of visual data, encompassing RGB images, videos, HSIs, and multi-temporal MSIs. The results demonstrate that the NGR regularizer consistently achieves good performance across all data types, in contrast to previous state-of-the-art methods that are only applicable to a limited range of data types. This highlights the versatility of NGR, and implies it could be generally usable for a broad range of data types, thereby solidifying its potential for widespread applicability in the field of image processing.

\begin{figure}[t]
	\centering
	\subfigure[\centering\scriptsize Observation]    {\includegraphics[width=0.31\linewidth]{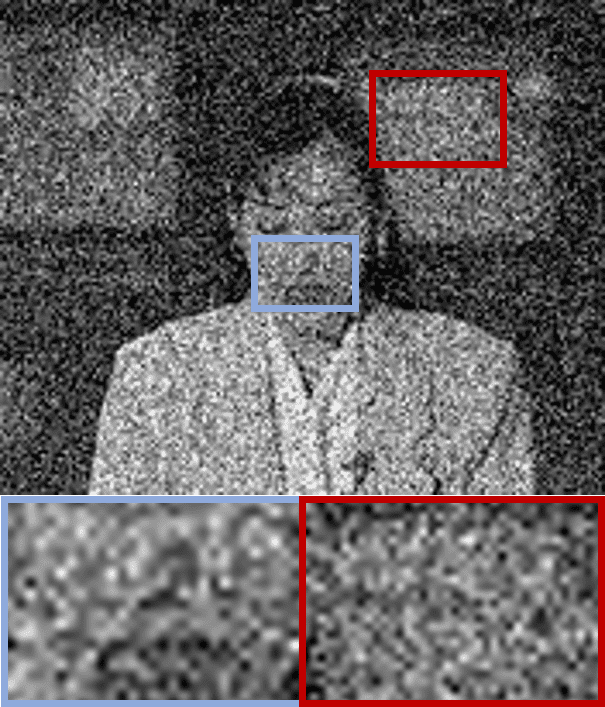} }
	\subfigure[\centering\scriptsize VBM3D(32.79)]   {\includegraphics[width=0.31\linewidth]{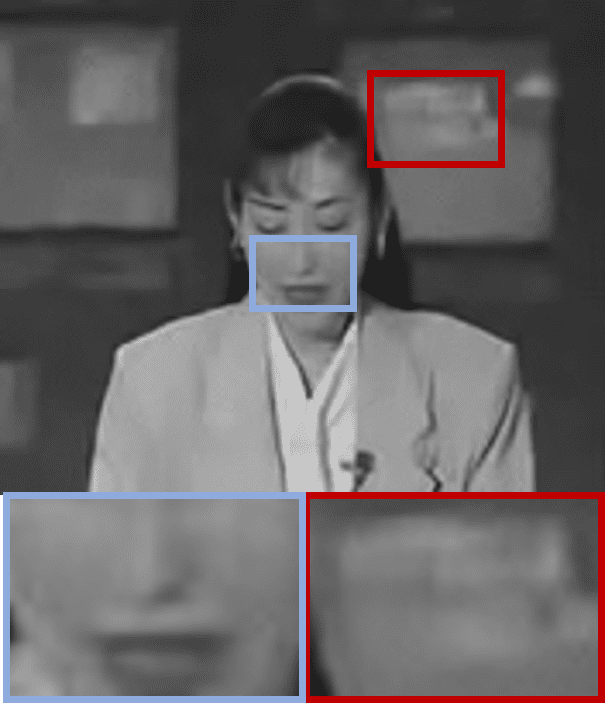} }
	\subfigure[\centering\scriptsize VBM4D(31.77)]   {\includegraphics[width=0.31\linewidth]{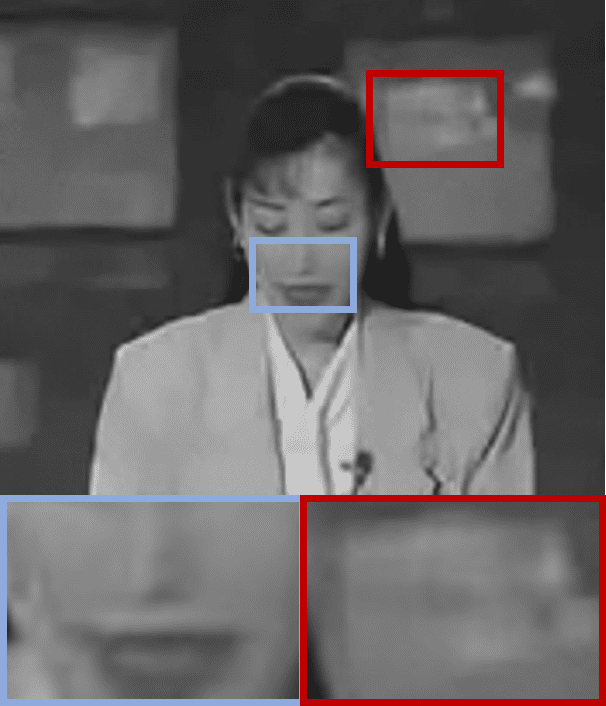} }
	\subfigure[\centering\scriptsize CTV(31.66)]     {\includegraphics[width=0.31\linewidth]{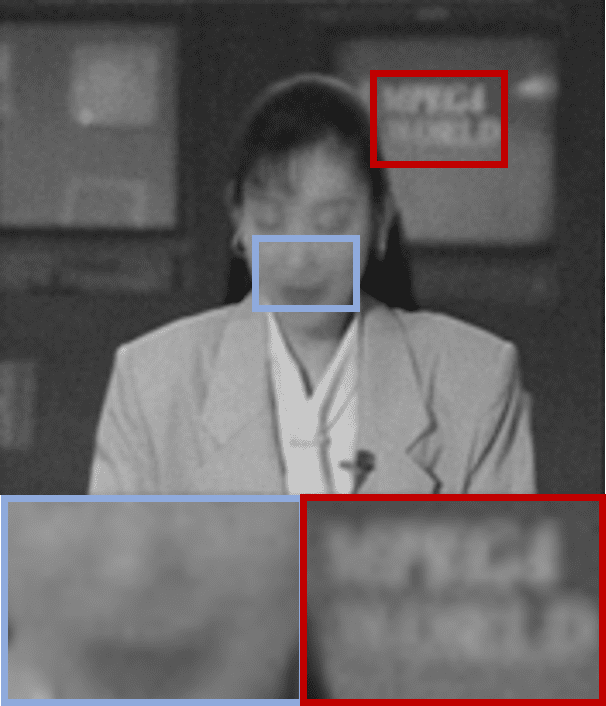} }
	\subfigure[\centering\scriptsize t-CTV(27.34)]   {\includegraphics[width=0.31\linewidth]{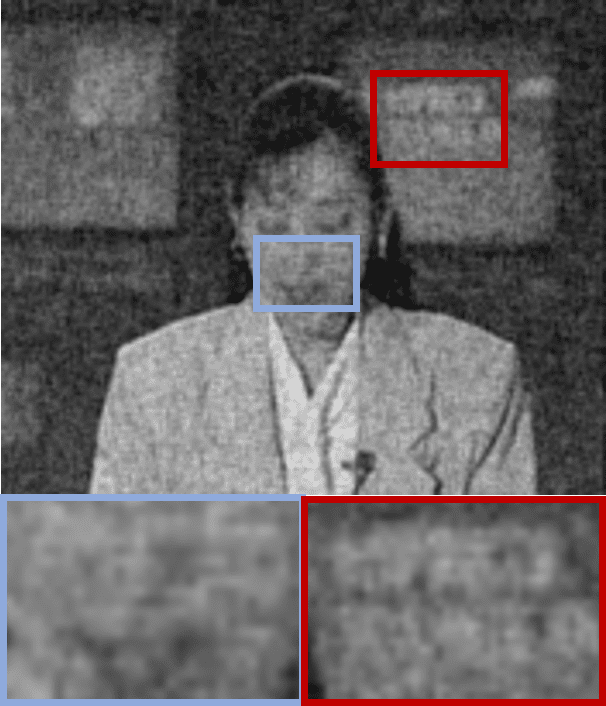} }
	\subfigure[\centering\scriptsize DIP(33.99)]     {\includegraphics[width=0.31\linewidth]{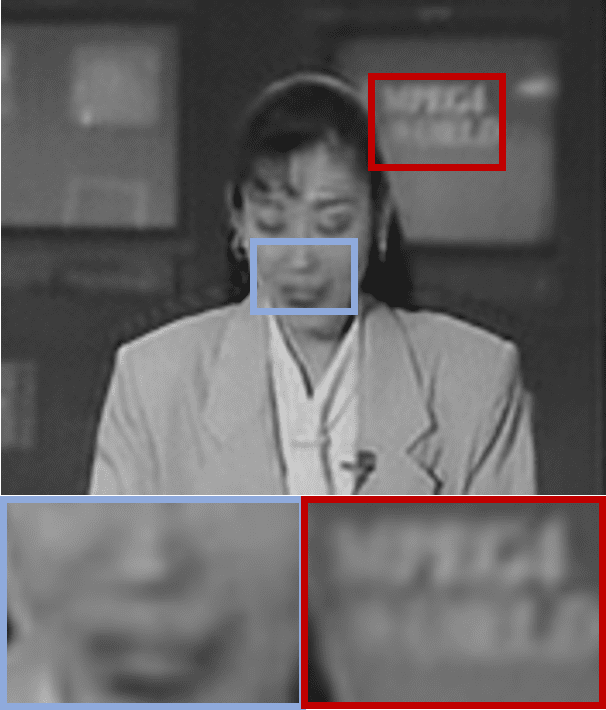} }
	\subfigure[\centering\scriptsize S2DIP(35.18)]   {\includegraphics[width=0.31\linewidth]{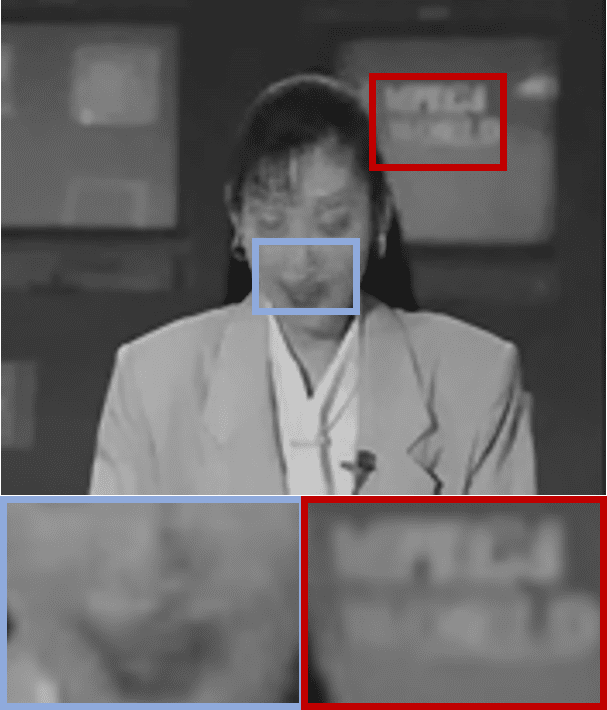} }
	\subfigure[\centering\scriptsize NGR(35.90)]     {\includegraphics[width=0.31\linewidth]{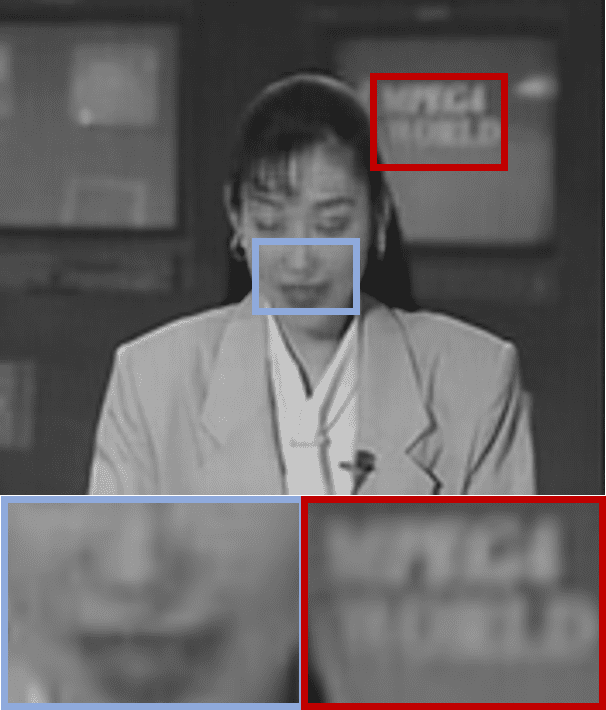} }
	\subfigure[\centering\scriptsize GT]             {\includegraphics[width=0.31\linewidth]{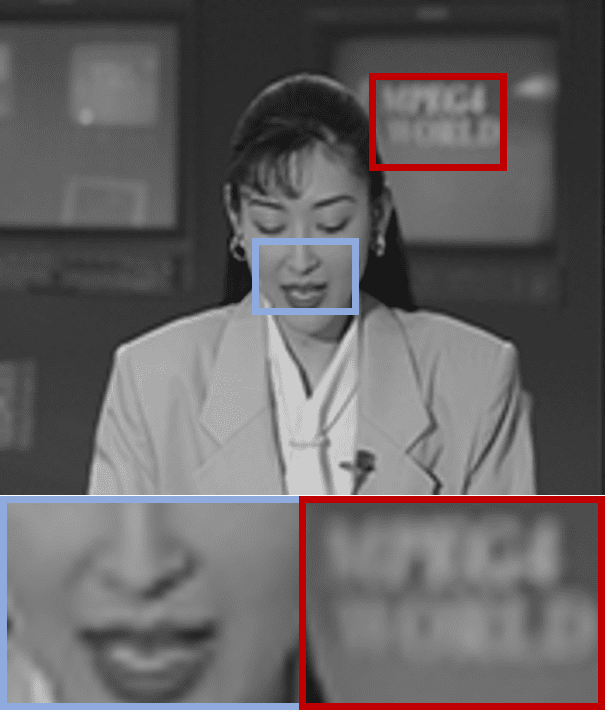} }
	\caption{GT, observed video frame, denoising visual results and  corresponding PSNR values by methods for comparison on akiyo dataset ($\sigma$ = 0.15). The 277-th frame is selected to display.\label{video_denoising fig}}
\end{figure} 
\begin{table}[!t]
	\caption{The quantitive results of videos denoising. The best results are marked in bold.\label{video_denoising table}}
	\centering
	\resizebox{\linewidth}{!}{
		\begin{tabular}{c|cc|cc|cc|cc}
			\Xhline{1.3pt}
			$\sigma$ & \multicolumn{2}{c|}{0.05} & \multicolumn{2}{c|}{0.1} & \multicolumn{2}{c|}{0.15} & \multicolumn{2}{c}{0.2} \\ \hline
			Metrics & PSNR & SSIM & PSNR & SSIM & PSNR & SSIM & PSNR & SSIM \\ \Xhline{0.8pt}
			VBM3D & 38.44  & 0.966  & 34.04  & 0.926  & 30.87  & 0.883  & 28.32  & 0.838  \\ 
			VBM4D & 37.58  & 0.958  & 33.08  & 0.911  & 30.18  & 0.866  & 27.92  & 0.822  \\
			CTV & 32.28  & 0.933  & 31.06  & 0.905  & 30.10  & 0.882  & 29.35  & 0.862  \\ 
			t-CTV & 34.32  & 0.901  & 30.18  & 0.767  & 27.32  & 0.644  & 25.19  & 0.546  \\ 
			DIP & 36.79  & 0.961  & 33.68  & 0.940  & 31.16  & 0.914  & 29.11  & 0.889  \\ 
			S2DIP & 35.45 & 0.957 & 33.83 & 0.940 & \textbf{32.83} & 0.927 & \textbf{31.92} & 0.911 \\ 
			NGR & \textbf{38.53}  & \textbf{0.972}  & \textbf{34.97}  & \textbf{0.951}  & 32.41  & \textbf{0.933}  & 30.22  & \textbf{0.914} \\ \Xhline{1.3pt}
		\end{tabular}
	}
\end{table}
\subsection{Visual data denoising}
For denoising task, we compare our method with TV regularized low-rank matrix factorization (LRTV) \footnote[9]{http://www.lmars.whu.edu.cn/prof\_web/zhanghongyan/Resource
	\%20download.html} \cite{he2015total}, TV regularized low-rank tensor decomposition (LRTDTV) \footnote[10]{https://github.com/zhaoxile/Hyperspectral-Image-Restoration-via-Total-Variation-Regularized-Low-rank-Tensor-Decomposition} \cite{wang2017hyperspectral}, 
correlated TV (CTV) \footnote[11]{https://github.com/andrew-pengjj/ctv\_code} \cite{peng2022exact}, t-CTV \cite{wang2023guaranteed}, fast hyperspectral mixed noise removal (FastHyMix) \footnote[12]{https://github.com/LinaZhuang/HSI-MixedNoiseRemoval-FastHyMix} \cite{zhuang2021fasthymix}, DIP  \cite{ulyanov2018deep, sidorov2019deep}, S2DIP \cite{luo2021hyperspectral} and 3D quasi-recurrent and transformer based network (TRQ3D) \footnote[13]{https://github.com/LiPang/TRQ3DNet}, \cite{pang2022trq3dnet} for HSIs denoising. VBM3D \footnote[14]{https://webpages.tuni.fi/foi/GCF-BM3D/index.html} \cite{kostadin2007video}, VBM4D \cite{maggioni2012video}, CTV, t-CTV, DIP and S2DIP are for videos denoising comparison.

\subsubsection{Videos denoising}
Eight widely used videos are selected for this study. Four levels of Gaussian noise, characterized by $\sigma$ = 0.05, $\sigma$ = 0.1, $\sigma$ = 0.15, and $\sigma$ = 0.2, are considered. 

As depicted in Table \ref{video_denoising table}, the NGR performs superiorly to all comparison methods under low-intensity Gaussian noise conditions. When high-intensity Gaussian noise is present, the NGR achieves the second highest PSNR, slightly lower than S2DIP. Nonetheless, the NGR attains the highest SSIM value in this scenario. The high PSNR values of the S2DIP method on high-intensity Gaussian noise conditions can be attributed to its more sufficient use of proper priors on this data, like local smoothness (by additional TV and SSTV prior terms). However, the NGR's accurate estimation of gradient maps results in higher SSIM values, indicating that the NGR exhibits relatively stronger restoration capabilities in terms of preserving structural information. Furthermore, S2DIP exhibits unsatisfactory performance in the low-intensity Gaussian noise conditions, thereby indicating that NGR should be more robust to noise intensities. 

As illustrated in Fig. \ref{video_denoising fig}, it is evident that VBM3D and VBM4D are incapable of recovering the text on the background. The results obtained by CTV and S2DIP lead to a local blurriness in the foreground. Conversely, NGR exhibits its efficacy in effectively restoring the foreground and background of videos with strong noise, yielding relatively clear restorations. This demonstrates the superior performance of NGR in the video denoising task.

\begin{table*}[!t]
	\caption{The quantitive results of HSIs denoising on PA and WDC. The best results are marked in bold and the second are underlined.\label{HSI_denoising table}}
	\centering
	\begin{tabular}{c|c|cccc|cccc|cccc}
		\Xhline{1.3pt}
		\multirow{2}[0]{*}{Datasets} & Cases & \multicolumn{4}{c|}{Gaussian noise} & \multicolumn{4}{c|}{Weak mixed noise} & \multicolumn{4}{c}{Strong mixed noise} \\ \cline{2-14}
		~ & Metrics & PSNR & SSIM & SAM & ERGAS & PSNR & SSIM & SAM & ERGAS & PSNR & SSIM & SAM & ERGAS \\  \Xhline{0.8pt}
		\multirow{9}[0]{*}{PA} & LRTV & 32.95  & 0.921  & 8.012  & 12.09  & 25.27  & 0.728  & 22.748  & 36.74  & 24.01  & 0.685  & 23.953  & 39.57  \\ 
		~ & LRTDTV & 32.91  & 0.926  & 5.632  & 11.78  & 25.95  & 0.770  & 11.181  & 27.97  & 24.24  & 0.730  & 13.438  & 32.52  \\
		~ & CTV & 33.38  & 0.939  & 5.406  & 11.51  & 27.38  & \underline{0.853}  & \underline{9.709}  & 24.59  & 25.66  & 0.826  & \underline{11.456}  & 28.21  \\
		~ & t-CTV & 29.70  & 0.860  & 11.343  & 17.40  & 22.29  & 0.548  & 23.125  & 40.71  & 20.80  & 0.471  & 26.998  & 47.77  \\ 
		~ & FastHyMix & 35.16  & 0.960  & 4.161  & 9.37  & 25.82  & 0.826  & 10.381  & 29.58  & 24.60  & 0.772  & 13.089  & 35.58  \\ 
		~ & TRQ3D & 32.94  & 0.949  & 5.999  & 12.00  & 27.26  & 0.826  & \textbf{8.143}  & \underline{23.31}  & \textbf{26.99}  & 0.818  & \textbf{8.981}  & \textbf{23.80}  \\
		~ & DIP2D & 32.94  & 0.948  & 5.065  & 11.74  & 26.48  & 0.782  & 13.716  & 29.09  & 24.32  & 0.724  & 17.439  & 35.47  \\ 
		~ & S2DIP & \textbf{35.91}  & \textbf{0.969}  & \textbf{3.805}  & \textbf{8.24}  & \underline{27.90}  & 0.846  & 11.033  & 24.26  & 25.94  & \underline{0.835}  & 14.037  & 28.410  \\ 
		~ & NGR & \underline{35.37}  & \underline{0.969}  & \underline{3.879}  & \underline{9.14}  & \textbf{28.40}  & \textbf{0.880}  & 11.382  & \textbf{23.10}  & \underline{26.37}  & \textbf{0.853} & 13.668  & \underline{26.08}  \\ \hline
		\multirow{9}[0]{*}{WDC} & LRTV & 33.25  & 0.917  & 7.164  & 15.15  & 26.77  & 0.760  & 18.996  & 34.52  & 25.68  & 0.725  & 21.131  & 39.49  \\ 
		~ & LRTDTV & 31.58  & 0.890  & 6.530  & 17.71  & 26.62  & 0.760  & 10.225  & 31.76  & 25.53  & 0.726  & \underline{11.742}  & 36.66  \\ 
		~ & CTV & 31.53  & 0.897  & 9.155  & 19.43  & 28.79  & \underline{0.863}  & \textbf{10.428}  & 25.74  & 27.34  & \underline{0.826}  & 12.554  & \underline{31.33}  \\ 
		~ & t-CTV & 29.95  & 0.847  & 12.216  & 23.25  & 23.71  & 0.595  & 22.759  & 46.26  & 22.39  & 0.534  & 25.627  & 53.03  \\
		~ & FastHyMix & 34.04  & 0.948  & 6.419  & 15.42  & 24.50  & 0.787  & 15.643  & 53.32  & 22.09  & 0.696  & 20.935  & 61.63  \\ 
		~ & TRQ3D & 31.01  & 0.918  & 9.210  & 20.00  & 26.01  & 0.780  & 12.298  & 37.78  & \underline{25.82}  & 0.765  & \textbf{12.491}  & 37.96  \\
		~ & DIP2D & 31.88  & 0.916  & 6.645  & 18.37  & 27.94  & 0.827  & 12.828  & 29.59  & 25.84  & 0.749  & 15.705  & 37.59  \\ 
		~ & S2DIP & \textbf{35.71}  & \textbf{0.960}  & \textbf{5.246}  & \textbf{10.96}  & \underline{28.90}  & 0.845  & \underline{11.088}  & \underline{25.08}  & 27.19  & 0.793  & 13.984  & 31.41  \\ 
		~ & NGR & \underline{34.68}  & \underline{0.955}  & \underline{5.951}  & \underline{14.69}  & \textbf{29.10}  & \textbf{0.878}  & 12.019  & \textbf{24.30}  & \textbf{27.66}  & \textbf{0.839}  & 14.474  & \textbf{29.36} \\ \Xhline{1.3pt}
	\end{tabular}
\end{table*}

\begin{figure*}[t]
	\centering
	\subfigure[\centering\scriptsize Observation]     {\includegraphics[width=0.15\linewidth]{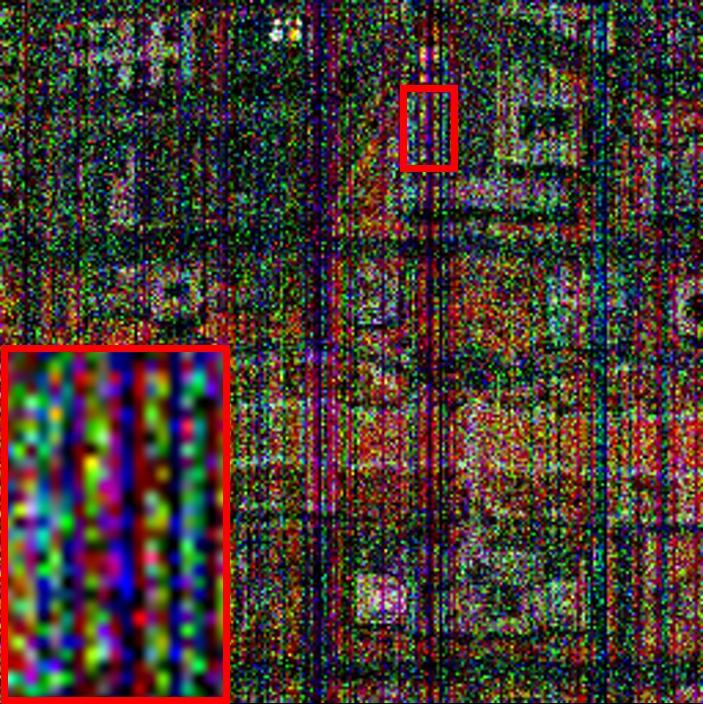} }
	\subfigure[\centering\scriptsize LRTV(25.68)]     {\includegraphics[width=0.15\linewidth]{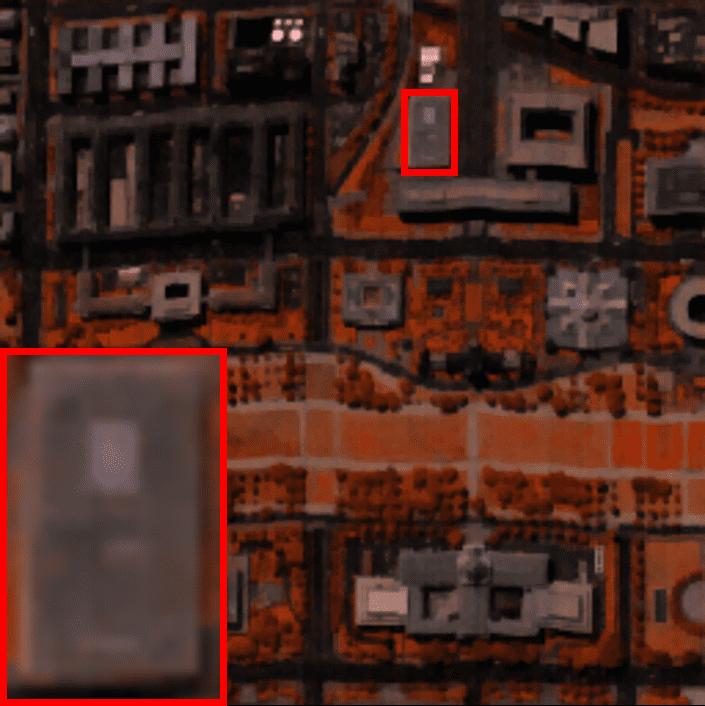} }
	\subfigure[\centering\scriptsize LRTDTV(25.53)]   {\includegraphics[width=0.15\linewidth]{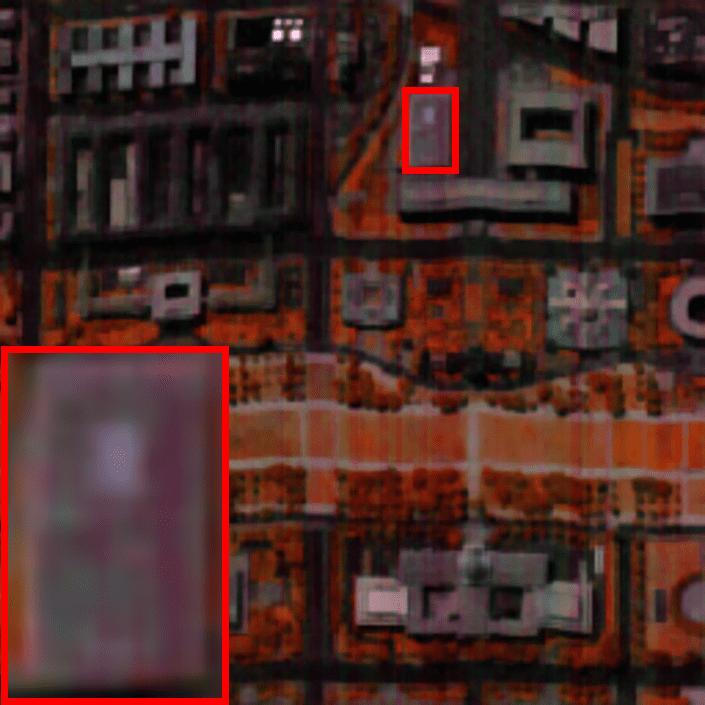} }
	\subfigure[\centering\scriptsize CTV(27.34)]      {\includegraphics[width=0.15\linewidth]{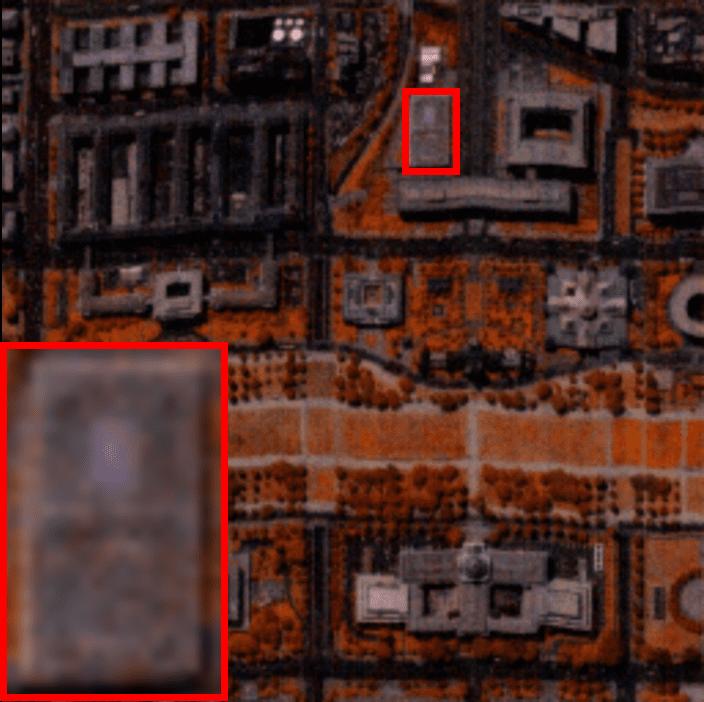} }
	\subfigure[\centering\scriptsize t-CTV(22.39)]    {\includegraphics[width=0.15\linewidth]{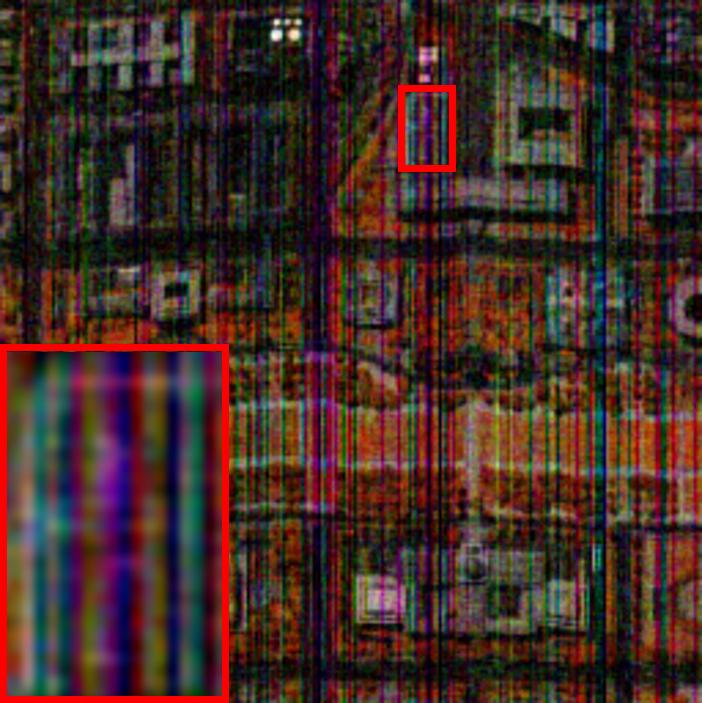} }
	\subfigure[\centering\scriptsize FastHyMix(22.09)]{\includegraphics[width=0.15\linewidth]{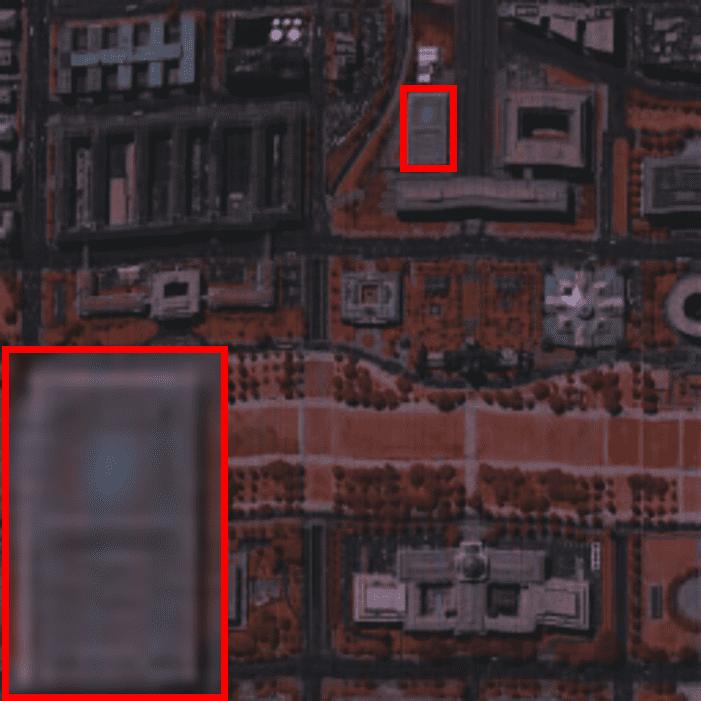} }
	\subfigure[\centering\scriptsize TRQ3D(25.82)]    {\includegraphics[width=0.15\linewidth]{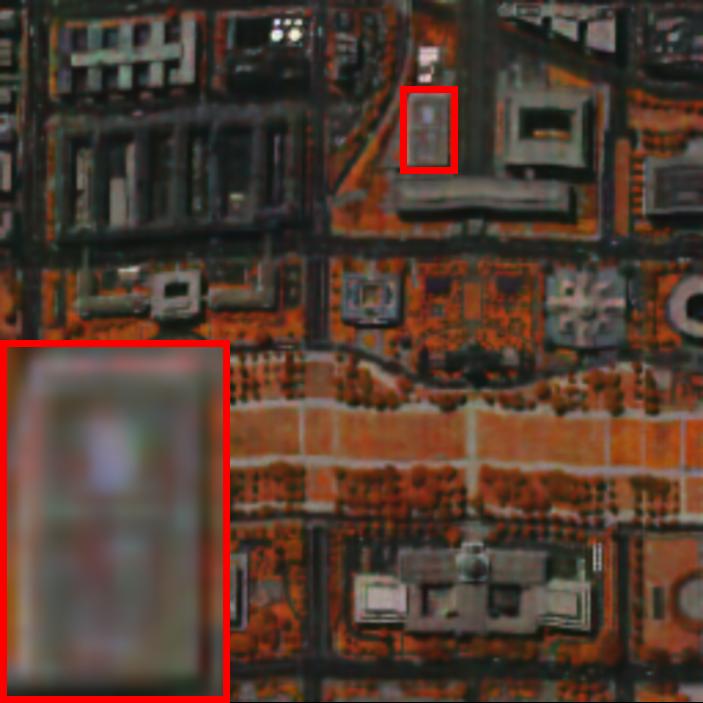} }
	\subfigure[\centering\scriptsize DIP(26.00)]      {\includegraphics[width=0.15\linewidth]{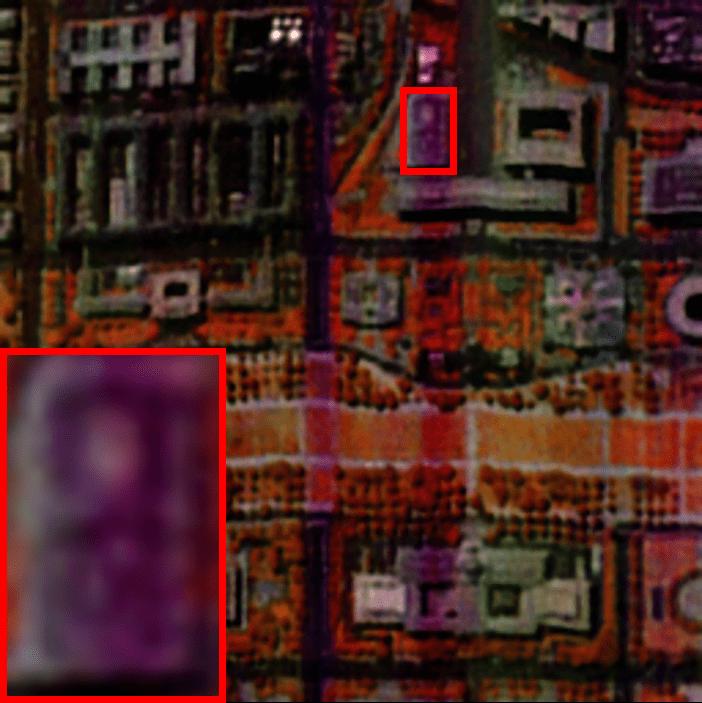} }
	\subfigure[\centering\scriptsize S2DIP(27.19)]    {\includegraphics[width=0.15\linewidth]{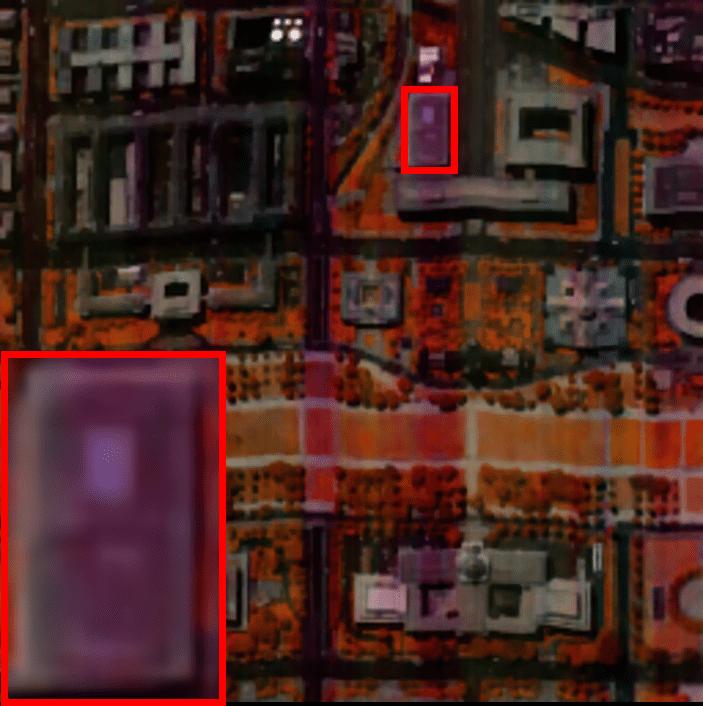} }
	\subfigure[\centering\scriptsize NGR(27.66)]      {\includegraphics[width=0.15\linewidth]{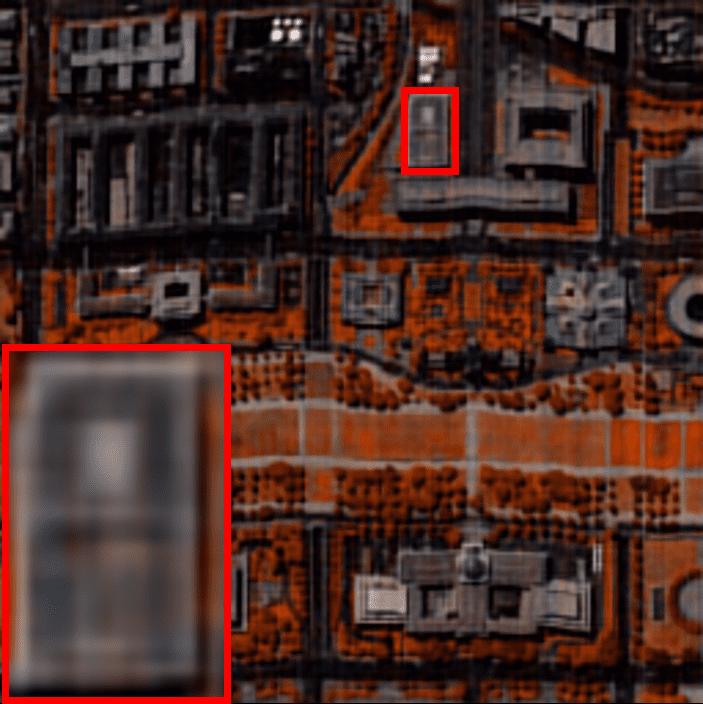} }
	\subfigure[\centering\scriptsize GT]      {\includegraphics[width=0.15\linewidth]{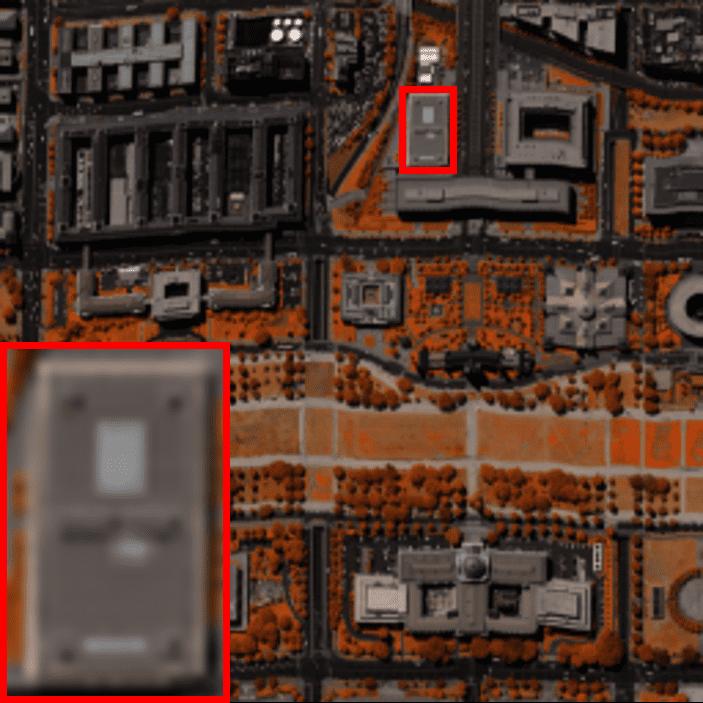} }
	\caption{GT, observed HSI, denoising visual results and corresponding PSNR values by methods for comparison on WDC (strong mixed noise). The pseudo images consisted of the 67-th, 56-th and 45-th bands are selected to display.\label{HSI_denoising fig}}
\end{figure*} 

\begin{figure*}[h]
	\centering
	\subfigure[\centering\scriptsize Observation] {\includegraphics[width=0.15\linewidth]{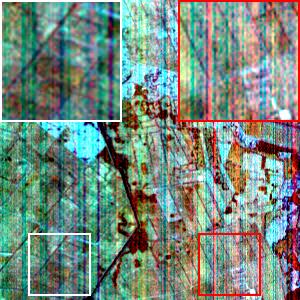} }
	\subfigure[\centering\scriptsize LRTDTV]      {\includegraphics[width=0.15\linewidth]{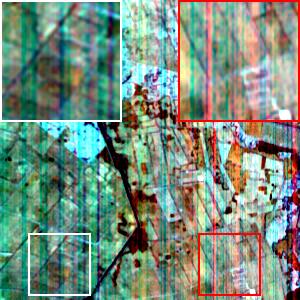} }
	\subfigure[\centering\scriptsize CTV]         {\includegraphics[width=0.15\linewidth]{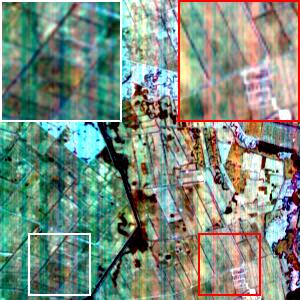} }
	\subfigure[\centering\scriptsize DIP]         {\includegraphics[width=0.15\linewidth]{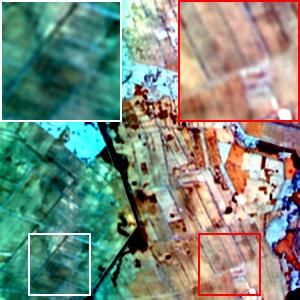} }
	\subfigure[\centering\scriptsize S2DIP]       {\includegraphics[width=0.15\linewidth]{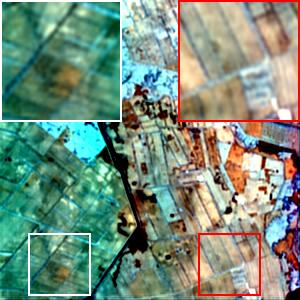} }
	\subfigure[\centering\scriptsize NGR]         {\includegraphics[width=0.15\linewidth]{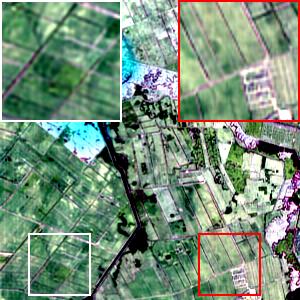} } \\
	\subfigure[\centering\scriptsize Observation] {\includegraphics[width=0.15\linewidth]{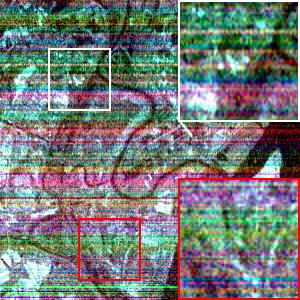} }
	\subfigure[\centering\scriptsize LRTDTV]      {\includegraphics[width=0.15\linewidth]{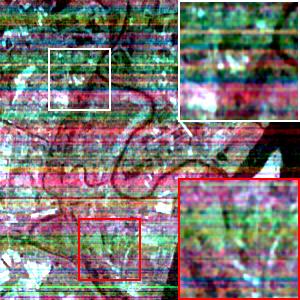} }
	\subfigure[\centering\scriptsize CTV]         {\includegraphics[width=0.15\linewidth]{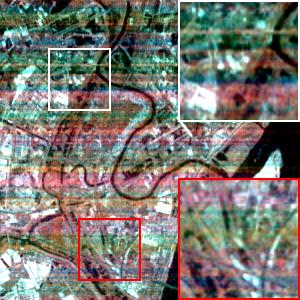} }
	\subfigure[\centering\scriptsize DIP]         {\includegraphics[width=0.15\linewidth]{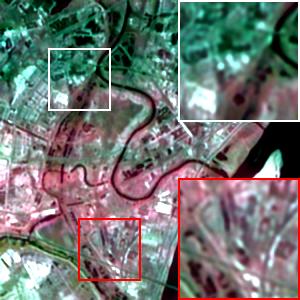} }
	\subfigure[\centering\scriptsize S2DIP]       {\includegraphics[width=0.15\linewidth]{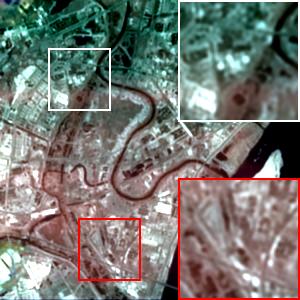} }
	\subfigure[\centering\scriptsize NGR]         {\includegraphics[width=0.15\linewidth]{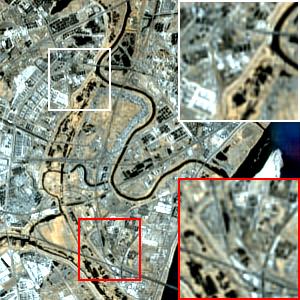} }
	\caption{The visual results on real-world data, Baoqing (the first row) and Wuhan (the second row), obtained by several representative methods. \label{real_hsi_denoising_fig}}
\end{figure*}

\subsubsection{HSIs denoising}
In order to demonstrate the effectiveness of NGR for the HSI denoising task, we also employ the PA and WDC. Considering the intricate nature of noise components in HSIs, we focus on three distinct scenarios for simulated data: (1) Independently and identically distributed (i.i.d.) Gaussian noise. In this scenario, the image cube is contaminated by i.i.d. Gaussian noise with an intensity of $\sigma = 0.1$. (2) Weak mixed noise. Each band is corrupted by Gaussian noise with an intensity randomly selected from 0.1 to 0.4, and impulse noise with a ratio of 0.1. Additionally, 20\% of the bands are corrupted by deadlines and stripes, with the number of each type of noise being 35. (3) Strong mixed noise. The intensity of impulse noise is increased to 0.25, and 50\% of the bands are affected by deadlines and stripes.
Through these simulations, we aim to comprehensively evaluate the performance of NGR in addressing various noise types and intensities commonly encountered in HSIs.

\begin{figure*}[h]
	\centering
	\subfigure[\centering\scriptsize $\bm{\mathcal G}^{(1000)}$]   {\includegraphics[width=0.15\linewidth]{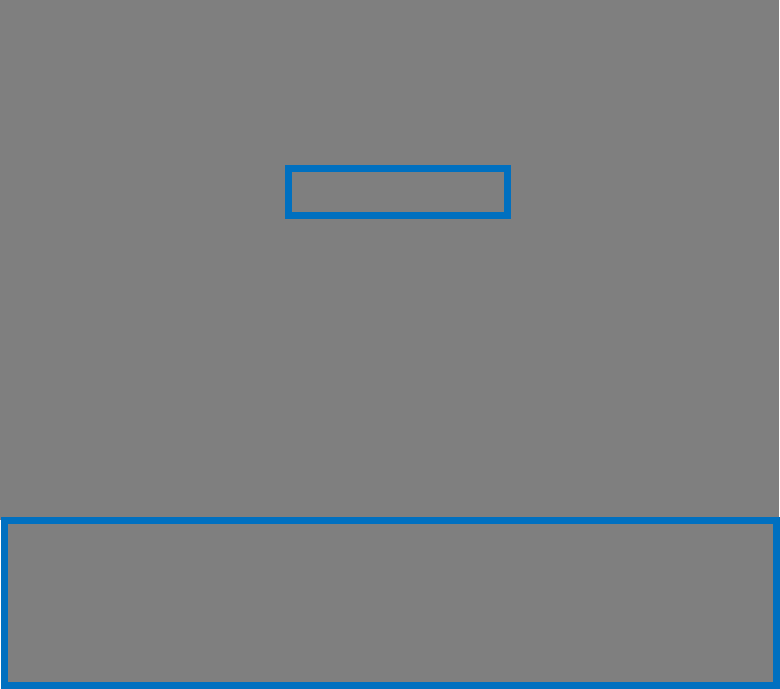} }
	\subfigure[\centering\scriptsize $\bm{\mathcal G}^{(2000)}$]   {\includegraphics[width=0.15\linewidth]{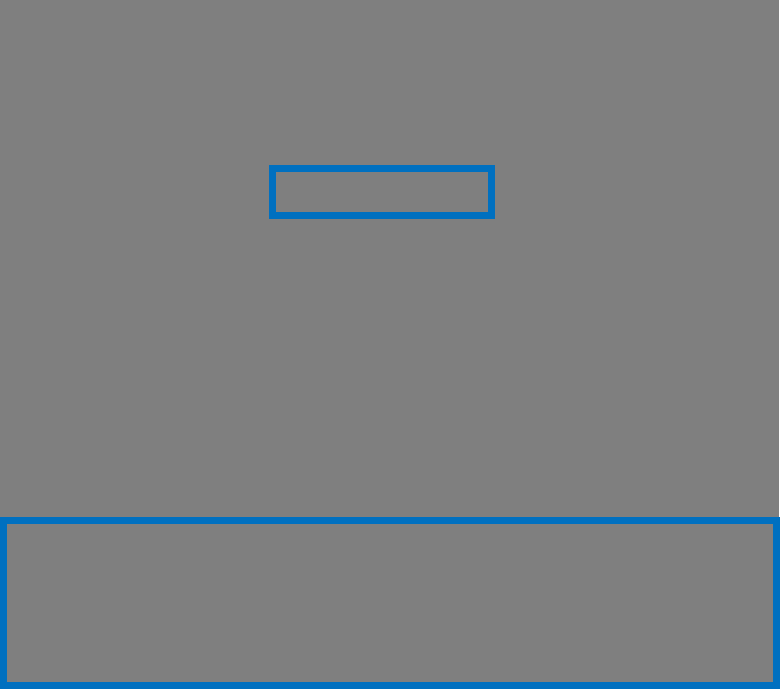} }
	\subfigure[\centering\scriptsize $\bm{\mathcal G}^{(3000)}$]   {\includegraphics[width=0.15\linewidth]{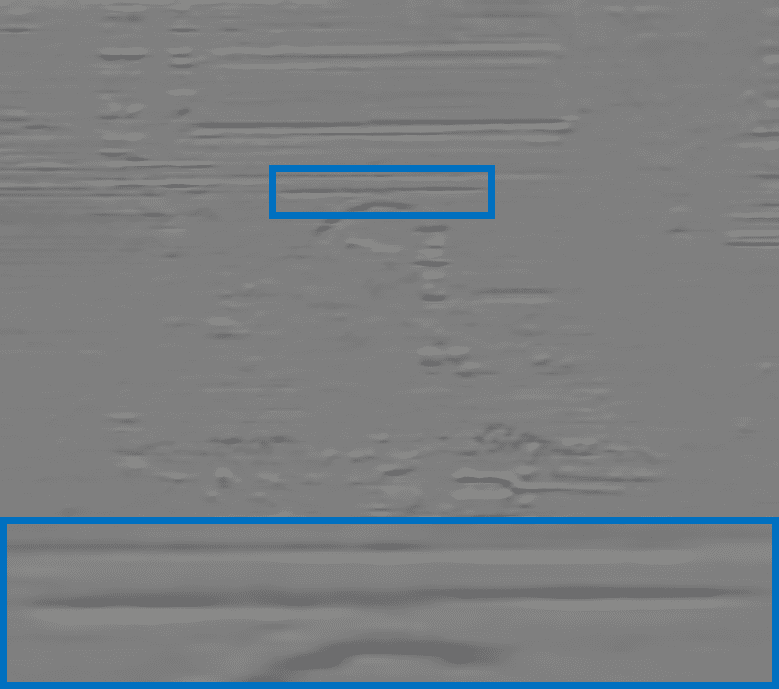} }
	\subfigure[\centering\scriptsize $\bm{\mathcal G}^{(4000)}$]   {\includegraphics[width=0.15\linewidth]{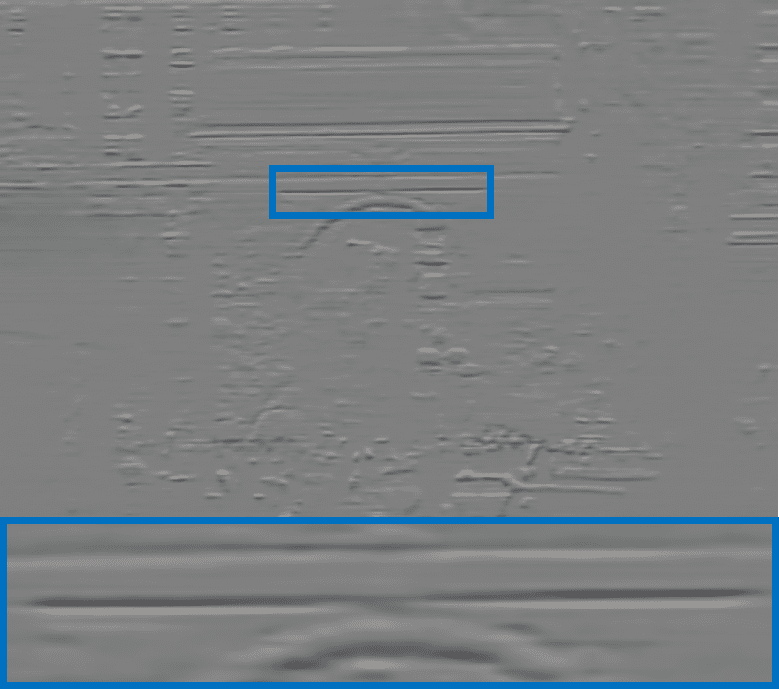} }
	\subfigure[\centering\scriptsize $\bm{\mathcal G}^{(5000)}$]   {\includegraphics[width=0.15\linewidth]{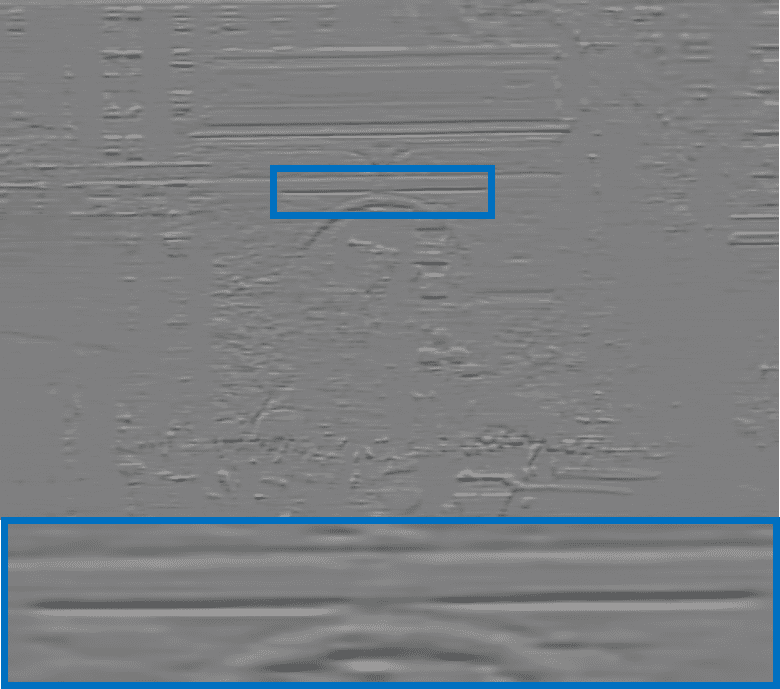} }
	\subfigure[\centering\scriptsize output]{\includegraphics[width=0.15\linewidth]{images/RGB_inpainting/NGR.jpg} }
	\caption{NGR's gradient estimation process on ``291000'' (SR = 10\%) from BSDS100 dataset. (a)-(e) denote the estimated gradient maps on step 1k, 2k, $\dots$, 5k. (f) is the final output of NGR.\label{analysis}}
\end{figure*} 

Quantitative results are presented in Table \ref{HSI_denoising table}. In terms of Gaussian noise, NGR exhibits superior performance in comparison to traditional low-rank based methods. Although the metrics for Gaussian scenarios are slightly lower than those of S2DIP, NGR achieves superior recovery results in most competing cases. Such performance superiority is more evident when dealing with mixed noise than other comparison methods, especially those deep learning ones, like FastHyMix and TRQ3D.. This demonstrates that NGR excels in complex noise removal and accurately restores degraded images. Visual results on the WDC dataset are presented in Fig. \ref{HSI_denoising fig}. These results are consistent with our analysis, indicating that the gradient map estimation provided by NGR aids in the recovery from severe degradation.

In addition to simulated experiments, we also employ two real-world datasets with severe degradation caused by stripes and  deadlines, namely Baoqing and Wuhan, which were both acquired by the GaoFen-5 satellite. Fig. \ref{real_hsi_denoising_fig} illustrates the denoised images obtained by several representative methods. It is observed that both LRTDTV and CTV struggle to handle these dense stripes and deadlines effectively. Although DIP and S2DIP can remove most of the stripes and deadlines, they do so at the expense of evident detail loss and color distortion. Conversely, NGR emerges as the best performer for real-world data, demonstrating its effectiveness in tackling such challenging scenarios.

\subsubsection{Brief summary}
In the context of image denoising, it has been demonstrated that NGR is capable of simultaneously eliminating noise and preserving textures from corrupted images. While S2DIP may achieve better metrics in certain instances, its performance highly relies on the involvement of more priors. Considering that NGR only depends on purely deep gradient prior automatically extracted from data, and the denoised images by NGR can always attain better visual quality, as depicted in the demonstrated figures, it should be rational to say that NGR is effective and potentially useful for more general scenarios.

\subsection{Discussion}

\subsubsection{Analysis of gradient estimation process}\label{sec:gradient_estimation_process}
The robust edge-preserving capability of NGR could possibly be attributed to its efficiency in extracting gradients with strong self-similarity. Leveraging the inductive bias (also known as the deep prior) of neural networks, NGR iteratively refines gradient details, such as edges and textures, enabling it to effectively capture fine-grained gradient maps.

Fig. \ref{analysis} illustrates the gradient estimation process for the same task shown in Fig. \ref{RGB_inpainting fig}. At the $i$-th iteration, based on Eq. \ref{eq:solve_theta}, the deep network can automatically extract fine details from $\bm{\mathcal X}^{(i)}$ and estimate a refined gradient map $\bm{\mathcal G}^{(i+1)}$ by exploiting the strong self-similarity of gradients. By employing the gradient map $\bm{\mathcal G}^{(i+1)}$ estimated by the network, a more refined $\bm{\mathcal X}^{(i+1)}$ tends to be obtained by using Eq. \ref{eq:solve_x}. This iterative process continues until a proper fine-grained gradient map is obtained. This coarse-to-fine gradient estimation process is crucial to the superior performance of NGR.

\subsubsection{Sensitivity to hyperparameters}
We then discuss the effects of hyperparameters on NGR. For the inpainting task on the Set5 dataset (SR = 10\%), these hyperparameters include $\lambda_i (i\in\{h,v,t\})$ and $\mu$ in Eq. \ref{eq:Lagrangian func}. The hyperparameter $\lambda_i$ is designed to control the weights on the $h, v$, and $t$ directions. As aforementioned, we always set $\lambda_h$ and $\lambda_v$ to 1, while regarding $\lambda_t$ as the sole hyperparameter to control the channel direction. $\mu$ is a hyperparameter from the ADMM algorithm, which is used to control the fidelity term.

\begin{figure}[tbp]
	\centering
	\includegraphics[width=7cm]{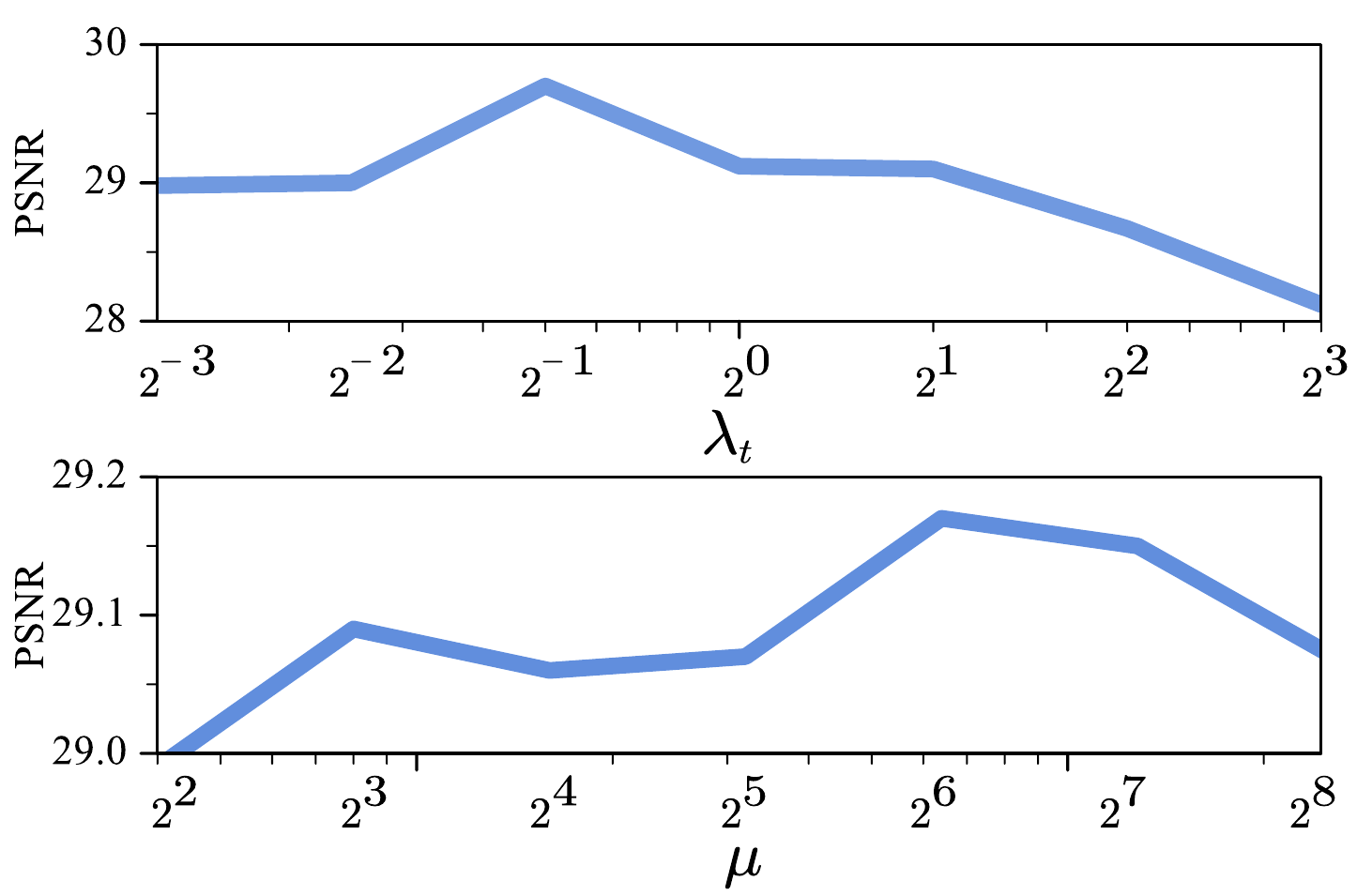} 
	\caption{The quantitative performances of NGR on Set5 dataset (SR = 10\%) with different value of hyperparameters ($\lambda_t$ and $\mu$).\label{hyper}}
\end{figure}

To comprehensively analyze the effects of different hyperparameters on the performance of our method, we vary the value of each hyperparameter while keeping the others fixed, and report the corresponding results. As shown in Fig. \ref{hyper}, we observe that our method is relatively robust, as NGR maintains a good performance across a wide range of hyperparameter values. Notably, for $\mu$, NGR achieves a PSNR higher than 29dB across the range of $\mu$ values from $2^2$ to $2^8$. This demonstrates that NGR is easily applicable in real-world scenarios, making it a more practical and robust choice for inpainting tasks.

\subsubsection{Influences of backbones}\label{backbone}
The gradients estimated by NGR are represented by neural networks $f_{\Theta}(\cdot)$. The experiments conducted above have demonstrated that NGR outperforms other methods based on untrained neural networks (namely, DIP and S2DIP). We now discuss whether NGR exhibits higher robustness to model backbones compared to these methods. In our study, we employed the following networks: ResNet \cite{targ2016resnet}, U-Net \cite{ronneberger2015u}, and a convolutional network composed of ten convolution-batch normalization-ReLU blocks. 

\begin{figure}[tbp]
	\centering
	\includegraphics[width=9cm]{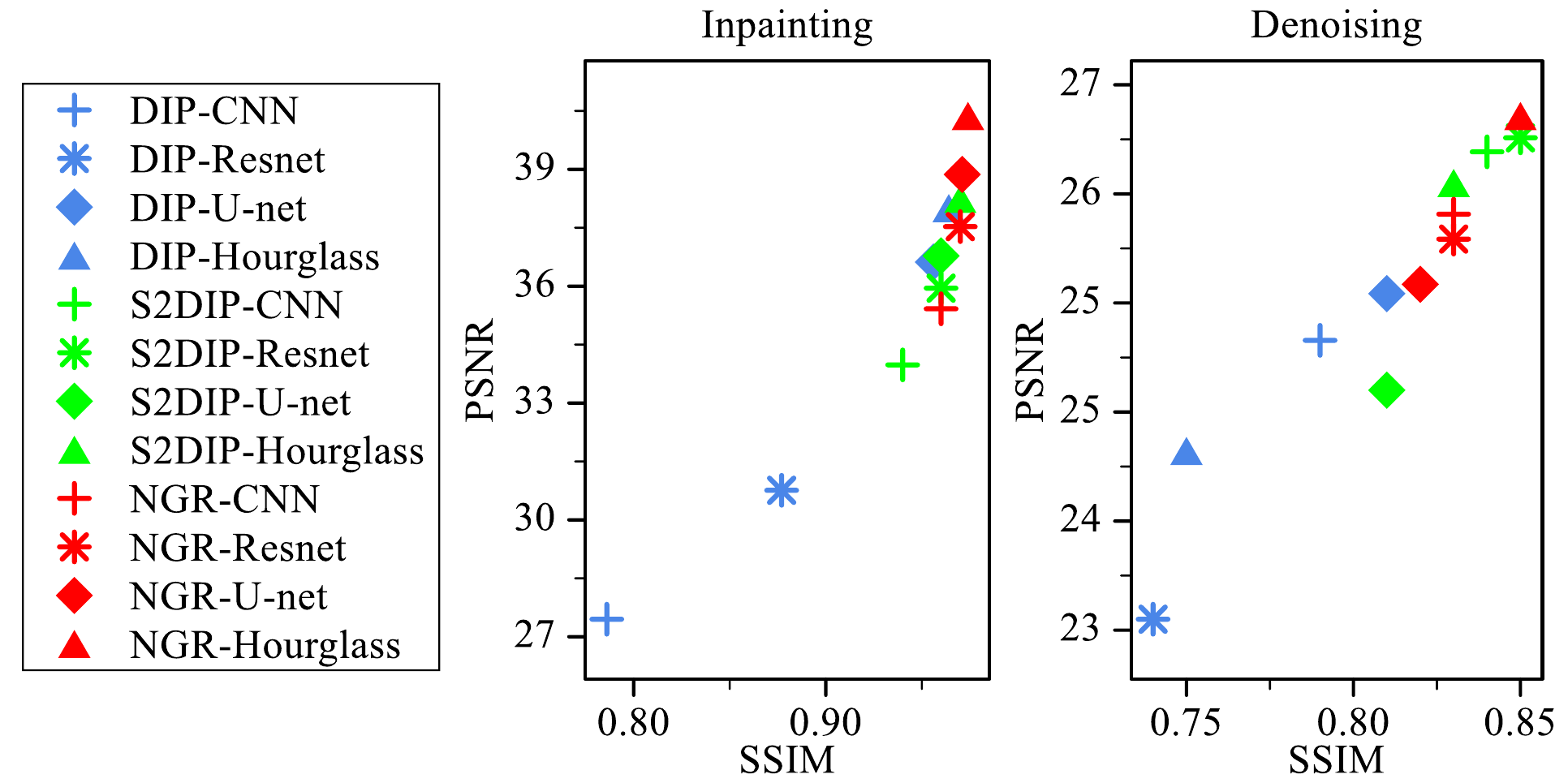} 
	\caption{The quantitative performances of NGR on Set5 dataset (SR = 50\%) for inpainting task and PA (strong mixed noise) for denoising task with different backbone.\label{backbone fig}}
\end{figure}

The results are presented in Fig. \ref{backbone fig}. The performance of DIP marked in blue shows significant variations under different architectures, and in some backbones, it even loses its ability to recover. On the other hand, S2DIP, as a result of incorporating TV, SSTV and DIP, exhibits improved robustness compared to DIP. However, the red markers representing NGR consistently demonstrate strong robustness to different backbone architectures. Regardless of the backbone network used, NGR consistently achieves good performance. This indicates that NGR is capable of adapting and maintaining its effectiveness across different model backbones.

\subsubsection{Comparison with other regularizers}
To demonstrate the advantage over other regularizers, we apply TV, TGV, HOTV, TV$_{l_p}(p=1/2)$, CTV, and t-CTV to the PA dataset with Case 3, which represents one of the most challenging scenario for the HSI denoising task. For fair comparison, all regularizers are equipped with $\ell_{1}$-norm based data fidelity, and the hyperparameters are tuned by grid search. TV, TGV, HOTV and TV$_{l_p}(p=1/2)$ belong to the sparse gradient prior category, but Table \ref{tab:compare_other_tv} shows that these TV variants do not result in significant improvements over the vanilla TV. CTV fuses sparsity and low-rank priors, achieving a 5.13dB gain over TV in terms of PSNR. This gain stems from CTV taking into account the additional prior knowledge (i.e., low-rank). The proposed NGR further improves over CTV by 0.71dB using the gradient estimation technique. In the future, it is promising to investigate the combination of sparse or low-rank prior with NGR, which may lead to even better results.

\begin{table}[tbp]
	\centering
	\caption{Metrics of different regularizers on the PA dataset with Case 3. LR is the abbreviation of low-rank. Grad. Esit. is the abbreviation of gradient estimation.}
	\resizebox{\linewidth}{!}{
		\begin{tabular}{c|cccc|ccc}
			\Xhline{1.3pt}
			& PSNR  & SSIM  & SAM   & ERGAS & Sparse & LR & Grad. Esti. \\
			\Xhline{0.8pt}
			TV    & 20.53  & 0.375  & 16.264  & 49.02  & $\checkmark$ &       &  \\
			TV$_{l_{p}}$ & 20.22  & 0.327  & 20.229  & 49.20  & $\checkmark$ &       &  \\ 
			TGV   & 20.88  & 0.434  & 16.850  & 47.45  & $\checkmark$ &       &  \\
			HOTV  & 20.96  & 0.438  & 16.653  & 46.94  & $\checkmark$ &       &  \\
			CTV   & 25.66 & 0.826 & \textbf{11.456} & 28.21 & $\checkmark$ & $\checkmark$ &  \\
			NGR   & \textbf{26.37} & \textbf{0.853} & 13.668 & \textbf{26.08} &       &       & $\checkmark$ \\
			\Xhline{1.3pt}
		\end{tabular}%
	}
	\label{tab:compare_other_tv}%
\end{table}%
\begin{figure}[tbp]
	\centering
	\includegraphics[width=9cm]{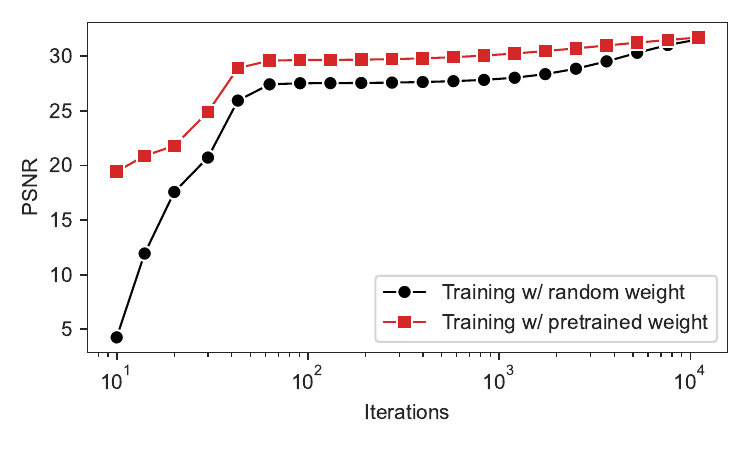} 
	\caption{The comparison of PSNR curves versus iteration for NGR with different initializations.  \label{fig:psnr_vs_iteration_two_models}}
\end{figure} 

\subsubsection{Transferring capability}
An intriguing inquiry pertains to the feasibility of transferring learned knowledge from one image to other images. As is well established, the learned knowledge of neural networks can be inherently embedded within the network weights. Consequently, an experimental investigation was conducted where the NGR model was trained with distinct initializations. In the first trial, the weights were initialized randomly, and the NGR was directly trained on the test image. In contrast, the second trial used the pretrained weights on another image, followed by finetuning the NGR on the test image.

The PSNR curves visualized in Fig. \ref{fig:psnr_vs_iteration_two_models} reveal that pretrained weights facilitate the attainment of higher PSNR values than random weights, with the gap diminishing after 8000 iterations. Ultimately, after 11000 iterations, the PSNR values for pretrained and random weights reached 31.71dB and 31.54dB, respectively. This outcome corroborates that utilizing pretrained weights yields superior performance.

In summary, this experiment unequivocally demonstrates that the  knowledge learned by NGR from one image can indeed be transferred and applied to other images.

\section{Conclusion}\label{sec:Conclusion}
In this research, we have proposed a novel approach for gradient prior modeling, referred to as the neural gradient regularizer (NGR). Distinct from conventional manually pre-designed sparse gradient priors, such as total variation and its variants, NGR does not depend on the gradient sparsity assumption. Instead, it is capable of automatically extract intrinsic priors underlying the gradient maps in an entirely data-driven and zero-shot learning manner. Especially, attributed to its representation by a neural network, more flexible and complex priors of gradient images, rather than only sparsity prior as conventional TV methods, are expected to be derived from data and help further enhance the performance of the image restoration task. Our comprehensive experimental evaluation demonstrates that the versatile NGR is applicable to a wide range of image processing tasks and data types, exhibiting superior performance compared to state-of-the-art methods. The effectiveness of the proposed method can thus been substantiated.

\bibliographystyle{IEEEtran}
\bibliography{ngr_ref}

\begin{thebibliography}{10}
\providecommand{\url}[1]{#1}
\csname url@samestyle\endcsname
\providecommand{\newblock}{\relax}
\providecommand{\bibinfo}[2]{#2}
\providecommand{\BIBentrySTDinterwordspacing}{\spaceskip=0pt\relax}
\providecommand{\BIBentryALTinterwordstretchfactor}{4}
\providecommand{\BIBentryALTinterwordspacing}{\spaceskip=\fontdimen2\font plus
\BIBentryALTinterwordstretchfactor\fontdimen3\font minus
  \fontdimen4\font\relax}
\providecommand{\BIBforeignlanguage}[2]{{%
\expandafter\ifx\csname l@#1\endcsname\relax
\typeout{** WARNING: IEEEtran.bst: No hyphenation pattern has been}%
\typeout{** loaded for the language `#1'. Using the pattern for}%
\typeout{** the default language instead.}%
\else
\language=\csname l@#1\endcsname
\fi
#2}}
\providecommand{\BIBdecl}{\relax}
\BIBdecl

\bibitem{Microscopy_Images_Denoising_Review}
W.~Meiniel, J.~Olivo{-}Marin, and E.~D. Angelini, ``Denoising of microscopy
  images: {A} review of the state-of-the-art, and a new sparsity-based
  method,'' \emph{{IEEE} Trans. Image Process.}, vol.~27, no.~8, pp.
  3842--3856, 2018.

\bibitem{LRLRR}
Z.~Zhang and K.~Zhao, ``Low-rank matrix approximation with manifold
  regularization,'' \emph{{IEEE} Trans. Pattern Anal. Mach. Intell.}, vol.~35,
  no.~7, pp. 1717--1729, 2013.

\bibitem{DCP}
K.~He, J.~Sun, and X.~Tang, ``Single image haze removal using dark channel
  prior,'' \emph{{IEEE} Trans. Pattern Anal. Mach. Intell.}, vol.~33, no.~12,
  pp. 2341--2353, 2011.

\bibitem{ROP}
J.~Liu, R.~W. Liu, J.~Sun, and T.~Zeng, ``Rank-one prior: Real-time scene
  recovery,'' \emph{{IEEE} Trans. Pattern Anal. Mach. Intell.}, vol.~45, no.~7,
  pp. 8845--8860, 2023.

\bibitem{PanHS017}
J.~Pan, Z.~Hu, Z.~Su, and M.~Yang, ``L\({}_{\mbox{0}}\)-regularized intensity
  and gradient prior for deblurring text images and beyond,'' \emph{{IEEE}
  Trans. Pattern Anal. Mach. Intell.}, vol.~39, no.~2, pp. 342--355, 2017.

\bibitem{ZhengLKXGK13}
Y.~Zheng, S.~Lin, S.~B. Kang, R.~Xiao, J.~C. Gee, and C.~Kambhamettu,
  ``Single-image vignetting correction from gradient distribution symmetries,''
  \emph{{IEEE} Trans. Pattern Anal. Mach. Intell.}, vol.~35, no.~6, pp.
  1480--1494, 2013.

\bibitem{Gradient_Profile_Prior}
J.~Sun, J.~Sun, Z.~Xu, and H.~Shum, ``Gradient profile prior and its
  applications in image super-resolution and enhancement,'' \emph{{IEEE} Trans.
  Image Process.}, vol.~20, no.~6, pp. 1529--1542, 2011.

\bibitem{Learn_Gradient_Profile}
C.~Ren, X.~He, Y.~Pu, and T.~Q. Nguyen, ``Learning image profile enhancement
  and denoising statistics priors for single-image super-resolution,''
  \emph{{IEEE} Trans. Cybern.}, vol.~51, no.~7, pp. 3535--3548, 2021.

\bibitem{total_variation_review}
T.~F. Chan, S.~Esedoglu, F.~E. Park, and A.~M. Yip, ``Total variation image
  restoration: Overview and recent developments,'' in \emph{Handbook of
  Mathematical Models in Computer Vision}, N.~Paragios, Y.~Chen, and O.~D.
  Faugeras, Eds.\hskip 1em plus 0.5em minus 0.4em\relax Springer, 2006, pp.
  17--31.

\bibitem{TV_deblurring}
W.~Dong, S.~Tao, G.~Xu, and Y.~Chen, ``Blind deconvolution for poissonian
  blurred image with total variation and $\ell_{0}$-norm gradient
  regularizations,'' \emph{{IEEE} Trans. Image Process.}, vol.~30, pp.
  1030--1043, 2021.

\bibitem{TV_inpainting}
M.~V. Afonso and J.~M.~R. Sanches, ``Blind inpainting using $\ell_{0}$ and
  total variation regularization,'' \emph{{IEEE} Trans. Image Process.},
  vol.~24, no.~7, pp. 2239--2253, 2015.

\bibitem{he2015total}
W.~He, H.~Zhang, L.~Zhang, and H.~Shen, ``Total-variation-regularized low-rank
  matrix factorization for hyperspectral image restoration,'' \emph{{IEEE}
  Trans. Geosci. Remote. Sens.}, vol.~54, no.~1, pp. 178--188, 2016.

\bibitem{NonConvexTV}
D.~Krishnan and R.~Fergus, ``Fast image deconvolution using hyper-laplacian
  priors,'' in \emph{Advances in Neural Information Processing Systems, 7-10
  December 2009, Vancouver, British Columbia, Canada}, 2009, pp. 1033--1041.

\bibitem{NonConvexTV2015}
I.~W. Selesnick, A.~Parekh, and I.~Bayram, ``Convex 1-d total variation
  denoising with non-convex regularization,'' \emph{{IEEE} Signal Process.
  Lett.}, vol.~22, no.~2, pp. 141--144, 2015.

\bibitem{Nonconvex_Hybrid_Total_Variation}
Y.~Sun, L.~Lei, D.~Guan, X.~Li, and G.~Kuang, ``{SAR} image speckle reduction
  based on nonconvex hybrid total variation model,'' \emph{{IEEE} Trans.
  Geosci. Remote. Sens.}, vol.~59, no.~2, pp. 1231--1249, 2021.

\bibitem{LDCKTV_IEEESPL_2021}
J.~Zhuang, Y.~Luo, X.~Zhao, and T.~Jiang, ``Reconciling hand-crafted and
  self-supervised deep priors for video directional rain streaks removal,''
  \emph{{IEEE} Signal Process. Lett.}, vol.~28, pp. 2147--2151, 2021.

\bibitem{High_Order_Directional_Total_Variation}
X.~Liu, Q.~Li, C.~Yuan, J.~Li, X.~Chen, and Y.~Chen, ``High-order directional
  total variation for seismic noise attenuation,'' \emph{{IEEE} Trans. Geosci.
  Remote. Sens.}, vol.~60, p. 5903013, 2022.

\bibitem{DTV_IEEESPL_2012}
I.~Bayram and M.~E. Kamasak, ``Directional total variation,'' \emph{{IEEE}
  Signal Process. Lett.}, vol.~19, no.~12, pp. 781--784, 2012.

\bibitem{E3DTV}
J.~Peng, Q.~Xie, Q.~Zhao, Y.~Wang, Y.~Leung, and D.~Meng, ``Enhanced 3dtv
  regularization and its applications on {HSI} denoising and compressed
  sensing,'' \emph{{IEEE} Trans. Image Process.}, vol.~29, pp. 7889--7903,
  2020.

\bibitem{peng2022exact}
J.~Peng, Y.~Wang, H.~Zhang, J.~Wang, and D.~Meng, ``Exact decomposition of
  joint low rankness and local smoothness plus sparse matrices,'' \emph{{IEEE}
  Trans. Pattern Anal. Mach. Intell.}, vol.~45, no.~5, pp. 5766--5781, 2023.

\bibitem{wang2023guaranteed}
H.~Wang, J.~Peng, W.~Qin, J.~Wang, and D.~Meng, ``Guaranteed tensor recovery
  fused low-rankness and smoothness,'' \emph{{IEEE} Trans. Pattern Anal. Mach.
  Intell.}, vol.~45, no.~9, pp. 10\,990--11\,007, 2023.

\bibitem{TV_HOTV}
M.~Lysaker and X.~Tai, ``Iterative image restoration combining total variation
  minimization and a second-order functional,'' \emph{Int. J. Comput. Vis.},
  vol.~66, no.~1, pp. 5--18, 2006.

\bibitem{Second_order_TV_SIAMJSC_2000}
T.~F. Chan, A.~Marquina, and P.~Mulet, ``High-order total variation-based image
  restoration,'' \emph{{SIAM} J. Sci. Comput.}, vol.~22, no.~2, pp. 503--516,
  2000.

\bibitem{Arbitrary_order_TV_PR_2023}
J.~Duan, X.~Jia, J.~Bartlett, W.~Lu, and Z.~Qiu, ``Arbitrary order total
  variation for deformable image registration,'' \emph{Pattern Recognit.}, vol.
  137, p. 109318, 2023.

\bibitem{TGV}
K.~Bredies, K.~Kunisch, and T.~Pock, ``Total generalized variation,''
  \emph{{SIAM} J. Imaging Sci.}, vol.~3, no.~3, pp. 492--526, 2010.

\bibitem{TGV_piecewise_constant_func}
L.~Baumg{\"{a}}rtner, R.~Bergmann, R.~Herzog, S.~Schmidt, and
  J.~Vidal{-}N{\'{u}}{\~{n}}ez, ``Total generalized variation for piecewise
  constant functions on triangular meshes with applications in imaging,''
  \emph{{SIAM} J. Imaging Sci.}, vol.~16, no.~1, pp. 313--339, 2023.

\bibitem{TDV}
E.~Kobler, A.~Effland, K.~Kunisch, and T.~Pock, ``Total deep variation for
  linear inverse problems,'' in \emph{Computer Vision and Pattern Recognition,
  {CVPR}, Seattle, WA, USA, June 13-19, 2020}, 2020, pp. 7546--7555.

\bibitem{TDV2}
------, ``Total deep variation: {A} stable regularization method for inverse
  problems,'' \emph{{IEEE} Trans. Pattern Anal. Mach. Intell.}, vol.~44,
  no.~12, pp. 9163--9180, 2022.

\bibitem{ulyanov2018deep}
D.~Ulyanov, A.~Vedaldi, and V.~S. Lempitsky, ``Deep image prior,'' in
  \emph{Computer Vision and Pattern Recognition, {CVPR}, Salt Lake City, UT,
  USA, June 18-22, 2018}, 2018, pp. 9446--9454.

\bibitem{DIP_IJCV}
------, ``Deep image prior,'' \emph{Int. J. Comput. Vis.}, vol. 128, no.~7, pp.
  1867--1888, 2020.

\bibitem{spectral_bias_dip}
P.~Chakrabarty and S.~Maji, ``The spectral bias of the deep image prior,'' in
  \emph{Advances in Neural Information Processing Systems Workshops, Vancouver,
  Canada, December 8-14, 2019}, 2019.

\bibitem{SB_DIP}
Z.~Shi, P.~Mettes, S.~Maji, and C.~G.~M. Snoek, ``On measuring and controlling
  the spectral bias of the deep image prior,'' \emph{Int. J. Comput. Vis.},
  vol. 130, no.~4, pp. 885--908, 2022.

\bibitem{liu2012tensor}
J.~Liu, P.~Musialski, P.~Wonka, and J.~Ye, ``Tensor completion for estimating
  missing values in visual data,'' \emph{{IEEE} Trans. Pattern Anal. Mach.
  Intell.}, vol.~35, no.~1, pp. 208--220, 2013.

\bibitem{yokota2016smooth}
T.~Yokota, Q.~Zhao, and A.~Cichocki, ``Smooth {PARAFAC} decomposition for
  tensor completion,'' \emph{{IEEE} Trans. Signal Process.}, vol.~64, no.~20,
  pp. 5423--5436, 2016.

\bibitem{lu2018exact}
C.~Lu, J.~Feng, Z.~Lin, and S.~Yan, ``Exact low tubal rank tensor recovery from
  gaussian measurements,'' in \emph{International Joint Conference on
  Artificial Intelligence, {IJCAI}, July 13-19, 2018, Stockholm, Sweden},
  J.~Lang, Ed., 2018, pp. 2504--2510.

\bibitem{lu2019low}
C.~Lu, X.~Peng, and Y.~Wei, ``Low-rank tensor completion with a new tensor
  nuclear norm induced by invertible linear transforms,'' in \emph{Computer
  Vision and Pattern Recognition, {CVPR}, Long Beach, CA, USA, June 16-20,
  2019}, 2019, pp. 5996--6004.

\bibitem{ji2016tensor}
T.~Ji, T.~Huang, X.~Zhao, T.~Ma, and G.~Liu, ``Tensor completion using total
  variation and low-rank matrix factorization,'' \emph{Inf. Sci.}, vol. 326,
  pp. 243--257, 2016.

\bibitem{sidorov2019deep}
O.~Sidorov and J.~Y. Hardeberg, ``Deep hyperspectral prior: Single-image
  denoising, inpainting, super-resolution,'' in \emph{International Conference
  on Computer Vision Workshops, Seoul, Korea (South), October 27-28,
  2019}.\hskip 1em plus 0.5em minus 0.4em\relax {IEEE}, 2019, pp. 3844--3851.

\bibitem{luo2021hyperspectral}
Y.~Luo, X.~Zhao, T.~Jiang, Y.~Zheng, and Y.~Chang, ``Hyperspectral mixed noise
  removal via spatial-spectral constrained unsupervised deep image prior,''
  \emph{{IEEE} J. Sel. Top. Appl. Earth Obs. Remote. Sens.}, vol.~14, pp.
  9435--9449, 2021.

\bibitem{martin2001database}
D.~R. Martin, C.~C. Fowlkes, D.~Tal, and J.~Malik, ``A database of human
  segmented natural images and its application to evaluating segmentation
  algorithms and measuring ecological statistics,'' in \emph{International
  Conference On Computer Vision, Vancouver, British Columbia, Canada, July
  7-14, 2001}, 2001, pp. 416--425.

\bibitem{bevilacqua2012low}
M.~Bevilacqua, A.~Roumy, C.~Guillemot, and M.~Alberi{-}Morel, ``Low-complexity
  single-image super-resolution based on nonnegative neighbor embedding,'' in
  \emph{British Machine Vision Conference, Surrey, UK, September 3-7, 2012},
  R.~Bowden, J.~P. Collomosse, and K.~Mikolajczyk, Eds.\hskip 1em plus 0.5em
  minus 0.4em\relax {BMVA} Press, 2012, pp. 1--10.

\bibitem{ji2020simultaneous}
S.~Ji, P.~Dai, M.~Lu, and Y.~Zhang, ``Simultaneous cloud detection and removal
  from bitemporal remote sensing images using cascade convolutional neural
  networks,'' \emph{{IEEE} Trans. Geosci. Remote. Sens.}, vol.~59, no.~1, pp.
  732--748, 2021.

\bibitem{wang2017hyperspectral}
Y.~Wang, J.~Peng, Q.~Zhao, Y.~Leung, X.~Zhao, and D.~Meng, ``Hyperspectral
  image restoration via total variation regularized low-rank tensor
  decomposition,'' \emph{{IEEE} J. Sel. Top. Appl. Earth Obs. Remote. Sens.},
  vol.~11, no.~4, pp. 1227--1243, 2018.

\bibitem{zhuang2021fasthymix}
L.~Zhuang and M.~K. Ng, ``Fasthymix: Fast and parameter-free hyperspectral
  image mixed noise removal,'' \emph{{IEEE} Trans. Neural Networks Learn.
  Syst.}, vol.~34, no.~8, pp. 4702--4716, 2023.

\bibitem{pang2022trq3dnet}
L.~Pang, W.~Gu, and X.~Cao, ``Trq3dnet: {A} 3d quasi-recurrent and transformer
  based network for hyperspectral image denoising,'' \emph{Remote. Sens.},
  vol.~14, no.~18, p. 4598, 2022.

\bibitem{kostadin2007video}
K.~Dabov, A.~Foi, and K.~O. Egiazarian, ``Video denoising by sparse 3d
  transform-domain collaborative filtering,'' in \emph{European Signal
  Processing Conference, Poznan, Poland, September 3-7, 2007}.\hskip 1em plus
  0.5em minus 0.4em\relax {IEEE}, 2007, pp. 145--149.

\bibitem{maggioni2012video}
M.~Maggioni, G.~Boracchi, A.~Foi, and K.~O. Egiazarian, ``Video denoising,
  deblocking, and enhancement through separable 4-d nonlocal spatiotemporal
  transforms,'' \emph{{IEEE} Trans. Image Process.}, vol.~21, no.~9, pp.
  3952--3966, 2012.

\bibitem{targ2016resnet}
K.~He, X.~Zhang, S.~Ren, and J.~Sun, ``Deep residual learning for image
  recognition,'' in \emph{Computer Vision and Pattern Recognition, Las Vegas,
  NV, USA, June 27-30, 2016}.\hskip 1em plus 0.5em minus 0.4em\relax {IEEE}
  Computer Society, 2016, pp. 770--778.

\bibitem{ronneberger2015u}
O.~Ronneberger, P.~Fischer, and T.~Brox, ``U-net: Convolutional networks for
  biomedical image segmentation,'' in \emph{Medical Image Computing and
  Computer-Assisted Intervention, Munich, Germany, October 5 - 9, 2015}, ser.
  Lecture Notes in Computer Science, N.~Navab, J.~Hornegger, W.~M.~W. III, and
  A.~F. Frangi, Eds., vol. 9351.\hskip 1em plus 0.5em minus 0.4em\relax
  Springer, 2015, pp. 234--241.

\end{thebibliography}

\end{document}